\pdfoutput=1
\documentclass[acmsmall]{acmart}
\usepackage[T1]{fontenc}
\usepackage[utf8]{inputenc}
\usepackage{amsmath}
\usepackage{subcaption}

\usepackage{algorithmic}
\usepackage{textcomp}
\usepackage{array}
\usepackage{fixltx2e}
\usepackage{stfloats}
\usepackage{url}
\usepackage{mathtools}
\usepackage{float}
\usepackage{soul}
\usepackage[normalem]{ulem}

\newcolumntype{P}[1]{>{\centering\arraybackslash}p{#1}}
\usepackage{verbatim}
\usepackage{ltablex}
\usepackage{lscape}
\usepackage[ruled]{algorithm2e}
\usepackage{wrapfig}
\usepackage{wrapfig,lipsum,booktabs}
\usepackage{color,soul}
\usepackage{longtable}
\usepackage{amsmath} 
\usepackage{multirow}
\usepackage{varwidth}
\usepackage{color, colortbl}

\usepackage{graphicx}
\usepackage{stfloats}
\usepackage{tikz}
\usepackage{diagbox}
\usetikzlibrary{shapes,snakes,patterns}
\usetikzlibrary{shapes.gates.logic.US,trees,positioning,arrows}
\usepackage{pgfplots}
\pgfplotsset{width=18cm, height=6cm}
\usepackage{pgf-pie}

\SetAlFnt{\small}
\SetAlCapFnt{\small}
\SetAlCapNameFnt{\small}
\SetAlCapHSkip{0pt}
\IncMargin{-\parindent}

\newcommand{\p}{\textbf{p}}
\newcommand{\x}{\textbf{x}}
\newcommand{\h}{\textbf{h}}

\usepackage{enumitem}
\setlist[itemize]{leftmargin=*}
\setlist[enumerate]{leftmargin=*}

\DeclareRobustCommand{\feng}[1]{\textcolor{green}{\textsf{}{#1}}}

\usepackage{enumitem}
\usepackage{pifont}

\definecolor{Gray}{gray}{0.9}
\newcommand{\fully}{\ding{52}}
\newcommand{\partly}{\ding{115}}
\newcommand{\notso}{\ding{56}}

\usepackage{comment}
\usepackage{bm}
\expandafter\newcommand\csname r@tocindent4\endcsname{4in}
\usepackage{cancel}

\AtBeginDocument{%
  \providecommand\BibTeX{{%
    \normalfont B\kern-0.5em{\scshape i\kern-0.25em b}\kern-0.8em\TeX}}}

\setcopyright{acmcopyright}
\copyrightyear{2022}
\acmYear{2022}

\begin{document}
\title{A Survey on Uncertainty Reasoning and Quantification for Decision Making: Belief Theory Meets Deep Learning}


\author{Zhen Guo*, Zelin Wan*, Qisheng Zhang}
\authornote{The authors with * have equally contributed to this work.}
\email{{zguo, zelin, qishengz19}@vt.edu}
\orcid{https://orcid.org/0000-0002-6563-5934, https://orcid.org/0000-0001-5293-0363, https://orcid.org/0000-0001-8785-8437}
\affiliation{%
  \department{Department of Computer Science}
  \institution{Virginia Tech}
  \streetaddress{7054 Haycock Road}
  \city{Falls Church}
  \state{VA}
  \country{USA}
  \postcode{22043}
}

\author{Xujiang Zhao*, Feng Chen}
\email{{xujiang.zhao, feng.chen}@utdallas.edu }
\orcid{https://orcid.org/0000-0003-4950-4018, https://orcid.org/0000-0002-4508-5963}
\affiliation{%
  \department{Department of Computer Science}
  \institution{University of Texas at Dallas}
  \streetaddress{800 W Campbell Rd}
  \city{Richardson}
  \state{TX}
  \country{USA}
  \postcode{75080}
}

\author{Jin-Hee Cho, Qi Zhang}
\email{{jicho, qiz21}@vt.edu}
\orcid{https://orcid.org/
0000-0002-5908-4662, https://orcid.org/0000-0002-3607-3258}
\affiliation{%
  \department{Department of Computer Science}
  \institution{Virginia Tech}
  \streetaddress{7054 Haycock Road}
  \city{Falls Church}
  \state{VA}
  \country{USA}
  \postcode{22043}
}

\author{Lance M. Kaplan}
\email{lance.m.kaplan.civ@army.mil}
\orcid{https://orcid.org/0000-0002-3627-4471, https://orcid.org/}
\affiliation{%
  \institution{US Army Research Laboratory}
  \streetaddress{2800 Powder Mill Rd.}
  \city{Adelphi}
  \state{MD}
  \country{USA}
  \postcode{20783}
}

\author{Dong H. Jeong}
\email{djeong@udc.edu}
\orcid{https://orcid.org/0000-0001-5271-293X}
\affiliation{%
  \department{Department of Computer Science of Information Technology}
  \institution{University of the District of Columbia}
  \streetaddress{4200 Connecticut Ave NW}
  \state{Washington, DC}
  \country{USA}
  \postcode{20008}
}

\author{Audun J{\o}sang}
\email{audun.josang@mn.uio.no}
\orcid{https://orcid.org/0000-0001-6337-2264}
\affiliation{%
  \department{Department of Informatics}
  \institution{University of Oslo}
  \streetaddress{Ole-Johan Dahls hus Gaustadalléen}
  \city{Oslo}
  \country{Norway}
  \postcode{23b 0373}
}


\renewcommand{\shortauthors}{Guo et al.}

\begin{abstract}
An in-depth understanding of uncertainty is the first step to making effective decisions under uncertainty. Deep/machine learning (ML/DL) has been hugely leveraged to solve complex problems involved with processing high-dimensional data. However, reasoning and quantifying different types of uncertainties to achieve effective decision-making have been much less explored in ML/DL than in other Artificial Intelligence (AI) domains. In particular, belief/evidence theories have been studied in KRR since the 1960s to reason and measure uncertainties to enhance decision-making effectiveness.  We found that only a few studies have leveraged the mature uncertainty research in belief/evidence theories in ML/DL to tackle complex problems under different types of uncertainty.  In this survey paper, we discuss several popular belief theories and their core ideas dealing with uncertainty causes and types and quantifying them, along with the discussions of their applicability in ML/DL. In addition, we discuss three main approaches that leverage belief theories in Deep Neural Networks (DNNs), including Evidential DNNs, Fuzzy DNNs, and Rough DNNs, in terms of their uncertainty causes, types, and quantification methods along with their applicability in diverse problem domains.  Based on our in-depth survey, we discuss insights, lessons learned, limitations of the current state-of-the-art bridging belief theories and ML/DL, and finally, future research directions.
\end{abstract}


\maketitle


\section{Introduction} \label{sec:intro}

\subsection{Motivation} \label{subsec:motivation}

In all sorts of business processes and our private life, we are confronted with various kinds of decisions involving multiple choices and relative uncertainty. A clear understanding of the uncertainty is a prerequisite of sound and effective decision making.  Although the topic of reasoning and decision making under uncertainty has been studied for decades in various artificial intelligence (AI) domains, including belief/evidence theory, game theory, and machine/deep learning (ML/DL), the different manifestations of uncertainty based on its root causes have not been investigated in-depth.  The era of the Internet and Big Data has brought a flood of information which can be leveraged for decision making. Under such a situation, the challenge for timely, accurate decision making is no longer the lack of information, but the risk from a lack of understanding and managing inherent uncertainty resulting from unreliable, incomplete, deceptive, and conflicting information. 

In AI, a series of belief or evidence theories have a long history studying reasoning and/or decision making under uncertainty.  However, there has been still limited understanding since uncertainty is not caused only by a lack of evidence or unpredictability.  In addition, ML/DL algorithms have considered uncertainty (e.g., aleatoric or epistemic uncertainty) to offer solutions for effective decision making.  However, there has been no common, solid understanding of multidimensional uncertainty where each domain has a different and/or limited understanding in uncertainty even if they are in the pursuit of a common goal of effective decision making. 

Our survey paper aims to conduct an in-depth survey on a series of belief models and introduce a new solution domain that can leverage uncertainty research in belief/evidence theory to develop ML/DL solutions for attaining effective decision making. In particular, we are interested in quantifying different types of uncertainty caused by different root causes. This will help provide solutions for ML/DL that can meet explainable AI, the so-called XAI, by providing how uncertainty derives from, what is the reason behind, and ultimately how it affects the effectiveness of decision making.

The state-of-the-art decision making research has been fully recognized the significant importance of considering uncertainty for effective decision making. However, there has been little study that has extensively surveyed existing belief models to study uncertainty and its applicability for decision making in the ML/DL domain. 

\subsection{Comparison of Our Survey Paper with Other Existing Survey Papers} \label{subsec:comparison-surveys}

In this section, we discuss existing survey papers that have discussed uncertainty research. And then, we identify the key differences between the existing survey papers and our survey paper.  

\citet{li2012dealing} discussed the causes of different uncertainties and how to process them in belief models for making effective decisions in various domains.  According to the nature of aleatory and epistemic uncertainty, they classified uncertainty types processing in probability theory, fuzzy theory, information-gap theory and derived uncertainty theory for their comparison.  They focused on how different uncertainties can be processed in data management techniques.  \citet{kabir2018neural} surveyed prediction interval techniques using deep neural networks (DNNs). The prediction interval techniques quantify the level of uncertainty or randomness and have been widely applied in the medical and electricity fields.  They discussed aleatoric and epistemic uncertainty (see Section~\ref{sec:causes-types-uncertainty} for their definitions) to explain uncertainty in prediction using DNNs.  They also discussed how a Bayesian method is used to optimize the weight of an NN during training and applied for NN-based prediction intervals in various fields.

\citet{Hariri19-survey} surveyed various AI techniques, including ML, Natural Language Processing (NLP), and computational intelligence, that can recognize and reduce uncertainty in Big Data.  \citet{Abdar21-info-fusion} reviewed over 700 papers that studied uncertainty quantification in ML/DL. They mainly focused on discussing Bayesian and ensemble techniques and their applications in various related areas, such as image processing, computer vision, medical applications, NLP, and text mining. 

We also discuss the contributions of the survey papers focusing on the uncertainty mainly in ML/DL in the following papers.  \citet{hullermeier2021aleatoric} distinguished aleatoric uncertainty from epistemic uncertainty.  They explained how these two uncertainties are represented in various ML problems or models and can contribute to decision making under the assessed uncertainty.  \citet{ulmer2021survey} surveyed the methods of quantifying uncertainty in evidential deep learning model based on the conjugate prior and posterior distributions and unknown outlier samples.  This model estimates uncertainty from the Dirichlet distribution by data (aleatoric) uncertainty, model (epistemic) uncertainty, and distributional uncertainty.  \citet{gawlikowski2021survey} provided a comprehensive survey on the uncertainty in DNNs. They discussed two types of uncertainties: {\em reducible uncertainty} and {\em unreducible uncertainty}.'  The concept of reducible uncertainty is aligned with that of epistemic uncertainty where the reducible uncertainty can be introduced by variability in a real-world, errors in model structure, or in training parameter (i.e., batch size, optimizer).  Unreducible uncertainty means uncertainty by noises in measurement (i.e., sensor noise) and is in line with aleatoric uncertainty. The authors classified uncertainty estimation methods based on the cross-combination of the nature (i.e., deterministic or stochastic) and number (i.e., single or multiple) of DNNs. Since the discussions of uncertainty in \cite{hullermeier2021aleatoric, ulmer2021survey, gawlikowski2021survey} are very limited in the scope, we did not include them in Table~\ref{tab:Comparison}.

Unlike the existing survey papers above~\cite{li2012dealing, kabir2018neural, Hariri19-survey, Abdar21-info-fusion}, our paper provides an in-depth survey of eight different belief models with emphasizing how to reason and quantify uncertainty based on the root causes and types of the uncertainty. In addition, we discussed how a belief model is applicable in the DL domain. This will allow researchers to leverage both the solid methodologies of uncertainty reasoning/quantification in belief models and DL techniques for attaining effective decision making.  Finally, in Table~\ref{tab:Comparison}, we summarize the key differences between our survey paper and the existing four survey paper on uncertainty research. We selected the key criteria based on the common discussion points covered by the existing survey papers considered in this paper as well as the key discussion points made in our survey paper.

\begin{table*}[t]
\centering
\scriptsize
\caption{\sc Comparison of Our Survey Paper with the Existing Surveys on Uncertainty Research}
\label{tab:Comparison}
\vspace{-4mm}
\begin{center}
\begin{tabular}{|p{5cm}|P{1.3cm}|P{1.3cm}|P{1.3cm}|P{1.3cm}|P{1.3cm}|}
\hline
\multicolumn{1}{|c|}{\bf Key Criteria} & {\bf Our Survey (2022)} &{\bf \citet{li2012dealing} (2012)}& {\bf \citet{kabir2018neural} (2018)} &  {\bf \citet{Hariri19-survey} (2019)} & {\bf \citet{Abdar21-info-fusion} (2021)}  \\
\hline
Ontology of uncertainty &\fully &\partly & \partly  & \partly  &  \partly\\
\hline
Causes of uncertainty &\fully &\fully & \fully  & \partly &  \fully\\
\hline
Uncertainty reasoning \& quantification in DST &\fully &\fully & \notso & \notso &  \notso\\
\hline
Uncertainty reasoning \& quantification in TBM &\fully &\notso & \notso & \notso &  \notso\\
\hline
Uncertainty reasoning \& quantification in DSmT &\fully &\notso & \notso & \notso &  \notso\\
\hline
Uncertainty reasoning \& quantification in IDM  &\fully &\partly & \notso & \notso &  \notso\\
\hline
Uncertainty reasoning \& quantification in TVL &\fully &\partly & \notso & \notso &  \notso\\
\hline
Uncertainty reasoning \& quantification in Fuzzy Logic
&\fully &\fully & \notso & \partly &  \notso\\ \hline
Uncertainty reasoning \& quantification in Bayesian Inference &\fully &\fully & \notso & \partly & \partly \\
\hline
Uncertainty reasoning \& quantification in Subjective Logic &\fully &\notso & \notso & \partly & \notso \\
\hline
Uncertainty reasoning \& quantification in Bayesian Deep Learning &\fully &\notso &\partly & \notso &  \fully\\
\hline
Applicability of Belief Models in Deep Learning &\fully &\notso & \partly & \notso &  \notso\\
\hline
Discussions of insights, lessons, and limitations of the existing uncertainty-aware approaches &\fully &\partly & \partly & \partly &  \fully\\
\hline
Discussions of future research directions &\fully &\fully & \fully & \fully &  \fully\\
\hline
\end{tabular}

\fully: Fully addressed; \partly: Partially addressed; \notso: Not addressed at all; DST: Dempster-Shafer Theory; TBM: Transferable Belief Model, DSmT: Dezert-Smarandache Theory; TVL: Three-Valued Logic.
\end{center}
\vspace{-7mm}
\end{table*}

\subsection{Research Questions} \label{subsec:research-questions}
In this work, we aim to answer the following research questions:
\begin{description}[leftmargin=9mm]
\item[RQ1.] {\em What are the key causes and types of uncertainty studied in belief theory and deep learning?}
\item[RQ2.] {\em How can the ontology of uncertainty be defined based on the multidimensional aspects of uncertainty studied in belief models and deep learning?}
\item[RQ3.] {\em How has each belief model considered and measured uncertainty?}
\item[RQ4.] {\em How has each belief model been applied in deep learning and vice-versa for effective decision making under uncertainty?}
\item[RQ5.] {\em What are the key differences of uncertainty reasoning and quantification in belief theory and deep learning?}
\item[RQ6.] {\em How can belief model(s) be applied in deep learning to solve complicated decision making problems?}
\end{description}
The research questions above will be answered in Section~\ref{sec:concluding-remarks}.

\subsection{Scope \& Key Contributions} \label{subsec:contributions-scope}

Although uncertainty has been considered in various domains, we limit the scope of our paper to belief models and their applications in DL algorithms.  Note that when we refer to `decision making', we mean a choice among multiple alternatives. For example, it can be a certain class in classification tasks to maximize prediction accuracy, an action chosen among multiple actions available to maximize a decision utility, or a strategy chosen for optimizing system performance.

In this paper, we made the following {\bf key contributions}: 
\begin{enumerate}[leftmargin=*]
\item We are the first conducting an extensive survey on identifying the causes and types of uncertainty studied in various belief models and deep learning and provides the ontology of uncertainty.
\item We first investigate how various belief theories have considered uncertainty and quantified it for effective decision making.
\item We also first discuss how belief theories can be effectively leveraged for deep learning-based solutions for decision making.
\item We identify the key commonalities and differences about how each belief theory reasons and quantifies uncertainty and how it is applied in the context of deep learning or along with it.
\item We provide the overall perspectives of insights and lessons learned as well as the limitations from our extensive survey and suggest promising future research directions.
\end{enumerate}

\subsection{Structure of the Paper}
The rest of the paper is organized as follows:
\begin{itemize}
\item Section~\ref{sec:causes-types-uncertainty} provides various classification types of uncertainty, the causes of different types of uncertainties, and a proposed uncertainty ontology based on the surveyed multidimensional concepts of uncertainty.

\item Section~\ref{sec:dm-belief-theory} provides the details of eight belief models and discusses belief formation, causes and types of uncertainty, uncertainty quantification, and its application as decision making application. The eight belief models include Dempster Shafer Theory (DST), Transferable Belief Model (TBM), Dezert-Smarandache Theory (DSmT), Imprecise Dirichlet Model (IDM), Kleene's Three-Valued Logic (TVL), Fuzzy Logic (FL), Bayesian Inference (BI), and Subjective Logic (SL).

\item Section~\ref{sec:belief-theory-meets-DL} discusses how a belief theory can be applied in the context of DL as decision making applications under uncertainty, particularly in terms of evidential neural networks, fuzzy deep neural networks, and rough deep neural networks.

\item Section~\ref{sec:summary-findings} provides the answers to the key research questions raised in Section~\ref{sec:intro}.

\item Section~\ref{sec:concluding-remarks} concludes our paper by discussing the limitations, insights, and lessons learned from our survey. In addition, we suggest promising future research directions in applying belief models to solve DL-based decision making problems.
\end{itemize}

{\bf Caveat of Mathematical Notations}: In Sections~\ref{sec:dm-belief-theory} and~\ref{sec:belief-theory-meets-DL}, we discuss a set of belief theories and deep learning theories leveraging belief models, including Subjective Logic, Fuzzy theory, and rough set theory. The discussion of a theory needs to use mathematical notations which are only used under the theory, not other theories. We keep the mathematical notations of original papers in order to deliver the common notations used to discuss each theory in the literature.

\section{Classification Types, Causes, and Ontology of Uncertainty} \label{sec:causes-types-uncertainty}

In this work, we deal with uncertainty in data or information.  We define an {\em uncertainty type} as a perceived state of data or information, such as fuzziness, discord, non-specificity, ambiguity, and so forth (see Section~\ref{subsec:classification-types-uncertainty}). We define the {\em causes of uncertainty} by the reason introducing uncertainty in a decision maker's judgment.  We also discuss the ontology of uncertainty where ontology is studied as a branch of philosophy which is defined as the ``science of what it is'' describing \feng{\sout{``}}the structures of objects, properties, events, processes, and relations in every area of reality~\cite{smith2012ontology}. In this section, we will describe the ontology of uncertainty in terms of its types, causes, and outcomes of decision making based on uncertainty reasoning and quantification.

\subsection{Classification of Uncertainty Types} \label{subsec:classification-types-uncertainty}

In the probabilistic uncertainty research community~\cite{Josang16, Kiureghian09}, two types of {\em uncertainty natures} are widely used and commonly discussed by:
\begin{itemize}
\item {\em Aleatoric uncertainty}: This refers to statistical uncertainty about the long term relative frequencies of possible outcomes~\cite{Josang16}.  For example, if we do not know whether a dice is loaded -- and thereby unfair -- then we are faced with aleatoric uncertainty. This uncertainty can be reduced to the true variance about the loaded dice by throwing the dice sufficiently many times. However, every time a dice is thrown, we cannot predict its outcome exactly but can have only a long-term probability~\cite{Josang16}. In this sense, the long-term probability can reduce epistemic uncertainty by more and more observations.  Therefore, aleatoric uncertainty is fundamentally related to the nature of randomness in which a variable is governed by a frequentist process~\cite{Kiureghian09, Josang16}. 

\item {\em Epistemic uncertainty}: This uncertainty is related to a situation that we cannot predict exactly an event because of a lack of knowledge.  A typical example is the assassination of President Kennedy in 1963~\cite{Josang16}, where the uncertainty is about whether he was killed by Lee Harvey Oswald and who organized it. The nature of epistemic uncertainty derives from a lack of knowledge or information (or data).  This type of uncertainty is also called {\em systematic uncertainty} or {\em model uncertainty}.  This means that the outcome of a specific future or past event can be known, but there is insufficient evidence to support it. The epistemic uncertainty can be reduced by more evidence, more advanced technology, and/or scientific principles to interpret the evidence (e.g., forensic science)~\cite{Kiureghian09}. This epistemic uncertainty follows a non-frequentist process representing the likelihood of an event~\cite{Josang16}.
\end{itemize}
Since the above two natures of uncertainty have been most widely discussed as the nature of uncertainty, we will discuss how a different type of uncertainty in different belief models and DL models is related to these two natures of uncertainty.

Uncertainty reasoning and quantification research has been heavily explored by several theories, such as probability theory, fuzzy sets theory, possibility theory, evidence theory, and rough sets theory. These theories can be seen as complementary as each of them is designed to deal with different types of uncertainty. In these theories, \citet{dubois1980fuzzy} identified three uncertainty types in terms of {\em fuzziness}, {\em discord}, and {\em nonspecificity}. The latter two terms, discord and non-specificity, are combined as the term {\em ambiguity}. Each type is represented by a brief common-sense characterization and several pertinent synonyms as follows:
\begin{itemize}
\item {\em Fuzziness}: This results from a lack of definite or sharp distinctions and has pertinent synonyms, such as vagueness, cloudiness, haziness, unclearness, indistinctness, or sharplessness.  
\item {\em Ambiguity}: In general, there exists \textit{ambiguity} when an object cannot be specified due to a lack of certain distinctions or detected as a single class due to conflicting evidence. Hence, these two cases can be further classified into the following two subclasses:
\begin{itemize}
\item {\em Discord}: This is associated with disagreement among several alternatives and interchangeably used with synonyms including dissonance, incongruity, discrepancy, or conflict. 
\item {\em Nonspecificity}: This refers to a situation in which two or more alternatives are left unspecified and is used with pertinent synonyms, such as variety, generality, diversity, equivocation, and imprecision. 
\end{itemize}
\end{itemize}
Based on our understanding, fuzziness introduces vagueness (i.e., failing in distinguishing one from another) while ambiguity belongs to epistemic uncertainty. In addition to the above, the most common uncertainty is also derived from a lack of evidence, called {\em vacuity} (or ignorance)~\cite{Josang16} as we do not know how to make a decision because of insufficient evidence, which belongs to epistemic uncertainty as more evidence can reduce vacuity.\\
\vspace{-5mm}
\begin{wrapfigure}{r}{0.7\textwidth}
\vspace{-2mm}
\centering
\includegraphics[width=0.7\textwidth]{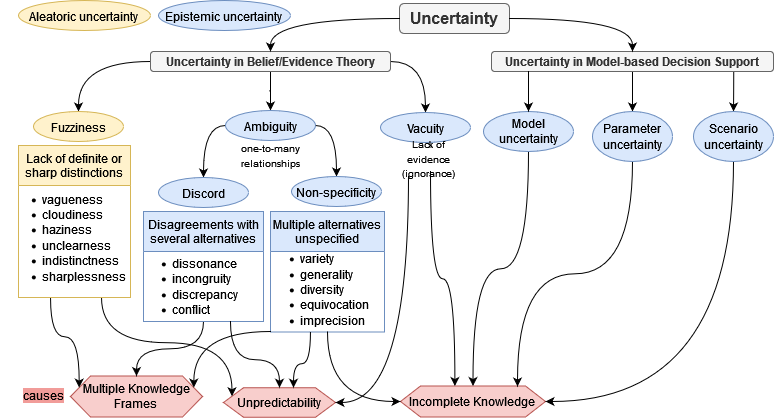}
\vspace{-5mm}
\caption{Classification of uncertainty types.}
\label{fig:uncertainty-types}
\vspace{-7mm}
\end{wrapfigure}

In the modeling and risk assessment research~\cite{linkov2003model, walker2003defining}, uncertainty associated with choices made by modelers has been studied, such as differences in problem formulation, model implementation, and parameter selection originating from subjective interpretation of the problem at hand. We call this `modeler uncertainty' which has been categorized by the following three types:
\begin{itemize}
\item {\em Parameter uncertainty}: This refers to uncertainty derived from the values of input parameters in a model, such as measurement errors, sampling errors, variability, and use of surrogate data. Hence, it is a type of epistemic uncertainty and can be reduced by collecting more reliable evidence to more accurately estimate the parameters used in the model. 
\item {\em Model uncertainty}: This indicates uncertainty about a model structure and the mathematical relationships of components defined in the model. For example, uncertainty can be introduced by making assumptions and simplifying mathematical equations in modeling real-world problems. Thus, this uncertainty is introduced by missing or incomplete information, which makes hard to fully define the model. This uncertainty belongs to epistemic uncertainty and can be reduced by gathering necessary, reliable information to accurately define the model.
\item {\em Scenario uncertainty}: This represents uncertainty caused by normative choices made on constructing scenarios, including the choice of a functional unit, time horizon, geographical scale, and other methodological choices. This uncertainty arises from uncertain problem formulations and theoretic assumptions, which are not statistical in nature. Due to this nature of uncertainty, we understand this uncertainty as epistemic uncertainty. This uncertainty can be reduced by collecting more evidence, using more advanced technology, and/or considering scientific principles to interpret the evidence, such as finding a better choice of a time horizon or a geographical scale for crime hotspot detection.
\end{itemize}
\subsection{Causes of Uncertainty} \label{subsec:causes-uncertainty}
From the perspective of framing research in decision making, where frames are heuristic representations of the external world, there are three main causes of uncertainty, including unpredictability, incomplete knowledge, and multiple knowledge frames~\cite{brugnach2008toward}.  From an engineering perspective, these three causes can be introduced by a lack of evidence, limited cognition to process a large amount of evidence, conflicting evidence, ambiguity, measurement errors, and subjective beliefs~\cite{zimmermann2000application}. The three causes are described by: 
\begin{itemize}
\item {\em Unpredictability}: A system (or entity/data) may exhibit chaotic, variable behavior in space/time.  In Statistics, confidence intervals have been used as a measure of uncertainty~\cite{zimmermann2000application}.  Statistical noise is a common  factor triggering uncertainty, leading to unpredictability.  Even if the system learns and adapts to dynamic, new conditions, it exhibits highly variable behaviors.  The variability may be due to unreliability in information, data, or an entity, which is caused by system/network dynamics, non-stationary environmental conditions, or adversarial attacks.  If this is the case, this type of uncertainty can be reduced by detecting and excluding the unreliable sources or data in decision making process~\cite{zimmermann2000application}. 
\item {\em Incomplete Knowledge}: This refers to situations where we do not know enough about a system to be managed or our knowledge about the system is incomplete (i.e., epistemic uncertainty)~\cite{zimmermann2000application}. This can be due to a lack of evidence (e.g., information/data) or a lack of knowledge because we may not have sufficient theoretical understanding (e.g., ignorance) or reliable information or data. This uncertainty can be reduced by considering more evidence or discarding unreliable evidence. In addition, when human decision makers receive a large amount of information, which is often highly complex, they cannot process them properly because of their limited cognition and processing power. To deal with this, people usually transform available data into information with a rougher `granularity' or focus on important features, neglecting other less important (or noisy) information or data. This uncertainty can be reduced by considering relevant information among the available information~\cite{zimmermann2000application}. 

\item {\em Multiple Knowledge Frames}: This refers to the case when same information (e.g., evidence or opinions) is interpreted differently, resulting in different, conflicting views.  \citet{Dewulf05} defined {\em ambiguity} as the presence of multiple, valid beliefs about a certain phenomenon. The ways of understanding the system (or the external world) can differ in where to put the boundaries of the system or what and who to put as the focus of attention. The differences can also emerge from the way in which the information about the system is interpreted.  Another major cause is \textit{conflicting evidence}, representing a situation where some of the information available may be incorrect, simply irrelevant, or the model to observe a system may not be correct at a given time. Further, multiple observers may provide different opinions based on their subjective views.
\end{itemize}
We demonstrated our view about the classifications of uncertainty types based on the existing classifications in Fig.~\ref{fig:uncertainty-types}.

\subsection{Ontology of Uncertainty} \label{subsec:ontology-uncertainty}

\begin{figure}
\centering
\includegraphics[width=\textwidth]{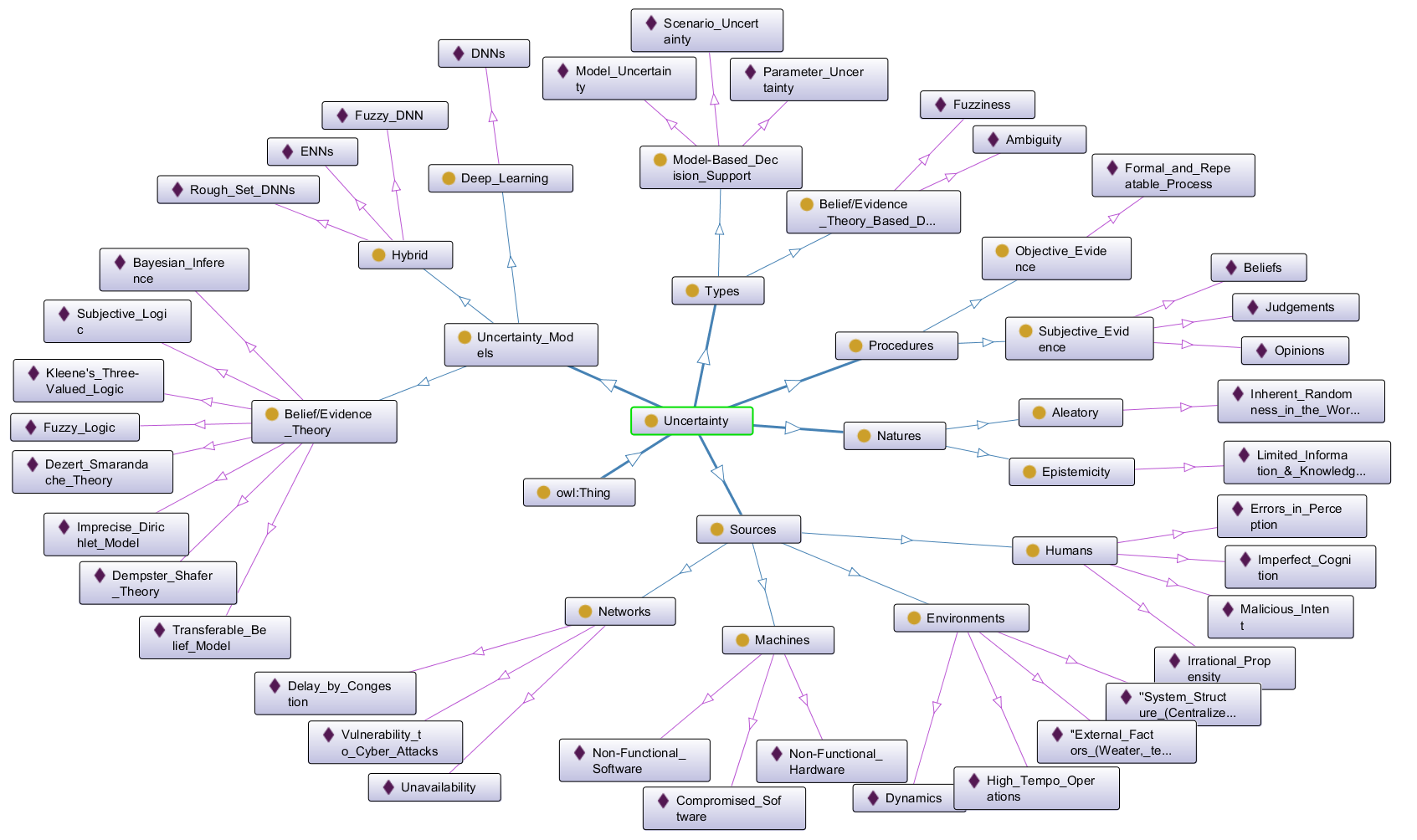}
\vspace{-5mm}
\caption{An Ontology of Uncertainty.}
\label{fig:uncertainty-ontology}
\vspace{-6mm}
\end{figure}

Ontology provides ``a definitive and exhaustive classification of entities in all spheres of being'' where the classification should be able to fully describe the details of the entities~\cite{smith2012ontology}. The information fusion research community has developed ``Uncertainty Representation and Reasoning Framework (URREF)''~\cite{costa2018urref}, which considers uncertainty ontology in the information processing systems. However, its scope was limited only to the information processing systems and only considered a subset of uncertainty types studied in the uncertainty research domain.  In order to more extensively understand the concepts of uncertainty and its multiple causes, we develop an ontology of uncertainty by using W3C Web Ontology Language (OWL)~\cite{owl}, which is demonstrated in Fig.~\ref{fig:uncertainty-ontology}. We do not show the subattributes of ambiguity and fuzziness (see Fig.~\ref{fig:uncertainty-types}) in Fig.~\ref{fig:uncertainty-ontology} due to the space constraint. Hence, we describe more about the key attributes of uncertainty used above in Appendix F of the supplement document. 

\section{Decision Making under Uncertainty in Belief Theory} \label{sec:dm-belief-theory}

Different types of uncertainty affect the assessment and analysis of a specific situation. Underlying uncertainty comes from how to view and model a given part of the world which we call a \emph{domain}. A domain is the abstract representations of states of the world, where analysts or decision makers can have beliefs about the true states of a domain. Beliefs about domains can be easily biased by an analyst or a decision maker, which is often called ``framing effect''~\cite{tversky1985framing, walker2003defining}, which can cause subjective beliefs about the world to deviate from ground truth of the world (e.g., past or future events)~\cite{walker2003defining}. The way a situation is formally modeled (i.e., elements in a domain) can also affect the types and levels of uncertainty perceived by a decision maker.

{\em Belief} has been based off for decision making process. In 1930s, \citet{kleene1938notation} proposed Three-Valued Logic (TVL) by defining algebra based on three values including false, unknown, and truth.  Many other theories defining a belief based on probabilities have been proposed since 1960's, including Fuzzy Logic~\cite{zadeh1965fuzzy}, Dempster-Shafer Theory~\cite{Shafer76}, Transferable Belief Model~\cite{Smets94},  Subjective Logic~\cite{Josang99, Josang01}, Dezert-Smarandache
Theory (DSmT)~\cite{Smarandache04}, Bayesian Inference~\cite{Fienberg06}, Imprecise Dirichlet Model (IDM)~\cite{walley1996inferences}.  The Dempster-Shafer Theory (DST)~\cite{Shafer76} first defined a ``frame of discernment,'' the set of propositions considered.  DST generalized Bayesian theory based on subjective probability~\cite{Shafer76}.  Transferrable Belief Model (TBM)~\cite{Smets94} has been proposed to deal with more knowledge and situations than DST.  \citet{zadeh1965fuzzy} introduced fuzzy set theory to represent a uncertain, subjective belief based on a membership function~\cite{Zadeh83} and has been applied in various trust-based systems~\cite{Nagy08, Lesani06, Chen09-fuzzy, Liao09, Luo08, Manchala98, Nefti05}.  DSmT~\cite{Smarandache04} extended DST to deal with conflicting evidence in trust management systems~\cite{Wang07-dsmt, Deepa14}.  

\subsection{Kleene's Three-Valued Logic (TVL)} \label{subsec:3-1-TVL}
\subsubsection{\bf Belief Formation.}  \citet{kleene1938notation} first proposed TVL in 1938 . Its truth table is shown in Table~\ref{truth_table}, where $p_1$ and $p_2$ are two logical variables. TVL's belief distribution is determined by the logical values of logical variables, i.e., $\mathbf{b}(p)_{q}=1$ when $p=q$ and $0$ otherwise. The $p$ is a logical variable and $q$ is a logical value from $\{T,U,F\}$ where $T$ is {\em true}, $U$ is {\em unknown}, and $F$ is {\em false}.

\begin{wraptable}{l}{.5\textwidth}
\vspace{-4mm}
\centering
\caption{\sc Truth Table of the TVL}
\label{truth_table}
\vspace{-3mm}
\parbox{.45\linewidth}{
\centering
\caption*{$p_1\wedge p_2$}
\vspace{-2mm}
    \begin{tabular}{|c|c|c|c|}
      \hline
      \diagbox{$p_1$}{$p_2$}   & $T$ & $U$ & $F$\\
      \hline
      $T$   & $T$ & $U$ & $F$\\
      \hline
      $U$   & $U$ & $U$ & $F$\\
      \hline
      $F$   & $F$ & $F$ & $F$\\
      \hline
    \end{tabular}
}
\parbox{.45\linewidth}{
\caption*{$p_1\vee p_2$}
\vspace{-2mm}
    \centering
    \begin{tabular}{|c|c|c|c|}
      \hline
      \diagbox{$p_1$}{$p_2$}   & $T$ & $U$ & $F$\\
      \hline
      $T$   & $T$ & $T$ & $T$\\
      \hline
      $U$   & $T$ & $U$ & $U$\\
      \hline
      $F$   & $T$ & $U$ & $F$\\
      \hline
    \end{tabular}
}
\vspace{-5mm}
\end{wraptable}
\vspace{1mm}
\textbf{\em Kleene Algebras and TVL.} Kleene's TVL is a special case of Kleene algebras. The properties of Kleene algebra, $\mathcal{K}=(K,\vee,\wedge,\sim,F,T)$, are: 
\begin{itemize}
\item $\mathcal{K}$ is a bounded distributive lattice; and 
\item $\forall a,b\in K$, $\sim(a \; \wedge b)=\sim a \; \vee \sim b$, $\sim\sim a = a$, and $a \; \wedge \sim a \leq b \; \vee \sim b$. 
\end{itemize}
Here we apply the semantics where $\vee$ means logical disjunction, $\wedge$ means logical conjunction, and $\sim$ means negation. It can be easily derived that Kleene’s TVL, in which $K=\{T,U,F\}$, is a Kleene algebra where $T=\sim F$, $F=\sim T$, and $U=\sim U$.

\vspace{1mm}
\textbf{\em Rough Sets and Kleene Algebras.} Kleene algebras are related to rough sets. The relationships between these two concepts are as follows.  Given an information system, $I=(S,\mathbb{A})$, where $S$ is a set of objects and $\mathbb{A}$ is a set of attributes $a:x\mapsto a(x)$ for any $x\in S$, we can define the set of equivalence relationships, $IND(I)$: $IND(I) = \{IND(A):A\subseteq \mathbb{A}\}$, where $IND(A)=\{(x,y)\in S^2: \forall a\in A,  a(x)=a(y)\}$. Given any equivalence relationship $R\in IND(I)$, a \textit{rough set} $\mathcal{X}\in (S\times S)/R$ is a pair $(\underline{R}X, \overline{R}X)$, where $\underline{R}X$ and $\overline{R}X$ are called the $R$-lower and $R$-upper approximation of $X$, respectively. More specifically,
\begin{eqnarray}
\underline{R}X = \bigcup\{Y\in S/R: Y\subseteq X\}, \; \; 
\overline{R}X = \bigcup\{Y\in S/R: Y\cap X\neq \emptyset\},
\end{eqnarray}
where $S/R$ is the collection of equivalence classes corresponding to $R$.

\begin{wrapfigure}{r}{0.5\textwidth}
\vspace{-3mm}
\centering
\includegraphics[width=0.5\textwidth]{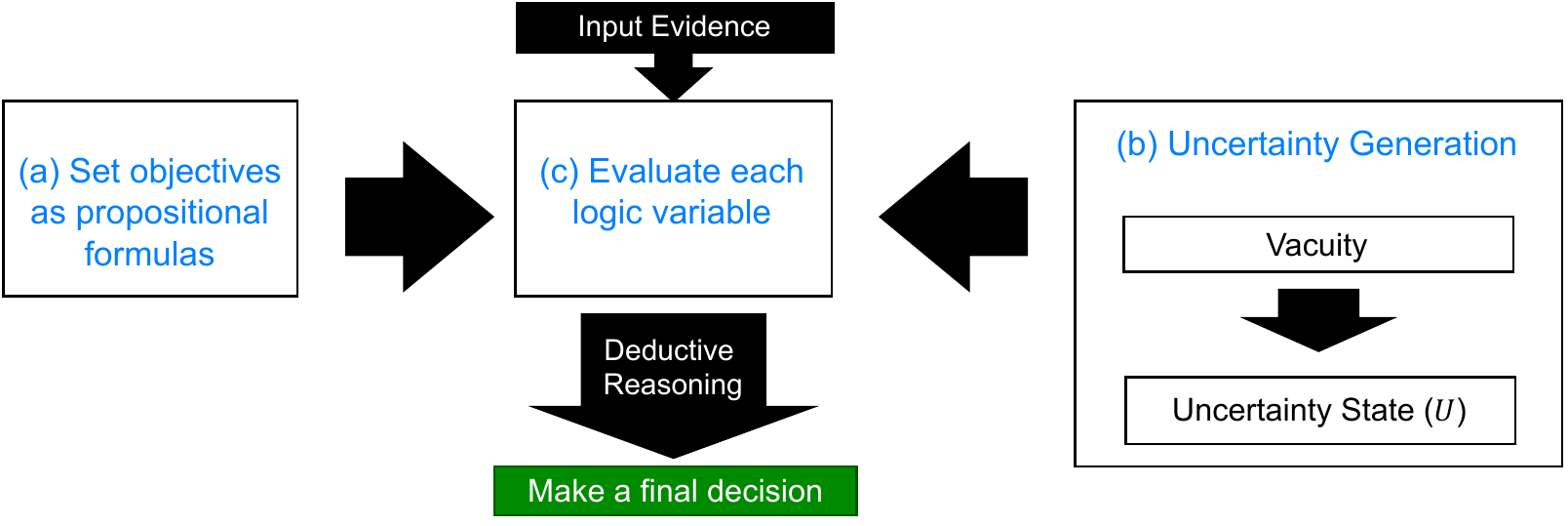}
\vspace{-4mm}
\caption{Uncertainty-aware decision making process using Kleene’s TVL.}
\label{fig:DM-TVL}
\vspace{-6mm}
\end{wrapfigure}

Given any set $S$ with $|S|\geq 2$, universal equivalence relationship $R:=S\times S$ and information system $I=(S,\mathbb{A})$, we can induce a three-valued algebra on a collection of rough sets, $\mathcal{RS}$, with the Kleene semantics by:
\begin{equation}
\mathcal{RS}=\{(\underline{R}A,\overline{R}A): A\subseteq S\}=\{(S, S), (\emptyset, S), (\emptyset, \emptyset)\}. 
\end{equation}
Here, if we define $\sim \mathcal{X}:=(\underline{R}X^c, \overline{R}X^c)$, we have $(S, S)=\sim(\emptyset, \emptyset)$, $(\emptyset, S)=\sim(\emptyset, S)$, and $(\emptyset, \emptyset)=\sim(\emptyset, \emptyset)$. This means $K\cong\mathcal{RS}$.  In general, given the set of all logic functions (propositional formula) denoted by $\mathcal{F}$, the set of all Kleene algebras by $\mathcal{A}_\mathcal{K}$, and the collections of all rough sets over all possible information systems by $\mathcal{A}_\mathcal{RS}$, the following theorem~\cite{kumar2017kleene} is held:
\begin{equation}
\forall \alpha, \beta \in \mathcal{F}, \alpha \vDash_{\mathcal{A}_\mathcal{K}} \beta \Leftrightarrow \alpha \vDash_{\mathcal{A}_\mathcal{RS}} \beta.
\end{equation}
The above can be read by: {\em For any logic functions $\alpha$ and $\beta$ in $\mathcal{F}$, if $\beta$ is a semantic consequence of $\alpha$ in $\mathcal{A}_\mathcal{K}$, then $\beta$ is a semantic consequence of $\alpha$ in $\mathcal{A}_\mathcal{RS}$}.  We summarize the uncertainty-aware decision making process using Kleene's TVL in Fig.~\ref{fig:DM-TVL}.

\subsubsection{\bf Causes and Types of Uncertainty}
Uncertainty is formalized as a logical value $U$ (unknown) and its relationship with two classical logical values $T$ (true) and $F$ (false) are shown in Table~\ref{truth_table}. The stated uncertainty here refers to {\em unpredictability} because of a lack of information or knowledge. For example, in rough sets, due to unpredictable noises, sets are represented by approximation spaces.

\subsubsection{\bf Uncertainty Quantification} 
Uncertainty in TVL represents an unknown or unspecified state in the decision making using TVL. This is related to vacuity uncertainty caused by a lack of information/knowledge or non-specificity.  Since uncertainty is regarded as a logical value, uncertainty value can be quantified through logical operations of logical variables. In Kleene's TVL, the three values of $T$, $U$ (uncertainty), and $F$ are often defined by 1, 0, and -1. As seen in Table~\ref{truth_table}, uncertainty, $U$, can be ignored under $\wedge$ to decide $T$ or $F$ while it can be used to support $T$ over $F$. 

\subsubsection{\bf Applications of TVL on  Machine/Deep Learning}   

\citet{kashkevich1979two} defined a function of TVL to solve classification problems in pattern recognition tasks. They viewed accepted accurate classification as $T$, accepted incorrect classification as $F$, and refused classification as $U$.
\citet{dahl1979quantification} leveraged TVL to construct a database used for natural language consultation. \citet{codd1986missing} applied TVL in the area of SQLs, with ``Null'' value behaving like the uncertain value $U$ in TVL. TVL is rarely observed for its applications in recent research. As the topological generalization of TVL, rough sets are used in ML/DL, as described in Section~\ref{subsec:rdnn}.

\subsection{Dempster Shafer Theory (DST)}
\label{subsec:dst}

DST is a fusion technique for decision-making based on the \textit{belief mass} (a.k.a. evidence) of various detection systems. Each system can be defined as a set of possible conclusions, called \textit{proposition}~\cite{Shafer76}.  The set of all propositions is denoted by $\Theta$ (a.k.a. the \textit{frame of discernment (FOD)}). Given the set $\Theta$, we can generate the power set $P(\Theta)$ (a.k.a. the \textit{powerset of FOD}), where the $P(\Theta)$ represents all possible combination of the set $\Theta$, including an empty set $\emptyset$. So, $2^{|\Theta|}$ is the size of the $P(\Theta)$.  As an example of $P(\Theta)$, if $\Theta=\{W,Z,L\}$, then $P(\Theta) = \{\emptyset,  \{W\}, \{Z\}, \{L\}, \{W,Z\}, \{W,L\}, \{Z,L\}, \{W,Z,L\}\}$.


\subsubsection{\bf Belief Formation}
The belief mass is an observed probability based on evidences. For example, assuming we have a black ball, a black square, and a red ball. The mass for focal element $black$ is $m(black) = \frac{2}{3}$, and the mass for focal element $red$ is $m(red) = \frac{1}{3}$.  For a given system, we assign a belief mass to each element in power set $P(\Theta)$, and defined the mass function as $m:P(\Theta) \rightarrow [0,1]$. The mass function is also called {\em basic belief assignment} (bba), and the sum of the mass for each element in set $P(\Theta)$ is equal to one, that is~\cite{Shafer76}: 
\begin{equation} \label{eq:DST-mass}
m: \Theta \rightarrow [0,1], \sum_{A \in P(\Theta)} m(A) = 1, \; \; \textsf{where} \; \; m(\emptyset) = 0.
\end{equation}

Given the power set $P(\Theta)$ and the corresponding belief mass $m$ for each focal element (i.e., a subset) $A$ in $P(\Theta)$, we can calculate the belief interval of each focal element $A$, and represent it as $[{Bel}(A), {pl}(A)]$. The belief ${Bel}(A)$ is the lower bound and plausibility ${pl}(A)$ indicates the upper bound~\cite{smarandache2012comparative}. The ${Bel}(A)$ and ${pl}(A)$ are obtained by:
\begin{eqnarray}
{Bel}(A) = \sum_{B|B \subseteq A} m(B), \; \; {pl}(A) = \sum_{B|B \cap A \neq \emptyset} m(B), \; \; {Dis}(A) = 1 - {pl}(A).
\label{Eq: belief_and_plausibility}
\end{eqnarray}
\begin{wrapfigure}{r}{0.5 \textwidth}
\centering
\includegraphics[width =0.5 \textwidth]{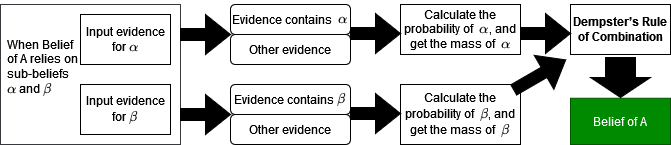}
\vspace{-2mm}
\caption{Dempster's Rule of Combination.}
\label{fig:DST_Combination}
\vspace{-3mm}
\end{wrapfigure}
\begin{wrapfigure}{r}{0.5\textwidth}
\centering
\includegraphics[width = 0.5\textwidth]{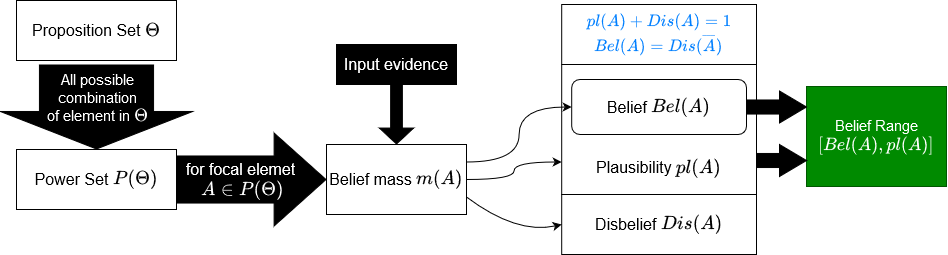}
\caption{Belief, Plausibility, and Disbelief in Dempster Shafer Theory.}
\label{fig:Dempster_Shafer_Theory}
\vspace{-5mm}
\end{wrapfigure}
For example, given $\Theta = \{W,Z,L\}$ and the belief mass $m(W)$, $m(Z)$, $m(W \text{ or } L)$, we can obtain beliefs of focal element $W$ and ($W$ and $Z$) by ${Bel}(W) = m(W)$ and ${Bel}(W \text{ and } Z) = m(W) \cdot m(Z)$, respectively, and plausibility of $W$ by $pl(W) = m(W) + m(W \text{ or } L)$. The belief interval for $W$ is denoted by $[m(W), m(W) + m(W \text{ or } L)]$. 

The disbelief of focal element $A$ is represented as ${Dis}(A)$, which equals ${Bel}(\overline{A})$, where $\overline{A}$ means the complement of $A$ (i.e., negation of $A$). The ${Dis}(A)$ is calculated by summing all masses of the focal elements that do not intersect with $A$. Another way to calculate the ${Dis}(A)$ is ${Dis}(A) = 1 - {pl}(A)$ where ${pl}(A)$ can be considered as an estimated uncertainty as a potential credit to increase the given belief. While uncertainty is often considered risk~\cite{vanAsselt00}, DST uses uncertainty as a credit to support a particular belief.  Fig.~\ref{fig:Dempster_Shafer_Theory} describes the key concept of DST.

\textbf{\em Dempster's Rule of Combination} is a belief mass combination function for two independent detection systems $i$ and $j$ over the same frame. The joint mass committed to focal element $A$ is given by:
\begin{equation}
m(A) = \kappa \sum_{A_i \cap B_j = A \neq \emptyset} m_1(A_i) \cdot m_2(B_j), 
\label{eq:dempster combination}
\end{equation}
where $A_i$ and $B_j$ are values in set $\Theta$ of two different systems $i$ and $j$ that contain target value $A$. The $\kappa$ is a renormalization constant, defined by $\kappa = (1 - \sum_{A_i \cap B_j = \varnothing} m_1(A_i) m_2(B_j))^{-1}$~\cite{Shafer76}.  For example, we have $m_1(W)$, $m_1(Z)$, $m_1(W,Z)$, $m_2(W)$, $m_2(Z)$, and $m_2(W \text{ or } Z)$. The joint mass for focal element $W$ is calculated by $m(W) = m_1(W) \cdot m_2(W) + m_1(W) \cdot m_2(W \text{ or } Z) + m_1(W \text{ or } Z) \cdot m_2(W) + m_1(W \text{ or } Z) \cdot m_2(W \text{ or } Z)$.  

We summarize the key concept of Dempter's rule of combination in  Fig.~\ref{fig:DST_Combination} based on our discussion above.  Many DST variants have been proposed. Due to the space constraint, we discuss some variants of DST in Appendix A of the supplement document.  

\subsubsection{\bf Causes and Types of Uncertainty} DST considers uncertainty in {\em plausibility} due to a lack of evidence.  This implies that uncertainty in DST is closely related to epistemic uncertainty or vacuity. Hence, DST can quantify an uncertain opinion as a subjective belief in a given proposition~\cite{Shafer76}.

\subsubsection{\bf Uncertainty Quantification}
\citet{smarandache2012comparative} measured uncertainty in DST based on its multiple dimensions, including {\em auto-conflict} (i.e., conflict in a belief function with conjunctive rule), {\em non-specificity} (i.e., a generic form of Hartley entropy with base 2), {\em confusion} (i.e., uncertainty by a lack of evidence), {\em dissonance} (i.e., all beliefs are mostly the same), {\em aggregate uncertainty measure (AU)} (i.e., generalized Shannon entropy), and {\em ambiguity measure (AM)} (i.e., non-specificity and discord). We provide the detail of each certainty explained above in Appendix A of the supplement document.  \citet{Blasch13} defined the {\em Interval of Uncertainty} (IOU) by:
\begin{equation}
IOU(A) = pl(A) - Bel(A) = 1 - Bel(\overline{A}) - Bel(A) = 1-Dis(A)-Bel(A).
\label{eq:iou}
\end{equation}
\citet{klir1990uncertainty} measured the total uncertainty in DST, denoted by $U^T(A)$, by considering two types of uncertainty, non-specificity and discord.  Both $U^T(A)$ and AM consider non-specificity and discord and differently capture them.  AM captures them in a level of each proposition (i.e., element $\theta \in \Theta$) while $U^T(A)$ obtains them at the level of sets, $A \subset \Theta$. 
Hence, $U^T(A)$ is given by~\cite{klir1990uncertainty}: 
\begin{equation}
U^T(A) = \sum_{A \subset \Theta} m(A) \log_2 \Bigg(\frac{|A|}{\sum_{B \subset \Theta} m(B) \frac{|A \cap B|}{|B|}}\Bigg).
\end{equation}
The key merit of DST is to combine an amount of uncertain evidence from multiple sources and select elements based on the combined belief mass.  However, the combination rule of DST fails to balance different sources, especially when sources provide conflicting evidence. Although many alternative combination rules have been proposed, \citet{dubois1988representation} argued that no single combination rule could be used as a universal solution to all encountered situations.

\subsubsection{\bf Applications of DST on  Machine/Deep Learning}  DST has been used with deep learning techniques.  \citet{soua2016big} proposed a framework to predict traffic flow using a Deep Belief Network (DBN), a class of deep neural network (DNN), to make an identical prediction based on two types of data, including data streams and event data. Then DST was used to fuse those two predictions.  \citet{tong2021evidential} studied a \textit{set-valued classification} (SVC) where a sample can be classified as multiple classes, not just a single class to identify outliers not represented in a training dataset. The authors proposed a technique of combining DST with Convolutional Neural Networks (CNNs) to improve the accuracy of the SVC. 
\citet{tian2020deep} proposed a new intrusion detection system (IDS) using DST with Long Short-Term Memory Recurrent Neural Network (LSTM-RNN) to combine the results from different classes. 
\citet{zhang2020fault} proposed a new fault diagnosis technique using an improved DST by fusing data from multiple sensors for fault classification.

\subsection{Transferable Belief Model (TBM)} \label{subsec:tbm}
TBM was developed as a variant of DST to resolve unreasonable results of the DST combination rule (see Section~\ref{subsec:dst}) when multiple sources provide conflicting evidence~\cite{Smets94}. TBM is based on the open-world assumption with two levels of belief reasoning: {\em credal level} and {\em pignistic level}. The \textit{credal level} quantifies and updates a belief through a belief function. The \textit{pignistic level} transfers a belief into a probability using the so-called {\em pignistic probability function} for making decision~\cite{Smets94}. 

\subsubsection{\bf Belief Formation} TBM defines basic belief masses the same as DST~\cite{Shafer76} (see Eq.~\eqref{eq:DST-mass}). The {\em credal level belief function} updating a belief upon the arrival of new evidence is formulated by~\cite{Smets94}: 
\begin{equation}
\label{eq:m_B(A)}
m_B(A) = 
\begin{cases}
\frac{\sum_{C \subseteq \overline{B}} m(A \cup C)}{1 - \sum_{C \subseteq \overline{B}} m(C)} \; \; \text{for} \; \; (A \subseteq B) \wedge (A \neq \emptyset); \\
  \; \; \; \;\; \; \;\; 0  \; \; \; \; \; \;\; \; \; \; \; \text{otherwise.} 
    \end{cases}
\end{equation}
Here $m_B(A)$ means the belief mass supporting propositions $A$ when conditional evidence does not support proposition $B$ as the truth~\cite{Smets94}.  The $\sum_{C \subseteq \overline{B}} m(C)$ refers the sum of beliefs supporting a set not supporting $B$ and $\sum_{C \subseteq \overline{B}} m(A \cup C)$ is the sum of beliefs not supporting $B$ or supporting $A$. For $m_B(A): \mathbb{R} \rightarrow [0, 1]$, Eq.~\eqref{eq:m_B(A)} needs to hold $\sum_{C \subseteq \overline{B}} m(A \cup C) < 1 - \sum_{C \subseteq \overline{B}} m(C)$.  

The probability transformed through the pignistic probability function for decision making is denoted by $BetP$, which is estimated by:
\begin{equation} \label{eq:betP-tbm}
BetP(x) = \sum_{x \in A \subseteq X} \frac{m(A)}{|A|} = \sum_{x \in A \subseteq X} m(A) \frac{|x \cap A|}{|A|},
\end{equation}
\begin{wrapfigure}{r}{0.35\textwidth}
\centering
\includegraphics[width=0.3\textwidth]{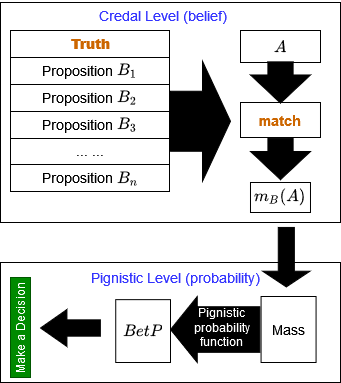}
\vspace{-3mm}
\caption{A belief at the credal level and pignistic level and their relationships in TBM. }
\label{fig:Transferable_Belief_Model}
\vspace{-8mm}
\end{wrapfigure}
where $\frac{m(A)}{|A|}$ means belief mass, $m(A)$, is evenly distributed into the atoms of $A$, a set of atoms, and $|A|$ means the number of atoms $x$ in set $A$ (i.e., $x \in A$). The $X$ is the \textit{Boolean algebra} of the subset of $\Omega$, where $\Omega$ is a set of worlds (truth). The probability distribution calculated from the pignistic probability function is used for decision making.

\subsubsection{\bf Causes and Types of Uncertainty} TBM considers epistemic uncertainty caused by a lack of evidence.

\subsubsection{\bf Uncertainty Quantification} In TBM, uncertainty has not been explicitly discussed. Since a belief at the pignistic level is for decision making in real-world settings, we can understand the pignistic probability function gives a belief mass that considers uncertainty in practice while the credal level belief function estimates a belief based on observed evidence.  We show how a belief is constructed at the credal level based on the arrival of evidence and how the credal level belief is transferred to the pignistic level belief for decision making in Fig.~\ref{fig:Transferable_Belief_Model}.

\subsubsection{\bf Applications of TBM on  Machine/Deep Learning} TBM has been used as a competitive algorithm to solve various types of classification problems~\cite{zhang2020tbm, guil2019associative, quost2005pairwise, honer2018motion, henni2019enhanced}. However, to the best of our knowledge, we have not found any prior work that uses TBM along with ML/DL.

 
\subsection{Dezert-Smarandache Theory (DSmT)} \label{subsec:dsmt}

\citet{Smarandache04} introduced DSmT theory for data and information fusion problems as a general framework that provides new rules of handling highly imprecise, vague, and uncertain sources of evidence and making decisions under them.  The main advantages of DSmT over DST are as follows. First, DSmT has a more general fusion space in a hyper-power set (discussed in the Appendix B.1 of the supplement document), compared to a power set. Second, DSmT fits free and a hybrid model compared to a strict DST model (see Appendix B.1 of the supplement document).  Third, DSmT also combines complex classes based on subsets or complements and introduces better fusion rules, such as proportional conflict redistribution rule 5 (PCR5), dynamic fusion by hybrid DSm rule (DSmH), a new probability transformation, qualitative operators for data with labels (e.g., linguistic labels in natural language), and new belief conditioning rules (BCRs), or new fusion rules for set-valued imprecise beliefs. 

\subsubsection{\bf Belief Formation}\label{subsubsec:dsmt-belief}   A {\em generalized basic belief assignment (gbba)} is formulated the same as a belief function of DST in Eqs.~\eqref{eq:DST-mass} and~\eqref{Eq: belief_and_plausibility}, but the domain of DSmT is the hyper-power set $D^\Theta$, compared to a power set $P(\Theta)$ of DST.  Note that $P(\Theta)\overset{\Delta}{=}(\Theta,\cup)$, $D^\Theta \overset{\Delta}{=}(\Theta,\cup,\cap)$, and $S^\Theta \overset{\Delta}{=}(\Theta,\cup,\cap,c(\cdot))$. If $\Theta=\{a,b\}$, $P(\Theta)\overset{\Delta}{=}(\emptyset,a, b, a\cup b)$, $D^\Theta \overset{\Delta}{=}(\emptyset,a, b, a\cup b,   a\cap b)$, and $S^{\Theta} = (\emptyset,a, b, a\cup b, a\cap b, c(\emptyset),c(a), c(b), c(a\cup b), c(a\cap b))$ where $c(X)$ refers to the complement of $X$.  
We provide the details of other various types of belief mass functions introduced in DSmT in Appendix B.1 of the supplement document.

\subsubsection{\bf Causes and Types of Uncertainty} DSmT handles various uncertainties as follows~\cite{smarandache2009advances}:
\begin{itemize}
\item {\em Precise, uncertain beliefs from multiple sources}: The beliefs from multiple sources contribute uncertainty even if the belief of each proposition is a precise $m(X)$, where each $m(X)$ is only represented by one real number in $[0,1]$ in $D^\Theta$. 
Uncertainty exists when a single source provides beliefs about partial elements or multiple sources provide conflicting beliefs.  For example, for $\Theta=\{\theta_1, \theta_2, \theta_3\}$, two independent sources provide beliefs $m_1(\theta_1)=0.6, m_1(\theta_3)=0.4$ and $m_2(\theta_2)=0.8, m_2(\theta_3)=0.2$, respectively. 

\item {\em Highly conflicting evidence from multiple sources}: If $k$ multiple sources have conflicting evidences toward a same event, there is uncertainty for which source to trust.  For example, for $\Theta=\{\theta_1, \theta_2, \theta_3\}$, two sources provide $m_1(\theta_1)=0.2, m_1(\theta_2)=0.1, m_1(\theta_3)=0.7$ and $m_2(\theta_1)=0.5, m_2(\theta_2)=0.4, m_2(\theta_3)=0.1$.  The decision is based on those conflicting evidence.

\item {\em Imprecise beliefs}: Imprecise beliefs are represented by the admissible imprecise.  Imprecise beliefs can be either quantitative or qualitative.  Quantitative imprecise beliefs $m^I(\cdot)$ are real subunitary intervals of $[0,1]$ or real subunitary sets over $D^\Theta$.  Qualitative $m^I(\cdot)$ is a set of labels $L=\{L_0, L_1, L_2, \ldots, L_m, L_{m+1}\}$ in order.  Imprecise beliefs are common in fusion problems because it is very hard to generate precise sources of evidence. For example, the set of ordered sentiment labels are $L=\{L_0, L_1, L_2\}= \{negative, neutral, positive\}$ and the set of elements is $\Theta=\{\theta_1, \theta_2\}$. The two sources can give qualitative beliefs by sentiment labels as $qm_1(\theta_1)=L_1, qm_1(\theta_2)=L_0$ and $qm_2(\theta_1)=L_2, qm_2(\theta_2)=L_1$, respectively.

\item {\em Subjective probability (DSm probability or DSmP) transformation, fusion space, and fusion rules}: The criteria (i.e., frame $\Theta$), the set of elements (i.e., $G^\Theta$), the choice of combination rule, the probability function, and controllable parameter $\epsilon$ for DSmP, all contribute to the uncertainty that can significantly impact decision making.
\end{itemize}

\subsubsection{\bf Uncertainty Quantification} DSmT does not provide its own uncertainty measure. It borrows other methods and helps decision making using the following uncertainty measures:
\begin{itemize}
\item In probability theory, uncertainty in proposition $A$ can be defined as~\cite{smarandache2012neutrosophic}:
\begin{equation}
U(A) = \sum_{\substack{B\in S^\Theta\backslash\{\emptyset\}, B \cap A\neq\emptyset, B \cap C(A)\neq\emptyset}} m(B),
\label{eq:dsmt-uncertainty}
\end{equation}
where $A$, $B$, and $C$ are three different elements (i.e., propositions), $\Theta$ is a set of the elements, and $S^\Theta$ is a super power set. The $C(A)$ is the complement of $A$.  Uncertainty and IOU can also be defined in the same way as DST in Eq.~\eqref{eq:iou}. 

\item Degree of uncertainty can be evaluated in the probability transformation.  Normalized Shannon's entropy is a measure of uncertainty in probability theory and given by:
\begin{equation}
    E_H = -\frac{\sum_{i=1}^n m(\theta_i)\log_2(m(\theta_i))}{H_{max}},
\end{equation}
where $H_{max}$ is the maximal entropy for the uniform distribution.

\item Probabilistic information content (PIC) score refers to the degree of certainty which can be estimated by $PIC = 1- E_H$. Less uncertainty (or higher certainty) can lead to a correct and reliable decision.
\end{itemize}

For decision making, DSmT extends the probability function, called {\em classical pignistic transformation} (CPT) in DST, into two ways. First, CPT can be generalized to the {\em Generalized pignistic transformation (GPT)}. Second, it can be generalized to a subjective probability measure of $m(\cdot)$ by $\epsilon\geq 0$ in the new probability transformation, $DSmP_{\epsilon}$, which is a probability transformation with a subjective measure, $\epsilon$.  Due to the space constraint, we discuss these two transformation methods in Appendix B of the supplement document. 

\begin{figure}
\centering
\includegraphics[width=0.7\textwidth]{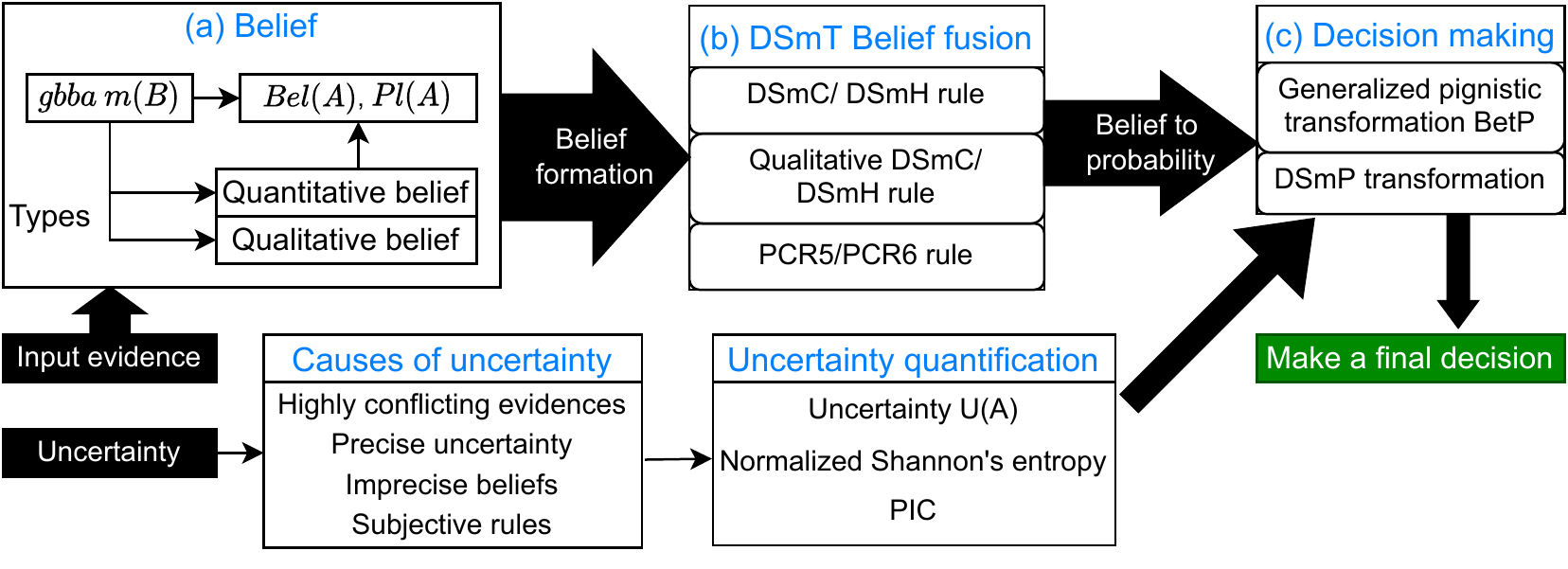}
\vspace{-5mm}
\caption{Uncertainty-aware decision making process using DSmT where the generalized basic belief assignment ($gbba$) is the formal name of $m(\cdot)$ and BetP refers to pignistic transformation in $gbba$ domain.}
\label{fig:DM-DSmT}
\vspace{-5mm}
\end{figure}

Fig.~\ref{fig:DM-DSmT} demonstrates the following steps/criteria to make a decision. In Fig.~\ref{fig:DM-DSmT}, (b) and (c) combines the belief masses from different sources. Uncertainty is not combined into input evidence and is used to make final decisions in (c).
\begin{enumerate}
\item Belief functions and models are defined in the proper frame $\Theta$ of a given problem.  In Appendix B of the supplement document, the closed finite set (i.e., frame), denoted by $\Theta$, has $n$ elements of hypotheses. These steps decide the elements in the given problem.
\item The belief functions are defined in the proper set of $G^\Theta$ (e.g., power, hyper, or super set) where $G^\Theta$ means any set of items, including power set $P^{\Theta}$, hyper power set $D^{\Theta}$, and super power set $S^{\Theta}$. DSmT works on any $G^{\Theta}$, but normally $D^{\Theta}$ is used to distinguish from $P^{\Theta}$ in DST. This step means the choice of $P^{\Theta}$, $D^{\Theta}$, or $S^{\Theta}$.
\item Choose an efficient rule to combine belief functions (in Appendix B.1 of the supplement document) for a given problem.
\item Before making a decision, one must use a probability function (e.g., GPT, see Eq.~(21) in Appendix B.2) or DSmP with a subjective measure) which is from the belief functions.  The maximum of the GPT function can be used as a decision criterion between two choices.  
\end{enumerate}
Making decisions by DSmP can improve the previous probabilistic transformations and increase the strength of a critical decision from the total knowledge.

\subsubsection{\bf Applications of DSmT on  Machine/Deep Learning} 
DSmT covers broad information fusion topics of data and sensors in robotics, biometrics, image fusion, trust management, situation analysis, or object tracking.  Various applications such as DSmH (hybrid), DSmP (probabilistic), and DSmT coordinate between machine processing and user coordination~\cite{smarandache2009advances}.  DSmT serves as an information fusion tool in binary class and multi-class classification problems in conjunction with ML/DL models~\cite{abbas2015effective, ji2021dsmt}.  As an extension to Support Vector Machine One-Against-All (SVM OAA) model for multi-class classification, DSmT models partial ignorance by combining conflicting evidence from two complementary SVM results through PCR6 rule (see Appendix B.1 of the supplement document) and reduces focal elements in the model.  A DSmT-based multi-classifier~\cite{ji2021dsmt} integrates PCR6 fusion rules into layered ML model structures, including Convolutional Neural Network (CNN),  Long Short-Term Memory (LSTM), and Random Forests (RF).  DSmT is applied to the final decision making process by combining multi-signal sources of fault characteristics. 

\subsection{Imprecise Dirichlet Model (IDM)} \label{subsec:IDM}

\citet{walley1996inferences} proposed IDM to derive beliefs based on objective statistical inference from multinomial data with no prior information. The inference is expressed in terms of posterior upper and lower probabilities. A typical application is predicting the color of the next marble from a bag whose contents are initially unknown. Objective Bayesian does not satisfy this principle because predicted outcome is unknown and we cannot formulate the sample space. In IDM, the inferences are expressed as the posterior upper and lower probabilities, $\overline{P}(A|n)$ and $\underline{P}(A|n)$, where $A$ refers to an event and $n$ is the number of observations towards event $A$.  In a multinomial sampling (i.e., $k \geq 2$), the sample space, any event of interest can be identified as a subset of $\Omega$. IDM generates the lower and upper bounds for each value in Beta/Dirichlet PDFs (Probability Density Functions). 

\subsubsection{\bf Belief Formation} \begin{wrapfigure}{r}{0.35\textwidth}
    \centering
    \includegraphics[width=0.35\textwidth]{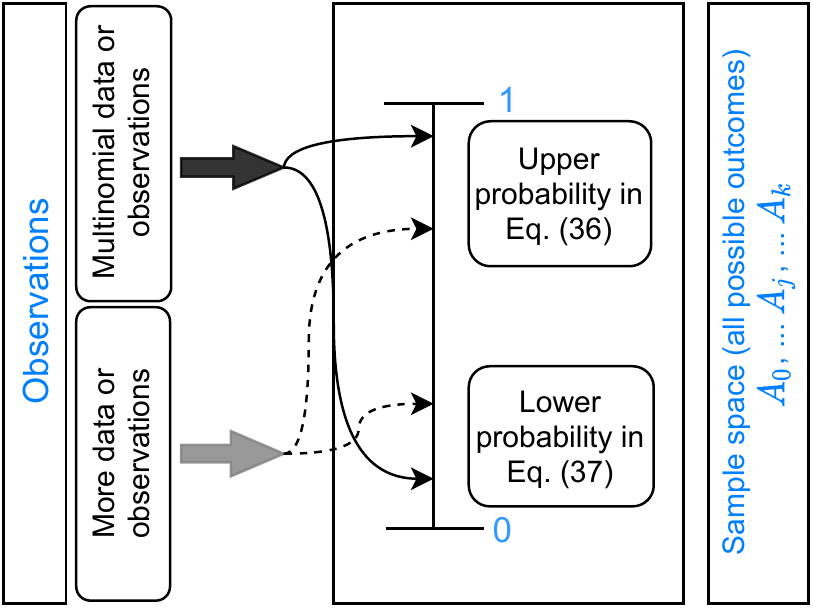}
    \vspace{-3mm}
    \caption{Derivation of the upper and lower bounds in IDM.}
    \label{fig:Imprecise_Dirichlet_Model}
    \vspace{-5mm}
\end{wrapfigure}
According to~\citet{walley1996inferences}, IDM can be defined as the set of all Dirichlet $(s, t)$ distribution, such that $0 < t_j <1$ for $j = 1, 2, \dots, k$ and $\sum_{j=1}^k t_j =1$ and $s$ is a specified positive constant that does not depend on $\Omega$. Walley suggests $s \leq 2$ where $s$ determines how quickly the upper and lower probabilities converge as the observation data accumulate. This is a prior set and denoted as $\mu_0$ to model the prior ignorance about chance $\theta$. Given $\theta=\{\theta_1,\theta_2,\dots,\theta_k\}$, which refers to the identical probability distribution of observations. The corresponding set of a posterior distribution, denoted by $\mu_N$, is composed of all Dirichlet $(N+s, \mathbf{t}^*)$ distribution (i.e.,  $\mathbf{t}^* = \{t_1^*, t_2^*, \ldots, t_j^*, \ldots, t_k^*\}$), where $t^*_j=\frac{n_j+s \times t_j}{N+s}$ and $n_j$ is the number of observations of category $\omega_j$ in $N$ trials.

For example, let $A_j$ be the event with outcome $\omega_j$ from next trial. The predictive probability $P(A_j|n)$ under Dirichlet $(N+s, \mathbf{t}^*)$ is equal to the posterior mean of $\theta$. By maximizing and minimizing $t^*_j$ with respect to $t_j$ (i.e., $t_j \rightarrow 1$ and $t_j \rightarrow 0$), the posterior upper and lower probabilities of $A_j$ are given by:
\begin{eqnarray}
\label{eq:IDM-upper-bound}
\overline{P}(A_j|n) = \frac{n_j+s}{N+s} \;\; \text{for $t_j \rightarrow 1$}; \;\;\;\;
\label{eq:IDM-lower-bound}
    \underline{P}(A_j|n) = \frac{n_j}{N+s} \;\; \text{for $t_j \rightarrow 0$.}    
\end{eqnarray} 

If $s$ is the hidden observation and $N$ is the number of revealed observations, those values can be interpreted as the upper and lower bound of relative frequency of $A_j$. For example, before making any observation, $n_j=N=0$, so that $\overline{P}(A_j|n)=\frac{s}{s}=1$ and $\underline{P}(A_j|n) = \frac{0}{s}=0$.  However, the interval of an IDM bound may be out of range under insufficient evidence ($s$) condition. For example, if a bag has nine red balls and one black ball, we randomly pick a ball and obtain a black ball. Now we have evidence $r(black)=1$, which gives $\underline{P}(black) = \frac{1}{2+1} = \frac{1}{3}$. However, we know that the actual probability of having a black ball is $p(black) = \frac{1}{10}$. So $\underline{P}(black) > p(black)$ when the number of trials is not sufficient. This case shows that actual probability may be outside the range of IDM under a lack of evidence~\cite{Josang16}.

\subsubsection{\bf Causes and Types of Uncertainty} In IDM, uncertainty decreases as a more amount of evidence is received. Hence, it is aligned with the concept of epistemic uncertainty, which can be reduced by increasing an amount of observations (or evidence).

\subsubsection{\bf Uncertainty Quantification} 
In IDM, the uncertainty is associated with the imprecision whose degree is captured by the difference between the posterior upper and lower probabilities as:
\begin{equation} \label{eq:IDM-upper-minus-lower}
\overline{P}(A_j|n) - \underline{P}(A_j|n) = \frac{s}{N+s}.
\end{equation}
From the above, we conclude that the imprecision does not depend on event $A_j$. That is, uncertainty due to the imprecision is based on the amount of hidden observations.

\subsubsection{\bf Applications of IDM on  Machine/Deep Learning} 


%
\citet{utkin2015imprecise} proposed an algorithm called {\em IDMBoost (Imprecise Dirichlet Model Boost)}, which is an improved version of AdaBoost, one of the well-known machine learning algorithms, particularly to improve the overfitting problem with a reduced number of iterations.  \citet{moral2020imprecise} solved imprecise classification problems by proposing an improved algorithm of Credal Decision Trees (CDTs). CDTs are Decision Trees using imprecise probabilities using the Non-Parametric Predictive Inference Model (NPI-M) and showed its outperformance over IDM. \citet{Corani10-idm-ml} presented a tree-augmented na\"{i}ve classifier, called a TANC, based on imprecise probabilities. TANC considered prior near-ignorance using the extreme IDM (or EDM). They proved their TANC provides an efficient and sensible approximation of the global IDM via extensive comparative performance analysis.



\subsection{Fuzzy Logic} 

\L ukasiewicz and Alfred Tarski~\cite{lukasiewicz1930untersuchungen} first proposed \textit{\L ukasiewicz logic}, which is the most typical case of many valued logic.  Our discussion focuses on the real-valued semantics of \textit{{\L}ukasiewicz logic} as the backbone of fuzzy logic. Assume $\alpha$ and $\beta$ are two propositional formulas with truth values $v(\alpha)=x$ and $v(\beta)=y$, we adopt these semantics in the following context:
\begin{eqnarray}
v(\alpha\vee \beta)=\max\{x,y\}, \; \; v(\alpha\wedge \beta)=\min\{x,y\}, \; \; v(\sim\alpha)=1-x,     
\end{eqnarray}
where $\vee$ refers to logical disjunction, $\wedge$ is logical conjunction, and $\sim$ indicates negation.

Fuzzy logic~\cite{zadeh1975fuzzy} is a kind of infinite-valued logic defined on type $1$ fuzzy sets~\cite{zadeh1965fuzzy}.  A fuzzy logic truth value set $\mathscr{T}$ is a set of linguistic truth-values, which is a language generated from  a context-free grammar $G$:
\begin{equation}
\mathscr{T}=L(G).    
\end{equation}
For each truth value $\tau\in\mathscr{T}$, $\tau$ is defined as a fuzzy subset of a truth-value set $l_\tau$ of \textit{\L ukasiewicz logic}, which is given by:
\begin{equation}
\tau = \int_{0}^{1} \frac{\mu_\tau(v)}{v},    
\end{equation}
where $\mu_{l_\tau}:[0,1]\rightarrow c_\tau\in[0,1]$ and $\mu_\tau:[0,1]\rightarrow[0,c_\tau]$ are defined as the membership function of $l_\tau$ and $\tau$, respectively.
Suppose $\tau$ has a finite support set $\{v_1,v_2,\ldots,v_n\}\subset [0,1]$, then we can write:
\begin{eqnarray}
\tau = \frac{\mu_1}{v_1} +\frac{\mu_2}{v_2}+\dots+\frac{\mu_n}{v_n},    
\end{eqnarray}
where $\mu_i = \mu_\tau(v_i)$ for $i\in[1,n]$ and `$+$' stands for an union operation.

Since truth values are fuzzy subsets of truth-value sets of \textit{\L ukasiewicz logic}, logic operations between them can be similarly defined by: 
\begin{eqnarray}
\mu_{\neg\tau_0}=1-\mu_{\tau_0},\; \; \mu_{\tau_0\vee\tau_1}=\max\{\mu_{\tau_0}, \; \; \mu_{\tau_1}\},\mu_{\tau_0\wedge\tau_1}=\min\{\mu_{\tau_0},\mu_{\tau_1}\},
\end{eqnarray}
where $\tau_0, \tau_1 \in \mathscr{T}$. From this, we can then derive $\mu_{\tau_0\Rightarrow\tau_1}=\mu_{\neg\tau_0\vee\tau_1}=\max\{1-\mu_{\tau_0},\mu_{\tau_1}\}$ based on \textit{Kleene-Dienes implication}.

In general, a type $n$ fuzzy set has a membership function defined based on the set of fuzzy sets of type $n-1$, where $n \geq 2$.
Fuzzy numbers~\cite{zadeh1975concept} can also be formulated as instances of fuzzy sets. In other words, each fuzzy number is attached to a membership function, that defines a fuzzy set.

A decision making process under fuzzy logic often consists of three phases: fuzzification, inference, and defuzzification.  A fuzzifier transforms crispy data into fuzzy sets. An inference engine does the logical deduction based on given fuzzy rules. A defuzzifier transforms the fuzzy relationships to crispy relationships and makes a final decision.

\subsubsection{\bf Belief Formation}

\citet{zadeh1968probability} 
defined $P(A)$ as the probability of a fuzzy event $A$:
\begin{equation}
P(A)=\int_{\mathbb{R}^n}\mu_A(x)dP=E(\mu_A),    
\end{equation}
where $A\subseteq \mathbb{R}^n$, $\mu_A:\mathbb{R}^n \rightarrow [0,1]$ is the membership function of $A$, and $P(A)$ represents the belief of a fuzzy event $A$.

\subsubsection{\bf Causes and Types of Uncertainty}  Uncertainty in fuzzy logic mostly comes from {\em linguistic imprecision} or {\em vagueness}, leading to generating unpredictability, multiple knowledge frames, and/or incomplete knowledge.

\subsubsection{\bf Uncertainty Quantification}  \citet{zadeh1968probability} defined two types of fuzzy sets: Type-1 fuzzy set and Type-2 fuzzy set. In Type-1 fuzzy sets, the uncertainty of fuzzy events introduces unpredictability and multiple knowledge frames. Zadeh formulated the uncertainty of a fuzzy event $A$ based on the entropy of event $A$, $H^P(A)$, which is given by:
\begin{equation}
H^P(A) = -\sum_{i=1}^n \mu_A(x_i) P(x_i) \log P(x_i),
\end{equation}
where $A=\{x_1,x_2,\dots,x_n\}$, $\mu_A$ is the membership function of $A$, and $P=\{P(x_1), P(x_2), \dots, P(x_n)\}$. Here, $P(x_i)$ refers to the probability of occurring event $x_i$.

Fuzzy logic research mainly focused on investigating uncertainty measures on Type-2 fuzzy sets, which can provide a way of accurately and effectively measuring fuzziness and uncertainty characteristics of fuzzy complex systems with two membership functions~\cite{zadeh1975concept}.  \citet{wu2007uncertainty} proposed five novel uncertainty metrics, called {\em centroid, cardinality, fuzziness (entropy), variance}, and {\em skewness}, to measure uncertainty in {\em interval} Type-2 fuzzy sets. They further evaluated these metrics with inter-uncertainty and intra-uncertainty raised in words paradigms~\cite{wu2009comparative}. 
\citet{zhai2011uncertainty} extended the five metrics to {\em general Type-2 fuzzy sets}.

\subsubsection{\bf Applications of Fuzzy Logic on  Machine/Deep Learning}

\begin{wrapfigure}{r}{0.5\textwidth}
\centering
\includegraphics[width=0.5\textwidth]{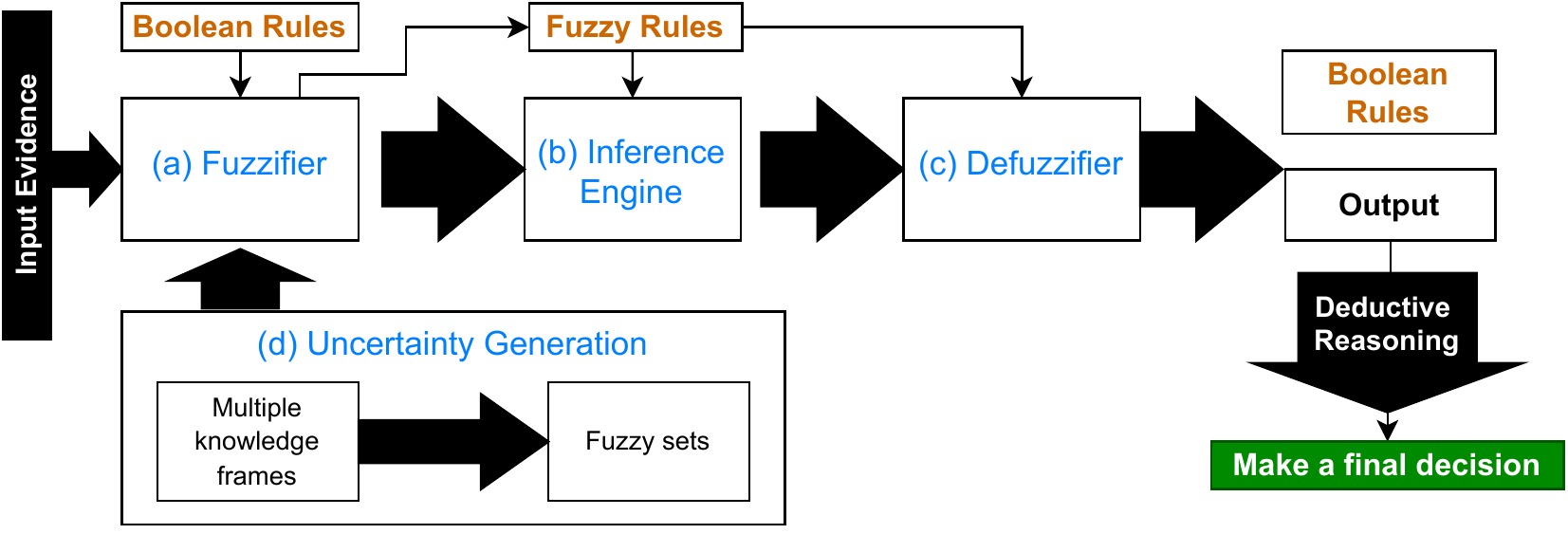}
\vspace{-5mm}
\caption{Uncertainty-aware decision making process in Fuzzy Logic.}
\label{fig:DM-Fuzzy}
\vspace{-5mm}
\end{wrapfigure}

Recently, {\em fuzzy deep neural networks} (FDNNs) are considered for a system using both fuzzy logic and deep neural networks (DNNs) to deal with uncertainty or ambiguity in data~\cite{das2020survey}. The methods using FDNNs fall into two categories: {\em integrated models} and {\em ensemble models}. The integrated models integrate fuzzy logic as a part of DL models. In particular, Pythagorean Fuzzy Deep Boltzmann Machine (PFDBM)~\cite{zheng2017pythagorean} was developed based upon the DBM~\cite{holyoak1987parallel}. PFDBM used the Pythagorean Fuzzy Set (PFS)~\cite{yager2013pythagorean} to replace standard real-valued parameters. \citet{el2018fuzzy} developed a DL model in which a network architecture was designed based on stacked-auto-encoders (SAE) where multiple hyper parameters, such as the learning rate and the momentum, were determined using fuzzy logic systems. 
The ensemble models refer to ensembles of DL and fuzzy logic systems with three models: models with fuzzy inputs, models with fuzzy outputs, and parallel models. \citet{wang2016damaged} proposed a DL model that takes fuzzy feature points for input for damaged fingerprint classification. \citet{zhang2014model} proposed a model that uses DL with fuzzy granulation features to predict time-series data. \citet{chopade2017hybrid} proposed an ensemble of fuzzy logic and DL to predict fuzzy memberships for document summarization.  \citet{deng2016hierarchical} proposed a DL architecture with DL layers and fuzzy membership functions running in parallel.  FDNNs have been applied in various application domains, such as traffic control~\cite{chen2018novel,hernandez2018unsupervised}, surveillance and security~\cite{chen2015fuzzy,zheng2016airline}, text processing~\cite{shirwandkar2018extractive,nguyen2018fuzzy}, image processing~\cite{ahmed2018surveillance}, and time-series prediction~\cite{luo2019evolving}. 

\subsection{Bayesian Inference (BI)} \label{subsec:bayesian-inference}

Bayesian theory has been evolved for more than a hundred years~\cite{Fienberg06}.  Bayesian inference (BI) is the process of inductive learning using Bayer's rule~\cite{hoff2009first}.  Inductive learning is the process of estimating characteristics of a population from a subset of members of the entire population~\cite{hoff2009first}. Although some literature treats BI as an ML technique due to its statistical nature~\cite{tipping2003bayesian}, we treat BI as a belief model because it deals with a subjective probability representing a belief~\cite{hoff2009first}.

\subsubsection{\bf Belief Formation} \label{subsubsec:bayesian-belief}  
\begin{wrapfigure}{r}{0.5\textwidth}
\vspace{-3mm}
\centering
\includegraphics[width=0.5\textwidth]{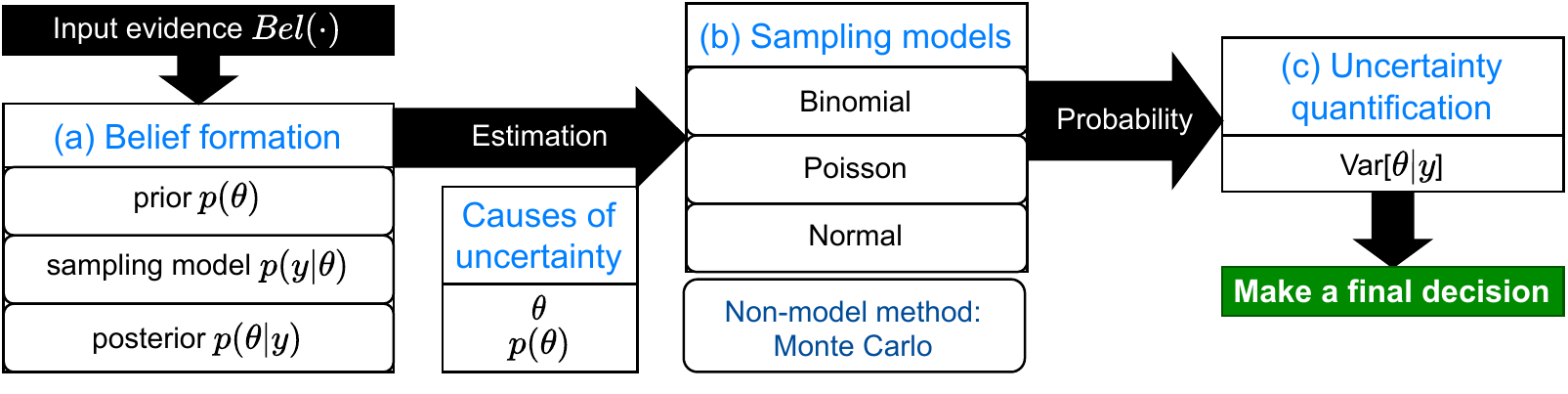}
\vspace{-5mm}
\caption{Uncertainty-aware decision making process using Bayesian inference where  $p(\theta)$ refers to the probability estimation of $\theta$ that causes uncertainty. $Bel(\cdot)$ is the belief of evidence in Eq.~\eqref{Eq: belief_and_plausibility}. When no model is identified, Monte Carlo approximation can be used.} 
\label{fig:DM-Bay}
\vspace{-5mm}
\end{wrapfigure}

Bayes' rule offers a rational tool of updating beliefs of unknown information, which connects probabilities and information~\cite{hoff2009first}.  Beliefs are the statements that can have overlapping domains, such as two beliefs $A$ and $B$ and $A \cap B \neq \emptyset$.  A higher value returned from a belief function indicates the higher degree of a given belief.  Bayesian inference estimates population characteristics $\theta$ from a single dataset sample $y$.  A belief is formed via three steps: 
\begin{enumerate}[leftmargin=*]
\item Prior distribution $p(\theta)$ describes that the belief of $\theta$ being true population characteristics.
\item Sampling model $p(y|\theta)$ shows the belief where $y$ means a sample of the huge sample space $\mathcal{Y}$ if $\theta$ is true and $y$ needs to be estimated.
\item Posterior distribution $p(\theta|y)$ updates the belief about $\theta$ from Bayes' rule based on observed datasets $y$~\cite{hoff2009first}, for the set of all possible parameter values in the parameter space, $\Theta$:
\begin{equation}
p(\theta|y) = \frac{p(y|\theta)p(\theta)}{\int_\Theta p(y|\tilde{\theta})p(\tilde{\theta}) d\tilde{\theta}}.
\end{equation}
\end{enumerate}
Bayesian inference includes conjugate (i.e., prior and posterior distribution are in the same class) prior distributions, posterior inference, predictive distributions, and confidence regions.  There are variants of quantifying the uncertainty of variables depending on different sampling methods. Due to the space constraint, we describe them in Appendix C of the supplement document. 
The probability of an event can be obtained via the following steps~\cite{hoff2009first}: (1) determine proper parameter $\theta$ and sample spaces; (2) select sampling model $p(y|\theta)$ and collect samples; (3) observe prior distribution $p(\theta)$ by experience or select uninformative prior; (4) calculate posterior distribution $p(\theta|y)$ based on prior and sampling methods; (5) perform sensitivity analysis for a range of parameter values; and (6) finalize general estimation of a population mean. The reliable estimate of $\theta$ contains a best guess and degree of its confidence.  We demonstrate the diagram of uncertainty and belief process of Bayesian inference in Fig.~\ref{fig:DM-Bay}.\\
\vspace{-5mm}
\subsubsection{\bf Causes and Types of Uncertainty}  A belief is formed with the unknown values of random variables.  In a population, the parameter of population characteristics $\theta$ may be unknown. This means that the conjugate prior belief $p(\theta)$ is unknown.  Before obtaining a dataset $y$, the subset of a population is also unknown. A sample of dataset $y$ can help to reduce the uncertainty about the population characteristics. This type of uncertainty is caused by a lack of evidence.

\subsubsection{\bf Uncertainty Quantification} In single-parameter sampling models, such as Binomial, Poisson, and Monte Carlo approximation, the posterior inference variance of the estimated mean $\theta$ is a measure of uncertainty from the current belief formation.  That is, the uncertainty is measured by a variance in Binomial model, Poisson model, and Monte Carlo sampling by:
\begin{eqnarray}
\mathrm{Var}^{Bin}[\theta|y]=\frac{\mathrm{E}[\theta|y]\mathrm{E}[1-\theta|y]}{a+b+n+1}, \;  
\mathrm{Var}^{Poiss}[\theta|y]=\frac{a+y}{b+n}, \;
\mathrm{Var}^{MC}[\theta|y]=\sum^S_{s=1}\frac{\theta^{(s)}-\overline{\theta})^2}{(S-1)},
\end{eqnarray}
where $n$ is the number of choices of $y$, $a$ and $b$ are the parameters in $\mathrm{Beta}(a,b)$ distribution for a Binomial model, $a$ and $b$ are the parameters of $\Gamma(a, b)$ distribution for a Poisson model, and $\theta$ is the estimation of parameters and $\overline{\theta}$ is the mean of $\theta$ for Monte Carlo sampling. 

In the normal model with mean $\theta$ and variance $\sigma^2$, a joint distribution can be transformed to a conditional probability by Eq.~(30) in the supplement document. 
The distribution $p(\theta|\sigma^2, y_1, \ldots,y_n)$ is defined by Eq.~(29) in the supplement document 
with variance $\tau_n^2 = 1/(\frac{1}{\tau_0^2}+\frac{n}{\sigma^2})$.  The posterior inverse variance $\frac{1}{\tau_n^2} = \frac{1}{\tau_0^2}+\frac{n}{\sigma^2}$ indicates that the posterior inverse variance (a.k.a. precision) $1/\tau_n^2$ combines sampling precision  $1/\sigma^2$ and prior precision $1/\tau_0^2$. 

\subsubsection{\bf Applications of Bayesian Inference on  Machine/Deep Learning}

\citet{tipping2003bayesian} introduced how Bayesian inference (BI) is used in ML. BI solves a non-deterministic relationship between dependent ($Y$) and independent ($X$) variables.  Given $N$ data examples and many parameters $\textbf{w}$, the model of probability of $Y$ given $X$ is computed by $P(Y|X) = f(X; \textbf{w})$~\cite{tipping2003bayesian}.  The distribution over parameters $\textbf{w}$ can be inferred from Bayes' rule.  Approximation techniques are the key points, such as least-square, maximum likelihood, and regularization.  The common choice of a prior is a zero-mean Gaussian prior.  The Bayesian way of estimating Maximum A Posteriori (MAP) is for posterior inference.  Marginalization serves an important role in the Bayesian framework~\cite{tipping2003bayesian}.  \citet{sofman2006improving} used improved robot navigation in a linear Gaussian model to estimate the posterior distribution of the general Bayesian features and the locale-specific features.  
\citet{tripathi2007selection} used relevance vector machines to predict uncertainty in hydrology.  
\citet{tian2011hybrid} analyzed brain image segmentation by applying Gaussian mixture model (GMM) with a genetic algorithm (GA) and variational expectation-maximization algorithm. 

In the ML/DL, the amount of parameters are expanded in Bayesian neural network (BNN)~\cite{wang2020bayesian-deep}.  BNN is naturally to capture the uncertainty for prediction by putting a prior distribution over its weights, such as Gaussian prior distribution: $\bm{\theta}\sim \mathcal{N} (0, I)$, where $\bm{\theta}$ is the model weights (parameters). Specifically, given a dataset $D=\{X=\{x_1, \ldots, x_N\},Y=\{y_1, \ldots, y_N\}\}$, instead of optimizing the deterministic model weights via maximum likelihood estimation (MLE), BNN refers to extending standard networks with posterior inference, which learns a posterior over model weights $p(\bm{\theta}|D)$ such that model output $f(x,\bm{\theta})$ is stochastic.

\subsection{Subjective Logic (SL)} \label{subsec:SL}

As a variant of DST, \citet{Josang16} proposed a belief model, called \emph{Subjective Logic} (SL) that describes subjectivity of an opinion in terms of multiple belief masses and uncertainty.  

\subsubsection{\bf Belief Formation}
Since a binomial opinion is a special case of multinomial opinions where the number of belief masses is two, for brevity, we only provide the descriptions of multinomial opinions and hyper-opinions.

{\bf Multinomial Opinions}: In SL, a multinomial opinion in a given proposition $x$ is represented by $\omega_X = (\bm{b}_X, u_X, \bm{a}_X)$ where a domain is $\mathbb{X}$, a random variable $X \in \mathbb{X}$, $\kappa = |\mathbb{X}| > 2$ (for a binomial opinion, $\kappa = |\mathbb{X}| = 2$), and the additivity requirement of $\omega_x$ is given as $\sum_{x \in \mathbb{X}} \bm{b}_X(x) + u_X = 1$ where each parameter refers to: (1) $\bm{b}_X$: {\em belief mass distribution} over $\mathbb{X}$; (2) $u_X$: {\em uncertainty mass} representing {\em vacuity of evidence}; and (3) $\bm{a}_X$: {\em base rate distribution} over $\mathbb{X}$.


The projected probability distribution of multinomial opinions is given by:
\begin{equation} \label{eq:multinomial-projected}
\mathbf{P}_X(x) = \bm{b}_X(x) + \bm{a}_X(x) u_X,\;\;\; \forall x \in \mathbb{X}.
\end{equation}  
The probability distribution of a multinomial opinion follows Dirichlet distribution~\cite{Josang16}.

{\bf Hyper-opinions}: Hyper-opinions represent multiple choices under a specific singleton belief $x$ where belief mass is allowed to be assigned to a composite value $x \in \mathscr{C}(\mathbb{X})$ consisting of a set of singleton values $x$'s. Belief masses assigned to composite values $x \in \mathscr{C}(\mathbb{X})$ can be used to estimate the vagueness of an opinion. {\em Hyperdomain}, denoted by $\mathscr{R}(\mathbb{X})$, is the reduced powerset of $\mathbb{X}$ which is the set of $\mathscr{P}(\mathbb{X})$ that excludes $\{\mathbb{X}\}$ and $\{\emptyset\}$. Hyperdomain can be defined by:
\begin{equation} \label{eq:hyperdomain}
\text{Hyperdomain}: \mathscr{R}(\mathbb{X}) = \mathscr{P}(\mathbb{X}) \backslash \{\{\mathbb{X}\}, \{\emptyset\}\}.
\vspace{-1mm}
\end{equation}
Given $X$ as a hyper variable in $\mathscr{R}(\mathbb{X})$, a hyper-opinion on $X$ is represented by $\omega_X = (\bm{b}_X, u_X, \bm{a}_X)$ where each opinion dimension includes: (1) $\bm{b}_X$: {\em belief mass distribution} over $\mathscr{R}(\mathbb{X})$; (2) $u_X$: {\em uncertainty mass} representing {\em vacuity of evidence}; and (3) $\bm{a}_X$: {\em base rate distribution} over $\mathbb{X}$, where $\sum_{x \in \mathscr{R}(\mathbb{X})} \bm{b}_X (x) + u_X = 1$.

\vspace{1mm}
The projected probability distribution of a hyper-opinion can be given by:
\begin{equation} \label{eq:hyper-projected-prob}
\small
\mathbf{P}_X(x) = \sum_{x_i \in \mathscr{R}(\mathbb{X})} \bm{a}_X (x| x_i)\bm{b}_X(x_i) + \bm{a}_X(x) u_X, \; \; \bm{a}_X (x| x_i) = \frac{\bm{a}_X (x \cap x_i)}{\bm{a}_X (x_i)}, 
\forall x, x_i \in \mathscr{R}(\mathbb{X}), 
\end{equation}  
where $\bm{a}_X (x| x_j)$ is the relative base rate and $\bm{a}_X (x_i) \neq 0$. For the binomial or multinomial opinions, the additivity requirement is met (i.e., $\sum_{x \in \mathbb{X}} \mathbf{P}_X (x) = 1$). However, for the hyper-opinion, the additivity requirement may not be met, but $\mathbf{P}_X (x)$ follows super-additivity (i.e., $\sum_{x \in \mathscr{R}(\mathbb{X})} \mathbf{P}_X (x) \geq 1$) with a hyperdomain, $\mathscr{R}(\mathbb{X})$.

Hyper-opinions can be represented by Dirichlet PDFs and the hyper-Dirichlet distribution~\cite{Hankin10}. To do so, we can project a hyper-opinion into a multinomial opinion based on~\cite{Josang16}.  The approximation by the projection of hyper-opinions to multinomial opinions removes vague information in the representation of opinions. This allows a decision maker  to see a particular opinion without the veil of vagueness, which facilitates a more direct and intuitive interpretation of the opinion.

\subsubsection{\bf Causes and Types of Uncertainty} 

In SL, three types of uncertainties are discussed as follows~\cite{Josang18-fusion}: {\em vacuity}, {\em vagueness}, and {\em dissonance}.  {\em Vacuity} uncertainty is caused by a lack of evidence or knowledge. {\em Vagueness} uncertainty is caused by vague observations, leading to failing in identifying a distinctive singleton belief. {\em Dissonance} uncertainty is introduced due to conflicting evidence, resulting in inconclusiveness.  Vacuity and dissonance can be understood as epistemic uncertainty, which can be reduced with more evidence. Vagueness is related to fuzziness, which triggers aleatoric uncertainty in its nature. 

\subsubsection{\bf Uncertainty Quantification}


Uncertainty measures across all belief masses are calculated based on the sum of uncertainty masses associated with individual belief masses, as discussed above.  They include {\em total vacuity} (same as $u_X$), {\em total vagueness} ( $b^{\mathrm{TV}}_{X}$), and {\em total dissonance} ($\dot{b}_{X}^{\mathrm{Diss}}$):
\begin{eqnarray} \label{eq:total-uncertainty}
u_X = \!\!\!\!
\sum \limits_{x \in \mathscr{R}(\mathbb{X})} \mathbf{u}^F_X (x), \;\;
b^{\mathrm{TV}}_{X} =  \!\!\!\!
\sum \limits_{x \in \mathscr{R}(\mathbb{X})} \!\!\!\! \bm{b}_{X}(x), \;\; \dot{b}_{X}^{\mathrm{Diss}} = \sum\limits_{x_{i}\in \mathbb{X}} \bm{b}^{\mathrm{Diss}}_{X}(x_{i}),
\end{eqnarray}
where $\mathbf{u}^F_X (x)$ refers to a focal uncertainty (vacuity per belief), $\bm{b}_{X}(x)$ is a belief mass supporting $x$, and $\bm{b}^{\mathrm{Diss}}_{X}(x_{i})$ indicates dissonance per singleton belief. All these belief masses are detailed in Appendix D.2 of the supplement document.

The uncertainty associated with each belief is elaborated in Appendix D of the supplement document. In addition,  \citet{Josang16} proposed a technique called {\em uncertainty maximization} to offset the amount of evidence received in the past to consider additional new evidence because additional new evidence does not change the current belief states significantly if uncertainty is very low. We provided the uncertainty maximization formulation in Appendix D of the supplement document. 

\begin{wrapfigure}{r}{0.7\textwidth}
\centering
\vspace{4mm}
\includegraphics[width=0.7\textwidth]{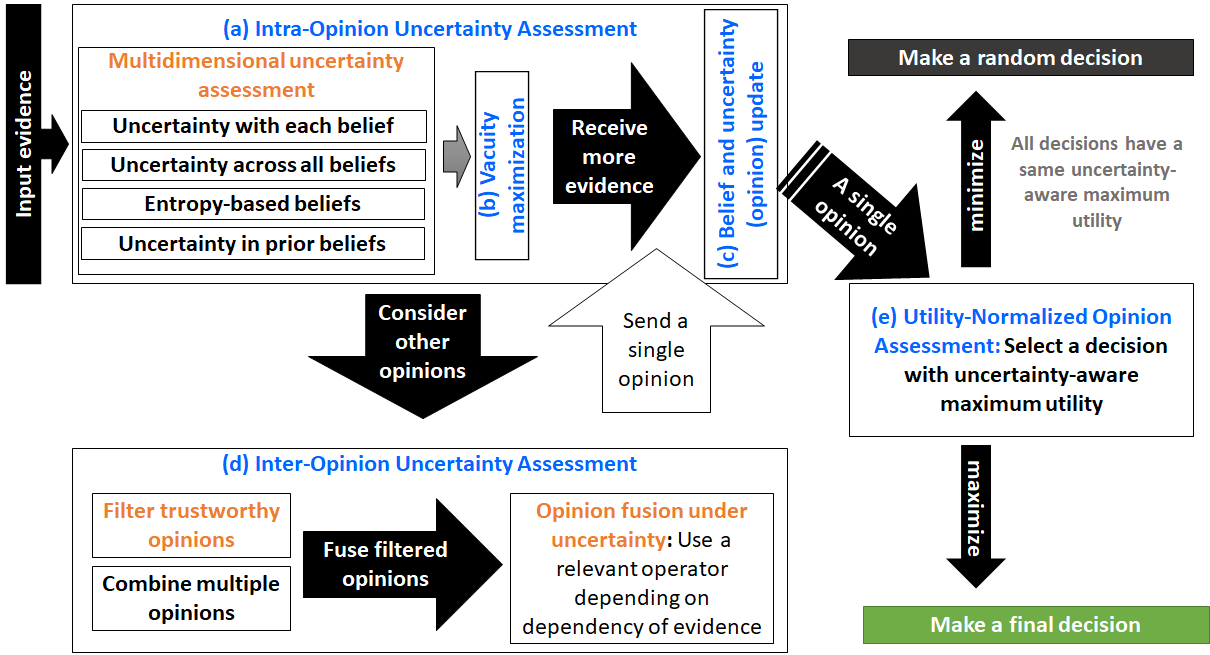}
\vspace{-7mm}
\caption{Uncertainty-aware decision making process using Subjective Logic.}
\label{fig:DM-SL}
\vspace{-9mm}
\end{wrapfigure}

When one makes a decision under uncertainty using SL, we can leverage SL's capability to estimate multidimensional uncertainty (i.e., vagueness, vacuity, and dissonance) to make effective decisions.  As in Fig.~\ref{fig:DM-SL}, after estimating multiple dimensions of uncertainty, one can use the vacuity maximization technique if more evidence is needed in order to allow considering more evidence even under low vacuity, representing high certainty due to a large volume of evidence collected. Recall that SL-based opinion cannot be updated or is rarely updated significantly if its vacuity is or close to zero. One can also consider other opinions by using a variety of fusion operators in SL~\cite{Josang16}, which can generate a single opinion with the updates of corresponding belief masses and vacuity values. The generated single opinion can be also assessed based on which decision has the most utility by normalizing the opinion based on each decision (i.e., belief mass)'s utility.  Most decision making problems can be solved by these processes and allows us to make a decision with minimum uncertainty and maximum utility. However, if all decisions have the same uncertainty-aware maximum utility, one can select a decision at random, which we want to avoid.
\vspace{-7mm}
\subsubsection{\bf Applications of SL on  Machine/Deep Learning}
Recently SL has been considered along with machine/deep learning. Uncertainty reasoning to solve classification tasks has been studied by leveraging SL to consider vacuity and dissonance uncertainty dimensions~\cite{sensoy2018evidential, zhao2019uncertainty}. In addition, SL-based opinion formulation is used to infer subjective opinions along with DL in the presence of adversarial attacks~\cite{Alim19, zhao2018icdm, zhao2018bigdata}. Further, SL-based opinions are considered along with deep reinforcement learning to propose uncertainty-aware decision making~\cite{Zhao19-fusion}. 

\section{Belief Theory Meets Deep Learning} \label{sec:belief-theory-meets-DL}
In this section, we review several hybrid frameworks that combine belief models and neural networks, including evidential (or subjective) neural networks,  fuzzy neural networks, and rough deep neural networks. 

\subsection{Evidential Neural Networks (ENNs)} \label{subsec:enns}

Evidential neural networks (ENNs)~\cite{sensoy2018evidential} is a hybrid framework of subjective belief models and neural networks. They are similar to classic neural networks for classification. The main difference is that the softmax layer is replaced with an activation function in ENNs, e.g., ReLU, to ensure non-negative output (i.e., range of $[0, +\infty]$), which is taken as the evidence vector for the predicted Dirichlet distribution, or equivalently, multinomial opinion. 

\subsubsection{\bf Key Formulation of ENNs}
Given the feature vector ${\bf x}$ of an input sample, let $f({\bf x} | {\bm \theta})$ represent the evidence vector by the network for the classification, where ${\bm \theta}$ is network parameters. Then the corresponding Dirichlet distribution has parameters ${\bm \alpha} = f({\bf x}_i | {\bm \theta}) + 1$, where the $k$-th parameter ${\alpha}_{k}$ denotes the effective number of observations of the $k$-th class, and the total number of classes is $K$.  Let $\p=(p_1, \dots, p_K)^T$ be the probabilities of the $K$ predefined classes. The Dirichlet PDF (i.e., $\text{Dir}({\bf p}; {\bm \alpha})$)
with ${\bf p}$ as a random vector is defined by:
\begin{eqnarray} \label{eq:multinomial-dir}
\mathrm{Dir}(\bm{p}| {\bm \alpha}) = \frac{1}{B({\bm \alpha})} \prod\nolimits_{k\in \mathbb{Y}} p_k ^{(\alpha_k-1)},
\end{eqnarray} 
where $\frac{1}{B({\bm \alpha})} = \frac{\Gamma (\sum_{k \in \mathbb{Y}} \alpha_k)}{\prod_{k \in \mathbb{Y}} (\alpha_k)}$, $\alpha_k \geq 0$, and $p_k \neq 0$, if $\alpha_k < 1$. The expected value of class probabilities $\p=(p_1, \ldots, p_K)^T$ is given by: 
\begin{align}
\mathbb{E}[p_k]=\frac{\alpha_k}{\sum_{j=1}^K \alpha_j}=\frac{e_k+a_kW}{\sum_{j=1}^K e_j + W}. 
\end{align}
The observed evidence in a Dirichlet PDF $\mathrm{Dir}(\bm{p}| {\bm \alpha})$ can be mapped to a multinomial opinion $(b_1, \cdots, b_K, u)$ as follows:
\begin{align}
\label{eq:dirichlet-opinion}
    b_k= \frac{e_k}{S}, \; \; u= \frac{W}{S}, \text{ for }  k=1, \cdots,  K, 
\end{align}
where $S = \sum_{k=1}^K \alpha_k$ refers to the Dirichlet strength. Without loss of generality, we set $a_k = 1/K$
and the non-informative prior weight (i.e., $W = K$), which indicates that $a_k \cdot W = 1$ for each $k \in \{1, \cdots, K\}$. Therefore, the output of an ENN can be applied to measure the subjective uncertainty about the predictive class variable $y$ in different types, such as vacuity and dissonance as defined based on a multinomial opinion (See Section~\ref{subsec:SL}). 

The Bayesian framework of ENNs was proposed in~\cite{malinin2018predictive}that considers a prior distribution on the network parameters ${\bm \theta}$, denoted by $P({\bm \theta})$. Let $P({\bm \theta} | \mathcal{D})$ be the posterier PDF, where $ \mathcal{D}$ refers to the training set. Let $\text{Cat}(y | {\bf p})$ be the PDF of the categorical distribution about the predictive variable $y$, where the class probabilities ${\bf p}$ are the parameters. We can then show terms associated with different uncertainty as follows:
\begin{eqnarray}
    P(y | {\bf x}, \mathcal{D} ) = \int \int \underbrace{\text{Cat}(y | {\bf p})}_{Data \;\; uncertainty} \;\;
     \underbrace{P({\bf p} | {\bf x}, {\bm \theta})}_{Subjective \;\; uncertainty} \;\; \underbrace{P({\bm \theta} | \mathcal{D})}_{Model \;\; uncertainty} d {\bm p}  d {\bm \theta},
\end{eqnarray}
where $P({\bf p} | {\bf x}, {\bm \theta}) = \text{Dir}({\bf p} | {\bm \alpha})$ and ${\bm \alpha} = f({\bf x}, {\bm \theta})$. 
In this expression, data (aleatoric), subjective (distributional), and model (epistemic) uncertainty are modeled by a separate term within an interpretable probabilistic framework. The data uncertainty is described by the point-estimate categorical distribution, $\text{Cat}(y | {\bf p})$. The subjective (or distributional) uncertainty is described by the distribution over predictive class variables $P({\bf p} | {\bf x}, {\bm \theta})$.  The model uncertainty is described by the posterior distribution over the parameters given the data. The relationship between uncertainties is made explicit - model uncertainty affects estimates of subjective uncertainty, which in turn affects the estimates of data uncertainty. This forms a hierarchical model with three layers of uncertainty: the posterior over classes, the per-data Dirichlet prior distribution, and the global posterior distribution over model parameters. 
The uncertainty due to the mismatch between testing and training distributions can be measured by two methods. First, as the Dirichlet distribution $P({\bf p}|{\bf x}, {\bm \theta})$ is equivalent to a subjective multinomial opinion based on the mapping defined in Eq.~\eqref{eq:dirichlet-opinion}, we can quantify subjective uncertainty types directly based on the Dirichlet distribution, such as vacuity and dissonance, where vacuity captures elements of distributional uncertainty.  Second, the distributional uncertainty can be measured based on mutual information between the categorical label $y$ and the class probabilities ${\bf p}$ as:
\begin{eqnarray}
\underbrace{\mathcal{I}[y, {\bf p}| {\bf x}, \mathcal{D})]}_{Epistemic\; \; uncertainty} = \underbrace{\mathcal{H}[\mathbb{E}_{P({\bf p} | {\bf x}; \mathcal{D})} [\text{Cat}(y | {\bf p})]]}_{Entropy} - \underbrace{\mathbb{E}_{P({\bf p} | {\bf x}; \mathcal{D})} [\mathcal{H}[\text{Cat}(y | {\bf p})]]}_{Aleatoric\; \; uncertainty}
\end{eqnarray}
We note that distributional uncertainty and vacuity are negatively correlated, if the parameters ${\bm \theta}$ are deterministic.  The former is maximized (and the latter is minimized) when all categorical distributions are equiprobable, which occurs when the Dirichlet distribution is flat.

\subsubsection{\bf Causes and Types of Uncertainty}
Since ENNs provide a hybrid framework of subjective belief models and neural networks, we can estimate evidential uncertainty, such as {\em vacuity (scenario uncertainty)} and {\em dissonance (discord uncertainty)}, based on a subjective opinion. Recall that vacuity is due to a lack of evidence introducing uncertainty by incomplete knowledge. Dissonance is due to conflicting evidence, resulting in multiple knowledge frames.

\subsubsection{\bf Uncertainty Quantification}
ENNs estimate Dirichlet distribution parameters directly, which can be transferred to a subjective opinion. After then, we can estimate vacuity ($u$) and dissonance ($diss$) based on SL-based subjective opinion where there are $K$ classes and $e_k$ number of evidence to support each class $k$ by:
\begin{align}
\small
u = \frac{K}{\sum_{k=1}^K e_k + K}, \;
diss=\sum_{k=1}^{K}\Bigg(\frac{b_k \sum_{j\neq k}b_j \text{Bal}(b_j,b_k)}{\sum_{j\neq k}b_j}\Bigg), \\
\small{\text{Bal}(b_j,b_k)=
  \begin{cases} 
  1-\frac{|b_j-b_k|}{b_j+b_k} & \text{if $b_i b_j \neq 0$}\\
  0 & \text{if $\min(b_i,b_j)=0$}
  \end{cases}}
\end{align}
where $b_k$ and $b_j$ refer to the belief masses supporting $k$ class and $j$ class, respectively.

\subsubsection{\bf Applications of ENNs}

There is a whole class of evidential neural networks that have the interpretation that evidence represents the number of nearby training samples of various classes relative to the sample under test. This includes the generative version from~\cite{sensoy2020uncertainty}, posterior networks based on density-based pseudo-counts in~\cite{charpentier2020posterior}, and  Epistemic Neural networks~\cite{osband2021epistemic} that allow a general interface to distinguish epistemic from aleatoric uncertainty. 
The ENNs have been applied on several applications in different domains, such as justified true belief models~\cite{virani2020justification,bhushan2020variational},  active learning on image data~\cite{weishi20}, misclassification and out-of-distribution detection on graph data~\cite{zhao-Neurips-2020,hu2020multidimensional}, event early detection on time series data~\cite{zhao2022seed}, and self-training on NLP task~\cite{xu2021boosting}.

\subsection{Fuzzy Deep Neural Networks (FDNNs)}
Fuzzy deep neural networks (FDNNs) is a hybrid framework of fuzzy logic systems and deep neural networks~\cite{das2020survey}. FDNNs is designed to address the drawback that deep neural networks are sensitive to the uncertainties and the ambiguities of real-world data. Multiple approaches are developed to implement an FDNN. Some models, such as Fuzzy Restricted Boltzmann Machines (FRBMs)~\cite{chen2015fuzzy}, 
consider the concept of fuzzy numbers to represent network
weights. Some models use fuzzy logic units to replace perceptrons in the network~\cite{park2016intra}. Fuzzy systems are also used to train the network parameters of a deep neural network~\cite{el2018fuzzy}. In this section, we use Pythagorean Fuzzy Deep Boltzmann Machines (PFDBMs)~\cite{zheng2016airline}, a recent extension of FRBMs,  to demonstrate how fuzzy logic can be integrated as a part of deep neural networks, such as deep Boltzmann machines. 




\vspace{1mm}

\subsubsection{\bf Key Formulation of PFRBMs} 
\label{subsec:fdnn-formulation}
We will first introduce the building blocks, including deep Boltzmann machines and pythagorean fuzzy set, and then introduce the architecture design of PFRBMs. 
A deep Boltzmann machine (DBM) is an extension of the restricted Boltzmann machine~\cite{zhang2018overview} and considers multiple hidden layers to capture more complex correlations of the activities of the preceding layers~\cite{salakhutdinov2010efficient}. Considering a $L$ hidden layers DBM whose set of layers is $\{\x, \h_1, \ldots, \h_L\}$ where $x$ is a set of visible units $\x\in \{0, 1\}^D$, and  $\h_l$ is $l$-th hidden layerwith a set of hidden units $\h_l\in \{0, 1\}^{P_l}$. DBM is an energy-based probabilistic model which defines a joint probability distribution over $x$ as
\begin{equation}
P(\x; \theta)=\frac{1}{Z(\theta)} \sum_{\h_1} \cdots \sum_{\h_L} e^{-E\left(\x, \h_{1},\ldots, \h_{L}, \theta\right)} \label{objective}
\end{equation}
where $\theta=[W_1, \ldots, W_L]$ is a vector of the parameters, and $E(\x, \h_1,\ldots, \h_{L}; \theta)$ is the energy function~\cite{smolensky1986information} n defined as
\begin{equation}
E(\x, \h_1, \ldots, \h_{L}; \theta) = -\x^T W_1 \h_1 -\sum_{l=2}^L \h_{l-1}^T W_l \h_l,\label{energy function}
\end{equation}
and $Z(\theta)$ is the partition function defined as
\begin{equation}
    Z(\theta) = \sum_{\x}  \sum_{\h_1} \cdots \sum_{\h_L} e^{-E\left(\x, \h_{1},\ldots, \h_{L}, \theta\right)} 
\end{equation}
DBM aims to maximize the joint probability $P(\x; \theta)$, which has the same effect to minimize energy function $E(\cdot)$.  

Pythagorean fuzzy sets (PFS) is an extension of the basic fuzzy sets in two perspectives. First, it introduces a non-membership degree besides the standard membership degree []. Second, it considers the restriction that the sum of the squares of the membership degree is between 0 and 1. PFS is defined by the mathematical object: \begin{equation}
\mathcal{P}=\left\{\langle x,\mu_{p}(x), v_{p}(x) \rangle \mid x \in S\right\},
\end{equation}
where $\mu_p(x):S\rightarrow [0, 1]$ is the membership degree (how much degree of $x\in S$) of element $x$ to $S$ in $P$, and $\nu_p(x):S\rightarrow [0, 1]$ is the non-membership degree (how much degree of $x\notin S$) as well. In addition, we have $\mu^2_p(x)+ \nu^2_p(x)\le 1$, and the hesitant degree, neither membership nor non-membership
degree, may consider as uncertainty degree. The hesitation degree (uncertainty degree) is the function that expresses lack of knowledge of whether $x \in S$ or $x \notin S$. It can be calculated by: \begin{equation}
\pi_{p}(x)=\sqrt{1-\mu_{p}^{2}(x)-v_{p}^{2}(x)}.
\end{equation}
Moreover, to simplify it, $\mathcal{P}(\mu_p(x), \nu_p(x))$ is called a Pythagorean fuzzy number (PFN) denoted $\beta=\mathcal{P}(\mu_\beta, \nu_\beta)$, where $\mu_\beta, \nu_\beta \in [0, 1]$ and $\mu_\beta^2+\nu_\beta^2 \le 1$.  We can use two metrics to rank a PFN by:
\begin{equation}
\begin{aligned}
&h(\beta)=\mu_{\beta}^{2}+v_{\beta}^{2}, &s(\beta)=\mu_{\beta}^{2}-v_{\beta}^{2},
\end{aligned}
\end{equation}
where $h(\beta)$ is the accuracy function of $\beta$ and $s(\beta)$ is the score function of $\beta$. The ranking of two PFNs, $\beta_1=P(\mu_{\beta_1}, \nu_{\beta_1})$ and $\beta_2=P(\mu_{\beta_2}, \nu_{\beta_2})$, is performed by: (1) If $s\left(\beta_{1}\right)<s\left(\beta_{2}\right)$, then $\beta_{1}<\beta_{2}$; and
(2) If $s\left(\beta_{1}\right)=s\left(\beta_{2}\right)$, then (a) if $h\left(\beta_{1}\right)<h\left(\beta_{2}\right)$, then $\beta_{1}<\beta_{2}$; and
(b) if $h\left(\beta_{1}\right)=h\left(\beta_{2}\right)$, then $\beta_{1}=\beta_{2}$.

\begin{wrapfigure}{R}{0.4\textwidth}
\centering
\vspace{-2mm}
\includegraphics[width=0.37\textwidth]{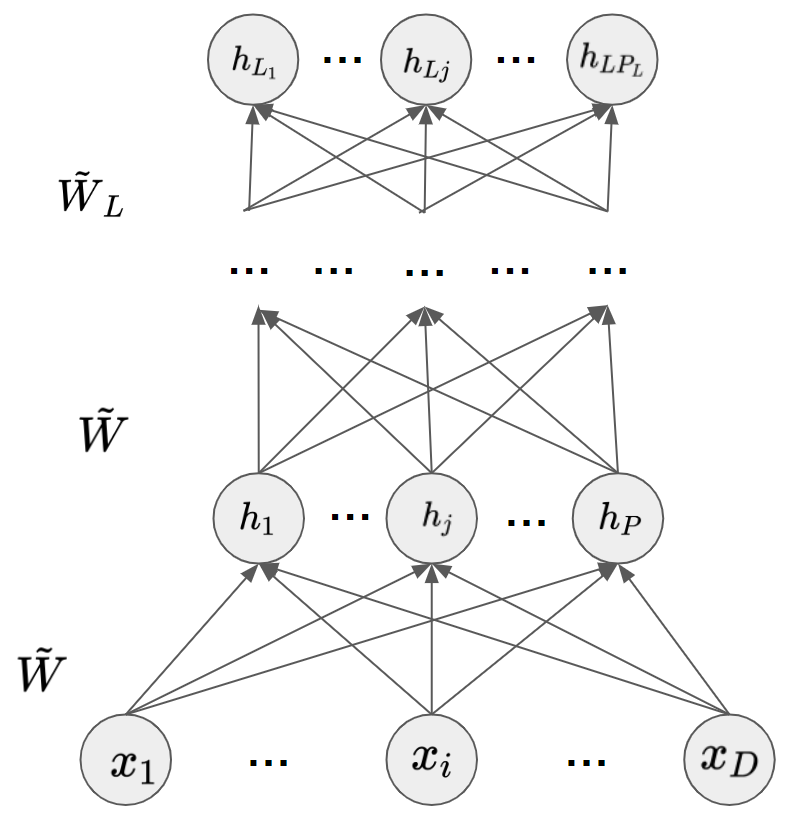}
\vspace{-3mm}
\caption{Pythagorean Fuzzy Deep Belief Network (PFDBN).}
\vspace{-5mm}
\label{fig:FDL2}
\end{wrapfigure}

The Pythagorean Fuzzy Restricted Boltzmann Machine (PFRBM) extends the DBM model by replacing the standard real-valued parameters with PFNs. 
PFRBM is able to handle fuzzy and/or incomplete data and the fuzzy parameters provide a better representation of the data using fuzzy probability. Fig.~\ref{fig:FDL2} describes the framework of a PFRBM with $L$ layers, denoted as $h_1, \ldots, h_L$. Let $\widetilde{\theta}=[\widetilde{W_{1}}, \ldots, \widetilde{W_{L}}]$ be the fuzzy parameters and $\x = (x_1, \cdots, x_D)$ be the input feature vector. The energy function and probability function of a PFRBM  are shown as follows:
\begin{eqnarray}
\widetilde{E}\left(\x, \h_{1}, \ldots, \h_{L}; \widetilde{\theta}\right)=-\x^{T} \widetilde{W_{1}} \h_{1}-\sum_{l=2}^{L} \h_{l-1}^{T} \widetilde{W_{l}} \h_{l}, \\
\widetilde{P}(\x;\bar{\theta})=\frac{1}{\widetilde{Z}(\widetilde{\theta})} \sum_{\h_{1}} \cdots \sum_{\h_{L}} e^{-\widetilde{E}\left(\x, \h_{1}, \ldots, \h_{L}, \widetilde{\theta}\right)}. 
\end{eqnarray}
Therefore, we consider the log likelihood as the objective function, 
\begin{equation}
\max _{\tilde{\theta}} \widetilde{\mathcal{L}}(\widetilde{\theta}, D)=\sum_{\x \in D} \log (\widetilde{P}(\x, \widetilde{\theta})).
\end{equation}
As fuzzy optimization problems are intractable, the PFDBM is trained using a combination of gradient descent and metaheuristic techniques.

\subsubsection{\bf Causes and Types of Uncertainty}
PFDBMs provides a hybrid framework of fuzzy sets and DNNs where uncertainty comes from a fuzzy set due to fuzzy and/or incomplete data, leading to unpredictability. The fuzziness has its root nature in aleatoric uncertainty.

\subsubsection{\bf Uncertainty Quantification}
Unlike traditional DNNs, PFDBMs with fuzzy parameters provide a better representation of data using a fuzzy probability to represent uncertainty. The fuzzy parameters can learn new features and allow investigating how much a certain (or uncertain) feature influences the output. 

\subsubsection{\bf Applications of FDNNs}
PFDBMs was proposed to develop an airline passenger profiling~\cite{zheng2016airline} and provide an early warning system for industrial accidents~\cite{zheng2017pythagorean}. 
Besides PFDBMs,~\citet{park2016intra} developed intra- and inter-fraction FDNNs to tracking lung-cancer tumor motion. Similar to~\cite{el2018fuzzy}, fuzzy logic was employed to train the learning parameters in FDNNs for traffic incident detection. In addition, some models consider fuzzy logic and deep learning in a sequential or parallel fashion. ~\citet{wang2016damaged} proposed a model that uses deep neural network  with fuzzy feature points for damaged fingerprint classification. ~\citet{zhang2014model} proposed a model utilizing fuzzy granulation and deep
belief network for predicting time-series data.

\subsection{Rough Deep Neural Networks (RDNNs)} \label{subsec:rdnn}

Rough neural networks (RNNs) have been studied for a decade by combining a rough set or rough neuron with DNNs to process the uncertainties and high-dimensional data~\cite{lingras1996rough}. The methods fall into two main categories: rough neural-based and rough set-based. Due to the space constraint, we discuss the rough neural-based method while providing the details of the rough set-based approaches in Appendix E.2 of the supplement document.

\begin{wrapfigure}{R}{0.36\textwidth}
\vspace{-3mm}
    \centering
    \includegraphics[width=0.35\textwidth]{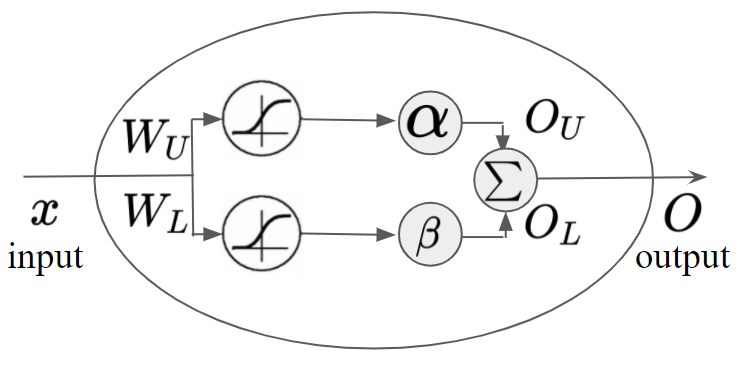}
    \caption{Rough neuron with six tunable parameters}
    \label{fig:rough_fig}
\end{wrapfigure}

\subsubsection{\bf Key Formulation of RDNNs}

Rough neural-based method considers a rough neuron in DNNs to improve the robustness of learning. For a traditional neural network, if the input feature is represented by a range, such as the temperature of climate (e.g., daily maximum and minimum temperature), the neural network cannot learn a good representation and the prediction error will be relatively large. The neural network based on a rough neuron can address this issue. 

Fig.~\ref{fig:rough_fig} shows how the rough neuron is applied for rough pattern recognition. This neuron consists of an upper bound neuron with parameters $\theta_U = \{W_U ,b_U, \alpha\}$, and a lower bound neuron with parameters $\theta_L = \{W_L, b_L, \beta\}$. Here $W_U$ and $b_U$ are the weight and bias of the upper bound, respectively, while $W_L$ and $b_L$ are those for the lower bound neuron, respectively. Output coefficients, $0 \le \alpha$ and $\beta \le 1$, determine the contribution of upper bound output $O_U$ and lower bound output $O_L$ to the overall neuron's output $O$.  A rough extension of auto-encode, called {\em rough auto-encoder} (RAE), uses rough neurons in its hidden layer and output layer. Here $W^k_U$, $b^k_U$, and $\alpha^k$ are the upper bound parameters of layer $k$ and $W^k_L, b^k_L$, and $\beta^k$ are the lower bound parameters of layer $k$, respectively. 

Given the RAE defined with input vector $h_0 = X$, the upper bound and lower bound outputs of the first hidden layer are shown where $W_{U}^{1}$ and $W_{L}^{1}$ are the learned parameters and $f^{1}W_{L}^{1} X+b_{L}^{1}$ can be larger than $f^{1}W_{U}^{1} X+b_{U}^{1}$. The $h_{U}^{1} (X)$ and $h_{L}^{1} (X)$ are defined by:
\begin{equation}
\begin{aligned}
&h_{U}^{1} (X)=\max \Big[f^{1}\left(W_{U}^{1} X+b_{U}^{1}\right), f^{1}\left(W_{L}^{1} X+b_{L}^{1}\right)\Big], \\
&h_{L}^{1} (X)=\min \Big[f^{1}\left(W_{U}^{1} X+b_{U}^{1}\right), f^{1}\left(W_{L}^{1} X+b_{L}^{1}\right)\Big],
\end{aligned}
\end{equation}
where $f^1$ is a {\em sigmoid} function. The latent representation in the hidden layer is computed by:
\begin{equation}
h^1 = \alpha^1 h^1_U +\beta^1 h^1_L.
\end{equation}
For the rough decoding process in the output layer, the upper bound and lower bound outputs are computed as:
\begin{equation}
\begin{aligned}
h_{U}^{2} &=\max \Big[f^{2}\left(W_{U}^{2} h^{1}+b_{U}^{2}\right), f^{2}\left(W_{L}^{2} h^{1}+b_{L}^{2}\right)\Big], \\
h_{L}^{2} &=\min \Big[f^{2}\left(W_{U}^{2} h^{1}+b_{U}^{2}\right), f^{2}\left(W_{L}^{2} h^{1}+b_{L}^{2}\right)\Big],
\end{aligned}
\end{equation}
where $f^2$ is considered to be a linear function. Therefore, we have the reconstructed input, 
\begin{equation}
    r=\alpha^2 h^2_U + \beta^2 h^2_L. \label{reconstructed}
\end{equation}

\subsubsection{\bf Causes and Types of Uncertainty}
In RDNNs, uncertainty is considered in a rough set introducing unpredictability and rough neuron introducing incomplete knowledge. Hence, the rough set and neuron can capture vagueness from model input and parameter uncertainty from model parameters.

\subsubsection{\bf Uncertainty Quantification}

The uncertainty in RDNNs can be estimated based on the rough set theorem introduced in Section~\ref{subsec:3-1-TVL}. The rough set theorem approximates an $M$-boundary region, which contains a set of objects
that cannot be clearly classified by only employing the set of attributes and represents vagueness. 

\subsubsection{\bf Applications of RDNNs}
Most RDNNs are proposed to reduce uncertainty.
\citet{zhang2006fuzzy} applied the fuzzy-rough neural network in vowels recognition. \citet{khodayar2017rough} proposed a rough extension of {\em stacked denoising autoencoder} (SDAE)  for ultrashort-term and short-term wind speed forecasting, incorporating a rough neural network into wind uncertainties. Sinusoidal Rough-Neural Network (SR-NN)~\cite{jahangir2020short} is proposed to predict wind speed by using rough neurons to handle the high intermittent behavior of wind speed.

\section{Summary of the Key Findings} \label{sec:summary-findings}

We summarize the key findings from our survey by answering the key research questions below:
\noindent {\bf RQ1.} {\em What are the key causes and types of uncertainty studied in belief theory and deep learning?}

\vspace{1mm}
\noindent {\bf Answer:} The majority of belief models, such as DST, TBM, IDM, SL, TVL, and Bayesian inference, consider uncertainty caused by a lack of evidence, which is called {\em vacuity} in SL. It is related to aleatoric uncertainty where a long-term probability can increase as more evidence is received. The second most common uncertainty type is considered in belief models, such as DSmT, SL, or Fuzzy Logic, is {\em discord} (or dissonance), caused by disagreement or conflicting evidence from multiple sources or observers, which generates multiple knowledge frames and results in inconclusiveness in decision making. Lastly, unpredictability is introduced by unclearness or impreciseness of observations or beliefs, which are considered as {\em fuzziness} in TVL, {\em vagueness} in SL, and {\em imprecise beliefs} in DSmT. In DL, two types of uncertainty natures are mainly considered: epistemic uncertainty and aleatoric uncertainty. Epistemic uncertainty, also called `model or systematic uncertainty,' represents the model (parameters) uncertainty due to the limited training data.  Aleatoric uncertainty indicates  data uncertainty introduced by the nature of randomness in data. 

\vspace{1mm}
\noindent {\bf RQ2.} {\em How can the ontology of uncertainty be defined based on the multidimensional aspects of uncertainty studied in belief models and deep learning?}

\vspace{1mm}
\noindent {\bf Answer:} We demonstrated the ontology of uncertainty in Fig.~\ref{fig:uncertainty-ontology}. The source of uncertainty can be from machines, networks, environmental factors, and humans that can generate a lot of various types of uncertain data. Uncertainty has a model of reasoning and quantifying various types of uncertainties to make effective decision making. We limited the models in belief theory and DL. Uncertainty has procedures to collect evidence, including both subjective and objective data or information. Uncertainty has its multiple types including ambiguity and fuzziness which also have been studied under different taxonomies (see Fig.~\ref{fig:uncertainty-types}), such as vagueness, imprecision, unclearness, and so forth. Uncertainty has its root nature in the most popularly used two types of uncertainty: aleatoric and epistemic uncertainty.

\vspace{1mm}
\noindent {\bf RQ3.} {\em How has each belief model considered and measured uncertainty?} 

\vspace{1mm}
\noindent {\bf Answer:} DST's combination rule helps the decision by combining beliefs from multiple information channels. IDM provides a belief range, rather than a single value, allowing a decision maker to be aware of the magnitude of uncertainty.  In DSmT, Shannon's entropy and the Probabilistic Information Content (PIC) score are used to indicate uncertainty caused by a lack of evidence where a decision is made based on GPT and DSmP. Bayesian inference theory uses a variance or co-variance to measure uncertainty representing unpredictability. In SL, one can use a projected belief that interprets uncertainty (i.e., vacuity) based on its prior belief (i.e., base rate). If there is very low vacuity but high dissonance, one can maximize vacuity by offsetting the amount of the smallest belief mass while increasing vacuity to have a high effect on new evidence. TVL uses an unknown status to model system uncertainty and defines a set of logical operations to decide the system status for decision making in system operations. Fuzzy Logic uses fuzzy entropy to quantify the unpredictability and multiple knowledge frames of fuzzy events.  

\vspace{1mm}
\noindent {\bf RQ4.} {\em How has each belief model been applied in deep learning and vice-versa for effective decision making under uncertainty?} 

\vspace{1mm}
\noindent {\bf Answer:} TVL is used to solve classification problems in pattern recognition tasks~\cite{kashkevich1979two} and leveraged to establish a database for natural language consultation~\cite{dahl1979quantification} in 1970s. We rarely found any recent work using TVL in ML/DL applications.  DST is mainly used to fuse data from multi-sensors before conducting neural network training, or fuse predictions from two identically trained models~\cite{soua2016big, tong2021evidential, tian2020deep}.  To our knowledge, TBM is also used to solve classification problems but not used with ML/DL.  DSmT is used along with ML/DL to solve classification problems where it is integrated with SVM, CNN, LSTM, and RF~\cite{abbas2015effective, ji2021dsmt}.  IDM is used to improve ML algorithms, such as AdaBoost~\cite{utkin2015imprecise}, Decision Tree~\cite{moral2020imprecise}, or n\"{a}ive classifier~\cite{Corani10-idm-ml}.  Fuzzy Logic is combined with DNNs, named fuzzy DNNs, to deal with ambiguity in data~\cite{das2020survey, holyoak1987parallel, el2018fuzzy, wang2016damaged, zhang2014model}.  Bayesian inference is mainly used to infer the posterior distribution of Bayesian features~\cite{sofman2006improving, tripathi2007selection, tian2011hybrid}.  SL's vacuity and dissonance uncertainty dimensions are considered in evidential neural networks for uncertainty-aware decision making in classification problems~\cite{sensoy2018evidential, zhao2019uncertainty}. Vacuity is used to detect out-of-distribution (OOD) samples while dissonance is used to detect missclassification samples. Rough set theory is combined with DNNs, named RDNNs, to 
deal with imprecise information and uncertainty in data (e.g., ranges as values for input and/or output variables)~\cite{zhang2007integrated, yasdi1995combining, lingras1996rough, khodayar2017rough}. 

\vspace{1mm}
\noindent {\bf RQ5.} {\em What are the key differences of uncertainty reasoning and quantification in belief theory and deep learning?}

\vspace{1mm}
\noindent {\bf Answer:} Deep learning (DL) has received high attention because of its powerful capability to deal with a large volume of high dimensional data and provide solutions to maximize decision performance. However, as DL is limited in dealing with uncertainty explicitly, it often faces the issue of unexplainability, a well-known issue of explainable AI (XAI), due to its nature of statistical inference. On the other hand, belief models provide rigorous mathematical formulation based on a limited number of parameters which can offer the capability to easily reason and quantify different types of uncertainties. This merit of {\em quantifiable uncertainty} in belief models can provide reasons to explain a decision made based on mathematical induction. However, belief models suffer from dealing with a large volume of data, which can be complemented by DL.  Therefore, our work discussed how a belief model (e.g., SL) has been bridged with DL to improve decision making capability based on the merits of both approaches to achieve XAI.


\vspace{1mm}
\noindent {\bf RQ6.} {\em How can belief model(s) be applied in DL to solve complicated decision making problems?}

\vspace{1mm}
\noindent {\bf Answer:} There may be various ways to leverage belief models considered in DL research. One example we discussed in Section~\ref{sec:belief-theory-meets-DL} is combining SL's opinions with DNNs by constructing evidential NNs (ENNs). That is, ENNs can be built to generate evidence to formulate a subjective opinion in SL, rather than using a common activation function, such as softmax, generating class probabilities.  Based on the estimated evidence in ENNs,  we can calculate vacuity and dissonance uncertainty by leveraging the operators in SL.  Depending on the degree of the quantified uncertainty values, such as vacuity, vagueness, or dissonance, diverse algorithms can be developed for effective decision making.  Based on our prior work~\cite{zhao2019uncertainty}, we found vacuity is a promising uncertainty type to detect out-of-distribution (OOD) samples while dissonance is an uncertainty type that can effectively detect misclassification samples. 

\section{Concluding Remarks} \label{sec:concluding-remarks}

\subsection{Insights, Lessons Learned, and Limitations}

\begin{itemize}
\item Recent efforts have been made to estimate different types of uncertainty in our prior work~\cite{Josang18-fusion} while vacuity and vagueness have been mainly considered in the past~\cite{Shafer76, Josang16, zadeh1965fuzzy}. However, reasoning and quantification of other dimensions of uncertainty still remains unaddressed in the literature.  Furthermore, the question of how different types of uncertainty can be helpful for effective decision making has not been addressed in the literature. 
\item When both all the belief masses and a prior belief supporting each belief are the same, decision making becomes more challenging because it leads to inconclusiveness.  Although utility-based belief masses have been studied~\cite{Yang09-belief-utility, Josang16}, their contributions are limited with the theoretical discussions based on simple examples. 

\item Most belief/evidence theories and their uncertainty reasoning show high maturity as they have been explored since the 1960s. However, they are mostly theoretical and have not been thoroughly validated based on real datasets and/or applications for effective decision making and learning.

\item Bayesian theorem and inference methods are the foundations of the advanced machine learning and deep learning algorithms.  However, the fitting of Bayesian inference from an one-parameter model to a deep learning model with a large volume of parameters can introduce non-trivial challenges in quantifying uncertainties.

\item Although belief models have considered various types of uncertainties, as described in Fig.~\ref{fig:uncertainty-types}, the terminologies of those types are often found very similar but their distinctions have not been clarified. Although our survey paper can help readers better understand the diverse types of uncertainties, one may want to argue about our clarification of uncertainty types in Section~\ref{sec:causes-types-uncertainty}. This implies that much more efforts should be made to investigate different types of uncertainties and their effect on diverse decision making settings and applications.

\item Some efforts leveraging both a belief theory and deep learning have been made, such as fuzzy deep neural networks (FDNNs) combining fuzzy logic and DNNs~\cite{das2020survey}, rough deep neural networks (RDNNs) combining rough logic and DNNs~\cite{lingras1996rough}, and evidential neural networks (ENNs) combining SL and DNNs~\cite{zhao-Neurips-2020}.  However, although belief theories have been explored for several decades and their uncertainty research has been matured more than any other fields, their applications in reasoning and quantifying uncertainty in DL are still in an infant stage.   
 
\item Although we can consider misclassification detection and out-of-distribution detection tasks to evaluate the accuracy of predictive uncertainty measured, the metrics of predictive uncertainty have not been validated as it is hard to determine the ground truth of measured uncertainty. To have more valid metrics of predictive uncertainty, we need to develop a way to evaluate the data generation process by considering the causes of uncertainty (e.g., how to generate data with vagueness).
\item Uncertainty can be easily introduced by intelligent adversarial attacks taking highly deceptive poisonous or evasion attacks. Detecting adversarial attacks with the intent to increase various types of uncertainty should be the first step to reduce noises or false information that can increase uncertainty before estimating uncertainty in data for effective decision making.
\end{itemize}

\subsection{Future Research Directions}

\begin{itemize}
\item Uncertainty quantification research can be explored more for the studies using qualitative labels.  Since belief theories, such as DST, DSmT, or TBM, can provide the capability to fuse qualitative beliefs, their applications in natural language processing (NLP) are promising. Other conventional NLP methods can be compared to DSmT in handling uncertainty for the qualitative beliefs and its effect on application performance.  

\item When a different belief model, that has a different way of estimating different types of uncertainty, is combined with deep learning, we can investigate how the different ways of measuring uncertainty can impact decision making performance.

\item A belief model, such as IDM, offers the ability to derive a belief without any prior knowledge about a given proposition. In the settings that do not allow any prior knowledge or information about the proposition, IDM can allow one to make decisions under uncertainty.

\item To solve sequential decision making problems, belief models can be combined with deep reinforcement learning. In particular, as IDM does not require having prior knowledge in the decision making process, it can be easily used for an RL agent to make decisions in the environment with no prior knowledge and to learn an optimal action via trials and errors.

\item Although many different types of uncertainty have been discussed in belief models, we can capture three main uncertainty types: vacuity caused by a lack of evidence, vagueness (or fuzziness) by failing in capturing a singleton belief, and discord (or dissonance) by conflicting evidence. This can be further examined to propose a unified mathematical belief framework to quantify various uncertainty types and belief masses for its broader applicability.

\item Most uncertainty measurements approaches are designed for a singleton prediction, such as image or node classifications. However, they may not be able to extend for time series application because they ignore temporal dependencies in uncertainty quantification. It is critical to developing novel uncertainty metrics considering temporal dependencies of time series data. For example, we may consider fusion operators in SL to fuse the subjective opinions from different time steps.

\item Most uncertainty estimation research focuses on unstructured tasks, such as classification and regression tasks. Meanwhile, for the structured prediction tasks, such as language modeling (e.g., machine translation and named entity recognition), we can further investigate a general, unsupervised, interpretable uncertainty framework.
\end{itemize}

\section*{Acknowledgement}
This work is partly supported by the Army Research Office under Grant Contract Number W91NF-20-2-0140 and NSF under the Grant Number 2107449, 2107450, and 2107451. The views and conclusions contained in this document are those of the authors and should not be interpreted as representing the official policies, either expressed or implied, of the Army Research Laboratory or the U.S. Government. The U.S. Government is authorized to reproduce and distribute reprints for Government purposes notwithstanding any copyright notation herein.

\bibliographystyle{ACM-Reference-Format}
\bibliography{ref}

\end{document}


\title{Supplement Document: A Survey on Uncertainty Reasoning and Quantification for Decision Making: Belief Theory Meets Deep Learning}

\author{Zhen Guo*, Zelin Wan*, Qisheng Zhang}
\authornote{The authors with * have equally contributed to this work.}
\email{{zguo, zelin, qishengz19}@vt.edu}
\orcid{https://orcid.org/0000-0002-6563-5934, https://orcid.org/0000-0001-5293-0363, https://orcid.org/0000-0001-8785-8437}
\affiliation{%
  \department{Department of Computer Science}
  \institution{Virginia Tech}
  \streetaddress{7054 Haycock Road}
  \city{Falls Church}
  \state{VA}
  \country{USA}
  \postcode{22043}
}

\author{Xujiang Zhao*, Feng Chen}
\email{{xujiang.zhao, feng.chen}@utdallas.edu }
\orcid{https://orcid.org/0000-0003-4950-4018, https://orcid.org/0000-0002-4508-5963}
\affiliation{%
  \department{Department of Computer Science}
  \institution{University of Texas at Dallas}
  \streetaddress{800 W Campbell Rd}
  \city{Richardson}
  \state{TX}
  \country{USA}
  \postcode{75080}
}

\author{Jin-Hee Cho, Qi Zhang}
\email{{jicho, qiz21}@vt.edu}
\orcid{https://orcid.org/
0000-0002-5908-4662, https://orcid.org/0000-0002-3607-3258}
\affiliation{%
  \department{Department of Computer Science}
  \institution{Virginia Tech}
  \streetaddress{7054 Haycock Road}
  \city{Falls Church}
  \state{VA}
  \country{USA}
  \postcode{22043}
}

\author{Lance M. Kaplan}
\email{lance.m.kaplan.civ@army.mil}
\orcid{https://orcid.org/0000-0002-3627-4471, https://orcid.org/}
\affiliation{%
  \institution{US Army Research Laboratory}
  \streetaddress{2800 Powder Mill Rd.}
  \city{Adelphi}
  \state{MD}
  \country{USA}
  \postcode{20783}
}

\author{Dong H. Jeong}
\email{djeong@udc.edu}
\orcid{https://orcid.org/0000-0001-5271-293X}
\affiliation{%
  \department{Department of Computer Science of Information Technology}
  \institution{University of the District of Columbia}
  \streetaddress{4200 Connecticut Ave NW}
  \state{Washington, DC}
  \country{USA}
  \postcode{20008}
}

\author{Audun J{\o}sang}
\email{audun.josang@mn.uio.no}
\orcid{https://orcid.org/0000-0001-6337-2264}
\affiliation{%
  \department{Department of Informatics}
  \institution{University of Oslo}
  \streetaddress{Ole-Johan Dahls hus Gaustadalléen}
  \city{Oslo}
  \country{Norway}
  \postcode{23b 0373}
}

\renewcommand{\shortauthors}{Guo et al.}

\maketitle

\appendices 

\section{Dempster Shafer Theory (DST)}

\subsection{Variants of DST} Those variant techniques usually aim to solve the problem of the original DST. For example, Dempster's rule requires a normalization operation to move the zero belief into a null set. However, Zadeh discovered that such normalization operation could lead to a counterintuitive result/belief in some situations~\cite{zadeh1979validity, yager1987dempster, zadeh1986simple}. To eliminate this problem, \citet{yager1987dempster} proposed a DST-concept-based technique for generating beliefs from evidence combining in Dempster Shafer framework.  
\citet{murphy2000combining} also solved an issue in Dempster's combination rule, introducing greater uncertainty with greater conflict evidence, by refining the normalization part with the average belief. This modification removes the mass assigned to null set $\emptyset$, and also combines weighted averages of the masses $n-1$ times in Eq. (6) of the main paper, where $n$ is the number of values in set $\Theta$. In addition, transferable belief model (TBM) is one of the well-known variants of DST, which is described in detail in Section 3.3 of the main document.

\subsection{Uncertainty Quantification} \citet{smarandache2012comparative} measured uncertainty in DST by its multiple dimensions as follows:
\begin{enumerate}[leftmargin=*, noitemsep]
\item {\em Auto-conflict} represents the conflict generated by a belief function $e$ with conjunctive rule~\cite{osswald2006understanding}. The auto-conflict of order $n$ ($a_n(e)$ with $n\geq 1$) fuses $n$ identical belief functions and is estimated by:
\begin{gather}
a_n(e) = m_{Conj(e_1, \ldots ,e_n)}(\emptyset), \\
\text{where  } m_{Conj}(X) = \sum_{Y_1 \cap \dots \cap Y_M = X} \prod_{j=1}^M m_j(Y_j)
\end{gather}
where 
$m_{Conj}(X)$ is non-normalized conjunctive rule, which is a multinormial form of DST combination rule shown in Eq. (6) of the main document 
and $Y_M$ is a proposition support from information source $M$. Hence, $m_{Conj(e_1, \ldots ,e_n)}(\emptyset)$ means each belief function $e$ is disjointed without any a set of intersection.



\item {\em Non-specificity} is a generalization of Hartley entropy, which is aligned with Shannon entropy when base 2 is used in a logarithmic term. That is, non-specificity increases when there is a large number of subsets $A \subseteq \Theta$ because it implies more subsets of $\Theta$ overlap in their elements.  This non-specificity can be treated as a weighted sum of the Hartley measure~\cite{hartley1928transmission} for different focal elements $A$ in $P(\Theta)$ and is measured by: 
\begin{equation}
N(m) = \sum_{A\subseteq \Theta}m(A)\log_2\vert A\vert.   
\end{equation}

\item {\em Confusion}~\cite{hohle1982entropy} refers to uncertainty caused by a lack of evidence supporting each set $A \subseteq \Theta$ and estimated by:
\begin{equation}
\mathrm{Confusion}(m) = -\sum_{A\subseteq \Theta}m(A)\log_2(Bel(A)).    
\end{equation}
As shown above, confusion increases when there exists a large number of subsets $A \subseteq \Theta$ and each belief in set $A$ is very small.

\item {\em Dissonance}~\cite{yager2008entropy} is similar to confusion but differs in using plausibility rather than belief. That is, dissonance increases when there are a large number of subsets $A \subseteq \Theta$ and its associated plausibility is very low. This implies that the difference between the lower bound (i.e., belief) and upper bound (i.e., plausibility) is small. Hence, dissonance is more likely to be higher when there are a sufficient amount of evidence (i.e., uncertainty due to a lack of evidence is low) supporting all possible beliefs (i.e., subsets) generating each $pl(A)$ with a very small probabilities. The dissonance is measured by: 
\begin{equation}
\mathrm{Dissonance}(m) = -\sum_{A\subseteq \Theta} m(A) \log_2(pl(A)).  
\end{equation}
Note that in~\cite{yager2008entropy}, natural logarithmic (i.e., $\ln$) is used rather than $\log_2$.

\item {\em Aggregate uncertainty measure} (AU) is a generalized {\em Shannon entropy} for {\em total uncertainty} and measured by:
\begin{equation}
\mathrm{AU}(Bel)=\mathrm{max}\Big[-\sum_{\theta\in\Theta} p_\theta \log_2 p_\theta \Big].    
\end{equation}
Recall that $\Theta = \{W, Z, L\}$ and $\theta$ refers an element in set $\Theta$. 

\item {\em Ambiguity measure} (AM)~\cite{jousselme2006measuring} represents non-specificity and discord where the \textit{discord} means disagreement in choosing several alternatives and \textit{non-specificity} refers to a situation with two or more alternatives left unspecified.  The AM is obtained by:
\begin{equation}
\mathrm{AM}(m) = \sum_{\theta\in\Theta} \mathrm{BetP_m}(\theta)\log_2(\mathrm{BetP_m}(\theta)),    
\end{equation}
where $\mathrm{BetP_m}$ is defined by the Generalized pignistic transformation (GPT), a pignistic~\footnote{The pignistic means a probability assigned to an option by a person who is required to make a rational decision~\cite{Smets00}.} generalized basic belief assignment in DSmT, as estimated in Eq. (10) of the main document. 
\end{enumerate}

\section{Dezert-Smarandache Theory (DSmT)}

\subsection{Belief Formation} \label{subsec: belief-formation} The belief masses is formulated via the following concepts: 
\begin{itemize}
\item {\em A hyper-power set}: The closed finite set (i.e., frame), denoted by $\Theta$, has $n$ hypotheses elements, represented by $\Theta = {\theta_1, \ldots, \theta_n}$.  The hyper-power set $D^\Theta \overset{\Delta}{=}(\Theta,\cup,\cap)$ is defined as all composite subsets built from elements of $\Theta$ with $\cup$ and $\cap$ operators.  In general, notation $G^\Theta$ covers the belief functions defined under any frames, such as a power set $P^\Theta\overset{\Delta}{=}(\Theta,\cup)$for DST, a hyper-power set $D^\Theta$ or a super-power set for DSmT~\cite{smarandache2009advances}.
\item {\em Free and hybrid DSm models}: DST is the most restricted hybrid DSm model, denoted by $\mathcal{M}^0(\Theta)$, because all the elements are exhaustive and exclusive.  However, in real world fusion problems, the hypothesis can be vague and imprecise.  The exclusive elements, $\theta_i$'s, may not be identified and separated.  When there is no constraint on the elements (i.e., hypothesis elements $\theta_i$'s can overlap), this model is the free DSm model as $\mathcal{M}^f(\Theta)$.  Hybrid DSm models, $\mathcal{M}(\Theta)$, take into account some exclusivity constraints and non-existential constraints (i.e., changes of frame $\Theta$ with time in a dynamic fusion problem).  DSmT works with any kind of hybrid models, such as free DSm model, hybrid model, or Shafer's model (i.e., DST)~\cite{smarandache2009advances}.

\item {\em Generalized belief functions}: The generalized belief and plausibility functions, $Bel(A)$ and $pl(A)$, are defined in a same way DST defines a belief by Eq. (5) of the main document. 

\item {\em Qualitative belief assignment}: The qualitative beliefs are a set of $m$ linguistic labels $L=\{L_0, L_1,$ $L_2, \ldots, L_m, L_{m+1}\}$ with a total order relationship $\prec$ where $L_1\prec L_2\prec \ldots \prec L_m$.  The example of ordered labels is $L=$ \{very poor, poor, good, very good\}.  $L_0$ and $L_{m+1}$ represent the boundary of numeric values in $[0, 1]$ for the labels.  With numeric values, the labels can have qualitative operators of label addition, label multiplication, and label division in~\cite{Smarandache04}.  The qualitative belief assignment is a mapping function by~\cite{smarandache2009advances}: \begin{equation}
\mathrm{qm}(\cdot):G^\Theta \mapsto L.   
\end{equation}

\item {\em Fusion of precise beliefs by the classic DSm rule of combination}: For the free DSm model, $\mathcal{M}^f(\Theta)$, the belief functions combination, $m_{\mathcal{M}^f(\Theta)} \equiv m(\cdot) \overset{\Delta}{=} [m_1 \oplus m_2](\cdot)$, from the two independent sources is the conjunctive consensus of the sources~\cite{Smarandache04}:
\begin{equation}
    \forall \; C\in D^\Theta, \enspace m_{\mathcal{M}^f(\Theta)} (C) \equiv m(C) = \sum_{A, B\in D^\Theta, A\cap B=C}m_1(A)m_2(B),
    \label{eq:precise-dsm}
\end{equation}
where belief functions from two sources $Bel_1(\cdot)$ and $Bel_2(\cdot)$ are related to generalized basic belief assignments, $m_1(\cdot)$ and $m_2(\cdot)$.  This function uses the same method as Dempster's rule of combination in Eq. (6) of the main document but the fusion space can be expanded to hyper-sets. 
\item {\em Fusion of precise beliefs by the hybrid DSm rule of combination}: The hybrid DSm rule (DSmH) works for the set with integrity constraints in a hybrid model, $\mathcal{M}(\Theta) \neq \mathcal{M}^f(\Theta)$.  For all $A \in D^\Theta$~\cite{Smarandache04}:
\begin{equation}
m_{DSmH}(A) = m_{\mathcal{M}(\Theta)}(A)\overset{\Delta}{=}\phi(A)[S_1(A) + S_2(A)+S_3(A)],
\label{eq:precise-dsmh}
\end{equation}
\begin{equation}
    S_2(A)=\sum_{\substack{X_1, X_2, \ldots,X_k \in \emptyset\\ 
    (\mathcal{U}=A) \vee (\mathcal{U}\neq\emptyset) \wedge (A=I_t)}} \prod^k_{i=1} m_i(X_i), \;
    S_3(A) = \sum_{\substack{X_1, X_2, \ldots,X_k \in D^\Theta, (X_1\cup X_2\cup\ldots\cup X_k)=A, \\
    X_1\cap X_2\cap\ldots\cap X_k)\in\emptyset}}\prod^k_{i=1} m_i(X_i),
    \label{eq:precise-dsmh2}
\end{equation}
where $\phi(A)$ indicates the non-emptiness of a set $A$, i.e., $\phi(A)=1$ if $A\notin\emptyset$.  $S_1(A) = m_{\mathcal{M}^f(\theta)}(A)$ is the classic rule based on the free DSm model as shown in Eq.~\eqref{eq:precise-dsm}. The  $S_2(A)$ is the relative or total ignorance with non-existential constraints in some dynamic problems, calculated from mass of empty sets $X_i$, where $i$ is one of the $k$ sources of evidence.  Total ignorance, $I_t$, is defined as $I_t \overset{\Delta}= \theta_1\cup\theta_2\cup \ldots \cup\theta_n$, meaning the union of $n$ hypotheses elements.  The mass of relatively empty sets in $S_2(A)$ is defined as the $X_1, X_2, \ldots, X_k\in\emptyset$ where $\mathcal{U}=u(X_1)\cup u(X_2)\cup\ldots\cup u(X_k)$ and $u(X)$ is the union of all the granules $\theta$ forming $X$.  $S_3(A)$ is the canonical disjunctive form of non-empty sets $X_i$, calculated from relatively empty sets. The details of $S_2(A)$ and $S_3(A)$ are defined in~\cite{Smarandache04} and their formulations are shown in Eq.~\eqref{eq:precise-dsmh2}.

\item {\em Fusion of imprecise beliefs by the classic DSm rule of combination}: For $k$ sources of evidence, the imprecise belief is defined by~\cite{Smarandache04}:
\begin{equation}
\forall \; A\neq\emptyset \in D^\Theta, \enspace m^I(A) =\sum_{\substack{X_1, X_2, \ldots,X_k \in D^\Theta\\
    (X_1\cap X_2\cap\ldots\cap X_k)=A}} \prod_{i=1,\ldots,k} m_i^I(X_i),
    \label{eq:imprecise-dsm}
\end{equation}
where $\sum$ and $\prod$ are for summation and product operation of sets, respectively.  The imprecise beliefs are defined by a set of points or continuous intervals in $[0,1]$.  The notations of intervals can either be open, closed or half-open intervals (i.e., $(a,b), [a,b]$, or $[a,b)$).
\item {\em Fusion of imprecise beliefs by hybrid DSm rule of combination}: The hybrid model DSm rule from the sets of imprecise beliefs are defined by~\cite{Smarandache04}: 
\begin{equation}
m_{DSmH}^I(A) = m_{\mathcal{M}(\Theta)}^I(A)\overset{\Delta}{=}\phi(A)\cdot[S_1^I(A) + S_2^I(A)+ S_3^I(A)],
\label{eq:imprecise-dsmh}
\end{equation} 
\begin{equation}
    S^I_2(A)=\sum_{\substack{X_1, X_2, \ldots,X_k \in \emptyset\\ 
    (\mathcal{U}=A) \vee (\mathcal{U}\neq\emptyset) \wedge (A=I_t)}} \prod^k_{i=1} m^I_i(X_i), \;
    S_3^I(A) = \sum_{\substack{X_1, X_2, \ldots,X_k \in D^\Theta, (X_1\cup X_2\cup\ldots\cup X_k)=A, \\
    X_1\cap X_2\cap\ldots\cap X_k)\in\emptyset}}\prod^k_{i=1} m^I_i(X_i),
    \label{eq:imprecise-dsmh2}
\end{equation}
where the $+$ and $\cdot$ operators are both for two sets.  $S_1^I(A)=m^I(A)$ in Eq.~\eqref{eq:imprecise-dsm} and $S_2^I(A)$ and $S_3^I(A)$ in Eq.~\eqref{eq:imprecise-dsmh2} are described similar to the precise belief DSm rule of combination in~\cite{Smarandache04}.
\item {\em Fusion of qualitative precise beliefs}: The qualitative DSm Classic rule (q-DSmC) is similar to Eq.~\eqref{eq:precise-dsm} by the qualitative conjunctive rule as below~\cite{Smarandache04}:
\begin{eqnarray}
\mathrm{qm}_{qDSmC}(A)=\sum_{\substack{X_1,\ldots,X_k \in D^\Theta\\
X_1\cap \ldots\cap X_k=A}} \; \; \prod^k_{i=1}\mathrm{qm}_i(X_i),\enspace \; \text{for} \; k\geq 2, \; \\ \text{where} \; \; \forall \; A\in D^\Theta\backslash\{\emptyset\}, \enspace \mathrm{qm}_{qDSmC}(\emptyset)=L_0, \nonumber
\end{eqnarray}
where $k$ refers to the number of sources.

The qualitative hybrid DSm rule (q-DSmH) is similarly defined to Eq.~\eqref{eq:precise-dsmh} by~\cite{Smarandache04}:
\begin{equation}
    \mathrm{qm}_{qDSmH}(A) \overset{\Delta}{=}\phi(A)[qS_1(A) + qS_2(A)+qS_3(A)] \enspace \text{where} \enspace \mathrm{qm}_{qDSmH}(\emptyset)=L_0,
    \label{eq:precise-qdsmh}
\end{equation}
\begin{equation}
    qS_2(A)=\sum_{\substack{X_1, X_2, \ldots,X_k \in \emptyset\\ 
    (\mathcal{U}=A) \vee (\mathcal{U}\neq\emptyset) \wedge (A=I_t)}} \prod^k_{i=1} qm_i(X_i), \;
    qS_3(A) = \sum_{\substack{X_1, X_2, \ldots,X_k \in D^\Theta, (X_1\cup X_2\cup\ldots\cup X_k)=A, \\
    X_1\cap X_2\cap\ldots\cap X_k)\in\emptyset}}\prod^k_{i=1} qm_i(X_i),
    \label{eq:precise-qdsmh2}
\end{equation}
where $qS_1(A)=\mathrm{qm}_{qDSmC}(A)$ and the details of $qS_2(A)$ and $qS_3(A)$ are defined in~Eq.~\eqref{eq:precise-dsmh2}.
\item {\em Proportional conflict redistribution (PCR) rules}:  Compared to DSmH, which considers a direct transfer of partial conflicts to partial uncertainties, PCR transfers total or partial conflicting belief functions to non-empty sets in the conflicts in proportion to the belief functions by sources.  The belief function redistribution generates several versions of PCR rules. We take the following two examples:   
\begin{itemize}
\item {\em PCR5} is the most sophisticated PCR rule.  The PCR5 formula of combining two sources, considering $m_{PCR5}(\emptyset)=0$ and $\forall X\in G^\Theta \backslash\{\emptyset\}$, is defined by~\cite{Smarandache04}:
\begin{equation}
m_{PCR5}(X) = m_{12}(X) +  \sum_{\substack{Y \in G^\Theta\backslash\{X\}\\
X\cap Y=\emptyset}} \Bigg(\frac{m_1(X)^2 m_2(Y)}{m_1(X)+ m_2(Y)} + \frac{m_2(X)^2 m_1(Y)}{m_2(X)+ m_1(Y)}\Bigg),
\label{eq:pcr5}
\end{equation}
where all sets are in canonical form and $m_{12}(X) \equiv m_{\cap}(X)$ for the conjunctive consensus on $X$ and all denominators are not zero.  If a denominator is zero, $m_{12}(X)=0$.  The PCR5 rule also provides good results for combining qualitative results to be a refined label. 

\item {\em PCR6} is a more intuitive PCR formula than other PCR formulas~\cite{Smarandache04}: 
\begin{equation}
m_{PCR6}(X) = m_{12 \ldots k}(X) +  \sum_{\substack{X_1, X_2, \ldots, X_k \in G^\Theta\backslash\{\emptyset\}\\
X_1\cap X_2\cap \ldots \cap X_k=\emptyset}} \Big(\sum_{r=1}^k\delta_{X_r}^X\cdot m_r(X_r)\Big)\cdot\frac{m_1(X)^2 m_2(Y)}{m_1(X)+ m_2(Y)},
\label{eq:pcr6}
\end{equation}
where $\delta_{X_r}^X=1$ if $X=X_r$; $\delta_{X_r}^X=0$ otherwise.  The $m_{12 \ldots s}(X) \equiv m_{\cap}(X)$ is the conjunctive consensus on $X$ for $s>2$ sources.  For $s=2$ sources, PCR5 and PCR6 have the same formulas.  \end{itemize}
\end{itemize}

\subsection{Decision Making in DST using CPT} For decision-making, DSmT extends the probability function, called {\em classical pignistic transformation} (CPT) in DST, into two ways:
\begin{enumerate}
\item {\em Generalized pignistic transformation (GPT)}: To make a rational decision, a subjective/pignistic probability function $BetP(\cdot)$ in Eq. (10) of the main paper is constructed, based on any generalized basic belief assignment $m(\cdot)$ from the DSmT rules of combination by expanding the set to $D^\Theta$ and counting the cardinal of propositions in the hyper-power set.

\item {\em DSmP transformation}: In GPT, the decision to map belief to the probability is in a fixed way but DSmP provides variations in subjective probabilities. DSmP is a new probability transformation which considers both the values and the cardinality of the elements involved in each ignorance in the proportional redistribution process.  This mapping is defined by $DSmP_\epsilon (\emptyset)=0$ and $\forall X\in G^\Theta \backslash\{\emptyset\}$:
\begin{equation}
    DSmP_\epsilon(X) = \sum_{Y\in G^\Theta}\frac{
    \sum_{\substack{Z \subseteq X\cap Y,\\
    |Z|_\mathcal{M}=1}} m(Z) + \epsilon\cdot |X\cap Y|_\mathcal{M}}
    {\sum_{\substack{Z \subseteq Y,\\
    |Z|_\mathcal{M}=1}} m(Z) + \epsilon\cdot |Y|_\mathcal{M}}m(Y),
    \label{eq:dsmp}
\end{equation}
where $|\cdot|_\mathcal{M}$ is the DSm cardinal of proposition in $D^\Theta$ and $\epsilon\geq 0$ is parameter for a subjective probability measure of $m(\cdot)$.  The smaller $\epsilon$, the bigger PIC value.  When $\epsilon=1$, $DSmP_{\epsilon=1} = BetP$ (i.e., Eq.~\eqref{eq:dsmp} is reduced to Eq. (10) of the main document).  DSmP is a more accurate distribution of the ignorance masses to the elements than $BetP$. DSmP and $BetP$ work in both DST (Shafer's model) and DSmT (free or hybrid models).
\end{enumerate}

\section{Bayesian Inference}

\subsection{Belief Formation} There are variants of quantifying the uncertainty of variables depending on different sampling methods as follows:
\begin{enumerate}[leftmargin=*]
\item {\em Binomial model}: A random variable $Y\in \{0,1,\ldots,n\}$ has a binomial distribution~\cite{hoff2009first} if
\begin{equation}
\mathrm{Pr}(Y=y|\theta)=\mathrm{dbinom}(y, n, \theta) = \left( \begin{array}{c}n\\y\end{array} \right) \theta^y(1-\theta)^{n-y}, \enspace \text{where} \enspace y\in {\{0, 1, \ldots, n\}},
\end{equation}
where E[$Y|\theta$]=$n\theta$ and Var[$Y|\theta$]=$n\theta(1-\theta)$. 
The prior is an uncertain variable $\theta$ from $\mathrm{Beta}(a,b)$ distribution~\cite{hoff2009first} if 
\begin{equation}
p(\theta)=\mathrm{B}(\theta,a,b)=\frac{\Gamma(a+b)}{\Gamma(a)\Gamma(b)}\theta^{a-1}(1-\theta)^{b-1}\enspace \mathrm{for}\enspace 0\leq\theta\leq 1,
\end{equation}
where $\Gamma(x+1)=x!$, E[$\theta$]=$\frac{a}{a+b}$, and Var[$\theta$]=$\frac{E[\theta]E[1-\theta]}{a+b+1}$. Then the posterior distribution~\cite{hoff2009first} is given by:
\begin{equation}
p(\theta|y) = \mathrm{dbeta}(\theta, a+y, b+n-y),
\end{equation}
where E[$\theta|y$]=$\frac{a+y}{a+b+n}$ and Var[$\theta|y$]=$\frac{\mathrm{E}[\theta|y]\mathrm{E}[1-\theta|y]}{a+b+n+1}$.  This reflects that beta priors are conjugate for the binomial sampling model.

\item {\em Poisson model}: A random variable $Y$ has a Poisson distribution~\cite{hoff2009first} if
\begin{equation}
\mathrm{Pr}(Y=y|\theta)=\mathrm{dpois}(y,\theta) = \theta^y \frac{e^{-\theta}}{y!} \enspace \mathrm{for} \enspace y\in \{0,1,2, \ldots\},
\end{equation}
with E[$Y|\theta$]=$\theta$ and Var[$Y|\theta$]=$\theta$.  The conjugate prior can be a gamma$(a,b)$ distribution~\cite{hoff2009first} if
\begin{equation}
p(\theta)=\mathrm{\Gamma}(\theta,a,b)=\frac{b^a}{\Gamma(a)}\theta^{a-1}e^{-b\theta}\enspace \mathrm{for}\enspace \theta,\; a,\;b > 0,
\end{equation}
with E[$\theta$]=$\frac{a}{b}$, Var[$\theta$]=$\frac{a}{b^2}$.  Combining the gamma family priors and Poisson sampling model, the posterior distribution of $\theta$ is given by~\cite{hoff2009first}:
\begin{equation}
p(\theta|y) = \mathrm{\Gamma}(a+y, b+n),
\end{equation}
where $b$ is the number of prior observations, $a$ is the sum of counts from $b$ prior observations, and E[$\theta|y$]=$\frac{a+y}{b+n}$.

\item {\em Monte Carlo approximation}: For the arbitrary posterior distributions, Monte Carlo approximation generates random sampling with large $S$ samples to model $p(\theta|y_1, y_2, \ldots, y_n)$~\cite{hoff2009first}.  The empirical distribution of the samples $\{\theta^{(1)}, \ldots, \theta^{(S)}\}$ can represent $p(\theta|y_1, y_2, \ldots, y_n)$.  For $S\rightarrow \infty$, the mean $\theta$ (i.e., $\overline{\theta}$) and the variance of the samples are given by:
\begin{eqnarray}
\overline{\theta} = \sum^S_{s=1} \frac{\theta^{(s)}}{S} \rightarrow \mathrm{E}[\theta|y_1, \ldots, y_n], \; \; 
\sum^S_{s=1}\frac{(\theta^{(s)}-\overline{\theta})^2}{(S-1)} \rightarrow \mathrm{Var}[\theta|y_1, \ldots, y_n].
\end{eqnarray}

\item {\em The normal model}: Normal model is a two-parameter model with mean $\theta$ and variance $\sigma^2$.  If the sampling model is normal, $\{Y_1, \ldots, Y_n|\theta, \sigma^2\}\sim$ normal $(\theta, \sigma^2)$, the joint sampling density is~\cite{hoff2009first}:
\begin{equation}
p(y_1, \ldots,y_n|\theta,\sigma^2)=\prod_{i=1}^n p(y_i|\theta,\sigma^2)=(2\pi\sigma^2)^{-\frac{n}{2}}e^{-\frac{1}{2}\sum \Big(\frac{y_i-\theta}{\sigma}\Big)^2}.
\end{equation}
When $\sigma^2$ is known, a posterior distribution can be inferred with conditional prior $p(\theta|\sigma^2)$.  The conjugate prior distribution is normal based on $\theta\sim \text{normal}\enspace (\mu_0, \tau_0^2)$.  The (conditional) posterior distribution is~\cite{hoff2009first}:
\begin{gather}
p(\theta|\sigma^2, y_1, \ldots, y_n) \propto \exp{\Big(-\frac{1}{2}\Big(\frac{\theta-\mu_n}{1/\tau_n}\Big)^2 \Big)},\\ \tau_n^2 = \frac{1}{\frac{1}{\tau_0^2}+\frac{n}{\sigma^2}}, \enspace 
\mu_n = \frac{\kappa_0}{\kappa_0+n}\mu_0+\frac{n}{\kappa_0+n}\bar{y}, \enspace \tau_0^2= \frac{\sigma^2}{\kappa_0}. \nonumber
\label{eq:normal-conditional}
\end{gather}

When $\sigma^2$ is also unknown, the joint inference for the posterior distribution can be calculated based on the prior distributions and sampling model normal $(\theta, \sigma^2)$ as: prior $\frac{1}{\sigma^2}\sim \mathrm{gamma} (\frac{\nu_0}{2}, \frac{\nu_0\sigma_0^2}{2})$, prior \{$\theta|\sigma^2\}\sim\mathrm{normal}(\mu_0, \frac{\sigma^2}{\kappa_0})$~\cite{hoff2009first}.  Then the joint posterior distribution is given by:
\begin{equation}
p(\theta,\sigma^2|y_1,\ldots,y_n) = p(\theta|\sigma^2, y_1,\ldots,y_n)p(\sigma^2|y_1,\ldots,y_n),
\label{eq:normal-joint}
\end{equation}
where 
\begin{gather}
\{\theta|\sigma^2, y_1, \ldots,y_n\}\sim\mathrm{normal}\Big(\mu_n, \frac{\sigma^2}{\kappa_n}\Big),\; \; \text{with}\;\; \kappa_n=\kappa_0+n, \;\; \mu_n=\frac{(\kappa_0\mu_0+n\bar{y})}{\kappa_n}, \\
\Big\{\frac{1}{\sigma^2}|y_1,\ldots,y_n\Big\}\sim\mathrm{gamma\Big(\frac{\nu_n}{2}, \frac{\nu_n\sigma_n^2}{2}\Big)}, \\
\text{with} \;\; \nu_n=\nu_0+n, \;\; \sigma_n^2=\frac{\Big[\frac{\nu_0\sigma_0^2}{2}+(n-1)s^2+ \frac{(\bar{y}-\mu_0)^2 \kappa_0n}{\kappa_n}\Big]}{\nu_n}. \nonumber
\end{gather}
\end{enumerate}

\section{Subjective Logic (SL)}

\subsection{Uncertainty Maximization} \citet{Josang16} also proposed a technique called {\em uncertainty maximization} where uncertainty refers to vacuity. Since we discussed different types of uncertainties in SL, we will call the uncertainty maximization `vacuity maximization.'  In subjective opinions formulated by SL, when vacuity (i.e., $u_X$) is 0 (i.e., complete certainty), then an opinion stops updating and is ended with its final state. However, if the opinion's dissonance is high, it leads to a situation of `multiple knowledge frames' and a decision maker still cannot make a decision even with zero uncertainty due to the same probabilities assessed for given belief masses.  To mitigate this effect, we can allow an opinion to receive more information or consider other opinions for the opinion being updated.  This can be enabled by maximizing vacuity based on the minimum belief mass~\cite{Josang16}.  Given opinion $\omega_X = (\mathbf{b}_X, u_X, \mathbf{a}_X)$ where $\mathbf{P}_X(x) = \bm{b}_X(x) + \bm{a}_X(x) u_X$ in Eq. (24) of the main document, 
the corresponding vacuity-maximized opinion is denoted by $\ddot{\omega}_X = (\ddot{\mathbf{b}}_X, \ddot{u}_X, \mathbf{a}_X)$ where $\ddot{u}_X$ and $\ddot{\mathbf{b}}_X$ are computed by:
\begin{equation}\label{eq:vacuity-maximization}
\ddot{u}_X = \min_i \Big[\frac{\mathbf{P}_X(x_i)}{\bm{a}_X(x_i)}\Big], \; \; \ddot{\mathbf{b}}_X (x) = \mathbf{P}_X(x) - \bm{a}_X(x) \ddot{u}_X, \; \; \text{for} \; \; x_i \in \mathbb{X}.    
\end{equation}
A vacuity-maximized opinion is an epidemic opinion based on the uncertain likelihood of the variable's value in unknown past or future for a given proposition. Notice $\mathbf{a}_X$ stays same in $\ddot{\omega}_X$ and the vacuity-maximized opinion is the same as an epidemic opinion.  Remind that the purpose of updating $\omega_X$ to $\ddot{\omega}_X$ is to allow the opinion to be further updated by receiving new evidence or being combined with other opinions which are possible only when $u_X > 0$.

\subsection{Uncertainty Quantification} Uncertainty associated with each belief mass are:
\begin{enumerate}[leftmargin=*]
\item {\em Belief vacuity} (a.k.a. focal uncertainty), $\mathbf{u}^F_X (x)$, is computed by:
\begin{equation} \label{eq:focal-vacuity}
\mathbf{u}^F_X (x) = \mathbf{a}_X (x) u_X,
\end{equation}
where $x \in \mathscr{R}(\mathbb{X})$ and $\mathbf{u}^F_X (x): \mathscr{R}(\mathbb{X}) \rightarrow [0, 1]$.
\item {\em Belief vagueness}, $\bm{b}^{\mathrm{V}}_{X}(x)$, is the vague belief mass associated with an individual belief $\bm{b}_{X}(x)$ where the belief has a composite value, $x \in \mathscr{C}(\mathbb{X})$.  
\begin{gather}
\bm{b}^{\mathrm{V}}_{X}(x) =
\sum \limits_{\stackrel{x_{i} \in \mathscr{C}(\mathbb{X})}{x_{i}\not\subseteq x}}  \bm{a}_{X}(x{|}x_{i})\;\bm{b}_{X}(x_{i})~,\; \; \; 
\forall x\in \mathscr{R}(\mathbb{X}),
\label{eq:belief-vagueness}
\end{gather}
where $\bm{a}_{X}(x{|}x_{i})$ is found in Eq. (27) of the main document. 
Note that this belief vagueness can be only measured for hyper opinions (i.e., $x_{i} \in \mathscr{C}(\mathbb{X})$).  

\item {\em Belief dissonance}, $\bm{b}^{\mathrm{Diss}}_{X}(x_{i})$, estimates the difference between belief $\bm{b}_{X}(x_{i})$ and other belief masses by:
\begin{gather}
\bm{b}^{\mathrm{Diss}}_{X}(x_{i}) = \frac{\bm{b}_{X}(x_{i})\!\!\!\! \sum\limits_{x_j \in \mathbb{X}\setminus x_i}\!\!\!\!\!\bm{b}_{X}(x_{j}) \mbox{Bal}(x_{j},x_{i})}{\sum\limits_{x_j \in \mathbb{X}\setminus x_i}\bm{b}_{X}(x_{j})},    
\label{eq:belief-dissonance}
\end{gather}
where the relative mass balance between belief masses, $\bm{b}_{X}(x_{j})$ and $\bm{b}_{X}(x_{i})$, is given by:
\begin{equation}
\label{eq:belief-balance}
\mbox{Bal}(x_{j},x_{i}) = 1 - \frac{|\bm{b}_{X}(x_{j})-\bm{b}_{X}(x_{i})|}{\bm{b}_{X}(x_{j})+\bm{b}_{X}(x_{i})}.
\end{equation}
\end{enumerate}

\section{Belief Meets Deep Learning}

\subsection{Fuzzt Set}
Let $S$ be a nonempty set. A fuzzy set $\mathcal{P}$ in $S$ is characterized by a membership function: $\mu_p:S\rightarrow [0, 1]$. That is
\begin{equation}
\mu_{\mathcal{P}}(x)= \begin{cases}1, & \text { if } x \in S \\ 0, & \text { if } x \notin S \\ (0,1) & \text { if } x \text { is partly in } S\end{cases}
\end{equation}
Alternatively, a fuzzy set $\mathcal{P}$ in $S$ is an object having the form
\begin{equation}
\mathcal{P}=\left\{\left\langle x, \mu_{p}(x)\right\rangle \mid x \in S\right\}
\end{equation}
where the function:  $\mu_p(x):S\rightarrow [0, 1]$ defines the degree of membership of the element, $x\in S$.

The closer the membership value $\mu_p(x)$to 1, the more $x$
belongs to $S$, where the grades 1 and 0 represent full membership and full non-membership. Fuzzy set is a collection of objects with graded membership, that is, having degree of membership.

\subsection{Rough Set}
Rough set theory is first introduced in~\cite{pawlak1982rough,pawlak1995vagueness} ] to deal with the problems of inexact, uncertain, or vague knowledge. An information system is defined by the four-tuple $S=<U, A, V, f>$ where $U$ is a finite nonempty set called the universe of primitive objects and $A$ is a finite nonempty set of attributes. Each attribute $a \in A$ is associated with a domain set $V_a$ and $V=\bigcup_{a\in A} V_a$. The mapping $f:U\times A\rightarrow V$ is an information function. Assume $S$  is an information system and $M\subseteq A$. Two objects $x, y \in U$ are indiscernible from each other by the set of attributes $M$ in $S$ if and only if for every $a\in M, f(x, a)=f(y, a)$. Therefore, every $M \subseteq A$ has a  the indiscernibility relation. Rough set theory defines two approximations for any concept set $X\subseteq U$ and attribute set $M \subseteq A$. Using the knowledge of $M, X$ can be approximated by the M-lower approximation $\underline{MX}$ and M-upper approximation $\overline{MX}$: 
\begin{equation}
\begin{aligned}
&\underline{M X}=\{x|[x]_M \subseteq X\} \\
&\overline{M X}=\{x|[x]_M \cap X \neq \varnothing \}
\end{aligned}
\end{equation}
Where $[x]_M$ is the equivalence classes of the M-indiscernibility relation. The M-boundary region of set $X$ is defined by
\begin{equation}
    BND_M(X) = \overline{MX} - \underline{MX}
\end{equation}
where $\underline{MX}$ is the set of all objects in $U$, which can be certainly classified as members of X with respect to the set of attributes $M$. $\overline{MX}$ is the set of objects in $U$, which can possibly be classified as members of X with respect to the set of attributes $M$. The boundary region is the set of objects that cannot certainly be classified to $X$ only by employing the set of attributes $M$. $BND_M (X)$ describes the vagueness of $X$. If
$BND_M (X) = \varnothing$, then $X$ is crisp (exact) with respect to $M$
and if $BND_M (X) \neq \varnothing$, then $X$ is called a rough (inexact) set.

\begin{table}[t]
\small 
\caption{\sc \centering Comparison of Belief and ML/DL Models In Their Uncertainty Consideration and Applications}
\label{tab:comparison-BMs}
\footnotesize
\centering
\begin{tabular}{|P{1.9cm}|P{1.9cm}|P{2cm}|P{2cm}|P{2cm}|P{2cm}|}
\hline
Model & Uncertainty type(s) & Uncertainty cause(s) & Uncertainty metric(s) & Application domain(s) & ML/DL Techniques \\
\hline
\hline
Kleene's Three-Valued Logic (TVL) & Unpredictability & Lack of information or knowledge & Unknown or unspecified state & pattern recognition & Rough DNNs\\
\hline
Dempster Shafer Theory (DST)  & Epistemic uncertainty & Lack of evidence & Plausibility - Belief & Information fusion & DBN, DNN, CNN, ENN, LSTM-RNN \\
\hline
Transferable Belief Model (TBM) & Epistemic uncertainty &  Insufficient evidence & unknown & Data or information fusion & MM-TBM, OGM algorithm \\
\hline
Dezert-Smarandache Theory (DSmT) & Fusion of precise, imprecise, qualitative beliefs & Conflicting, imprecise, subjective probability & Uncertainty, Shannon's entropy, PIC &robotics, biometrics, image fusion, trust management & SVM OAA, CNN, LSTM, RF \\
\hline
Imprevise Dirichlet Model (IDM) & Epistemic uncertainty & Lack of evidence & Eq. (14) & Utilizing upper and lower bound for conservative preference decision-making or avoiding overfitting & AdaBoost, lower expected utility \\
\hline
Fuzzy Logic (FL) & Multiple knowledge frames & Linguistic imprecision or vagueness & Entropy & Traffic control, surveillance and security, text processing, image processing,  time-series prediction & Fuzzy
deep neural networks\\
\hline
Bayesian Inference (BI) & Aleatoric, Epistemic  & data randomness,lack of data, label overlap & uncertainty, Shannon's entropy, mutual information & robotics, biometrics, image classification,face recognition & Reinforcement learning, active learning, OOD detection.\\
\hline
Subjective Logic (SL) & Vacuity, vagueness, and dissonance  & Lack of evidence, vague observations, conflicting evidence & Belief vacuity, vagueness, dissonance & Adversarial attacks, decision making & DL, DRL \\
\hline
Machine/Deep Learning & Aleatoric, Epistemic & Limited of training data, data noise, noise of measurement & Shannon's entropy, mutual information, variance, differential entropy & computer vision, natural language processing & MC-Dropout, prior network, ensemble method \\
\hline
\end{tabular}
\end{table}

\subsection{Regression Evidential Neural Networks (RENNs)}
In Section 4 of the main paper, we have discussed Evidential Neural Networks (ENNs) model for classification tasks. Next, we discuss the uncertainty quantification for regression tasks when consider the ENNs. Some metrics used in classification cannot apply to regression, such as vacuity and dissonance from Subjective Logic~\cite{zhao-Neurips-2020, josang2018uncertainty}. For regression tasks, we can directly get uncertainty information from Gaussian process~\cite{candela2004learning}. Similar to classification, epistemic and aleatoric uncertainties can be applied to regression tasks. Epistemic uncertainty can be calculated by the variance from dropout inference~\cite{kendall2015bayesian}, or estimated via an ensemble way~\cite{lakshminarayanan2016simple}. More specifically, Eq.~\eqref{regression:epistemic} gives an approximation formulation when we use dropout sampling or ensemble.
\begin{equation}
\operatorname{Var}(\mathbf{y}) \approx \frac{1}{T} \sum_{t=1}^{T} \mathbf{f}(\mathbf{x};\mathbf{\theta}_{t})^{T} \mathbf{f}(\mathbf{x};\mathbf{\theta}_{t})\left(\mathbf{x}_{t}\right)-E(\mathbf{y})^{T} E(\mathbf{y}),\mathbf{\theta}_{t} \sim q(\theta), \label{regression:epistemic}
\end{equation}
where $q(\theta)$ is the approximated posterior and $E(\mathbf{y}) \approx \frac{1}{T} \sum_{t=1}^{T} \mathbf{f}(\mathbf{x};\mathbf{\theta}_{t})$ is the predictive mean. For aleatoric uncertainty, \citet{kendall2017uncertainties} considered Gaussian likelihood as the objective function,
\begin{equation}
\mathcal{L}_{\mathrm{NN}}(\theta)=\frac{1}{N} \sum_{i=1}^{N} \frac{1}{2 \sigma\left(\mathbf{x}_{i};\theta\right)^{2}}\left\|\mathbf{y}_{i}-\mathbf{f}\left(\mathbf{x}_{i};\theta \right)\right\|^{2}+\frac{1}{2} \log \sigma\left(\mathbf{x}_{i};\theta\right)^{2}
\end{equation}
where $\sigma\left(\mathbf{x}_{i};\theta\right)$ is heteroscedastic (data-dependent) aleatoric uncertainty for each training sample, and can be learnt from a neural network. In addition, ~\citet{kendall2018multi} considered homoscedastic aleatoric uncertainty in multi-task objective, which is considered as Gaussian likelihood for the two tasks respectively, Eq.~\eqref{Mutil-task} shows a two-task objective,
\begin{equation}
\begin{array}{l}
-\log p\left(\mathbf{y}_{1}, \mathbf{y}_{2} \mid \mathbf{f}(\mathbf{x}_l;\theta)\right) \propto \frac{1}{2 \sigma_{1}^{2}}\left\|\mathbf{y}_{1}-\mathbf{f}(\mathbf{x};\theta)\right\|^{2}+\frac{1}{2 \sigma_{2}^{2}}\left\|\mathbf{y}_{2}-\mathbf{f}(\mathbf{x};\theta)\right\|^{2}+\log \sigma_{1} \sigma_{2}
\end{array} \label{Mutil-task}
\end{equation}
where $\sigma_1$ and $\sigma_2$ are the Homoscedastic (task-dependent) aleatoric uncertainty corresponding to each task. 

To address the limitation of prior network~\cite{malinin2018predictive}, ~\citet{malinin2020regression} proposed a regression prior network that considers Normal-Wishart distribution as the prior to estimate knowledge uncertainty (distributional uncertainty) for regression tasks. Normal-Wishart distribution is similar to the Dirichlet distribution, which is the conjugate prior to the categorical distribution (used in classification task on Prior Networks~\cite{malinin2018predictive}). The Normal-Wishart distribution is defined as follows:
\begin{equation}
\mathcal{N W}(\boldsymbol{\mu}, \boldsymbol{\Lambda} \mid \boldsymbol{m}, \boldsymbol{L}, \kappa, \nu)=\mathcal{N}(\boldsymbol{\mu} \mid \boldsymbol{m}, \kappa \boldsymbol{\Lambda}) \mathcal{W}(\boldsymbol{\Lambda} \mid \boldsymbol{L}, \nu),
\end{equation}
where $\boldsymbol{m}$ and $\boldsymbol{L}$ are the prior mean and inverse of the positive-definite prior scatter matrix, while $\kappa$
and $\nu$ are the strengths of belief in each prior, respectively. The parameters $\kappa$ and $\nu$ are conceptually similar to precision of the Dirichlet distribution. Usually, we use a neural network to estimate the parameters of the Normal-Wishart,
\begin{equation}
\mathrm{p}(\boldsymbol{\mu}, \boldsymbol{\Lambda} \mid \boldsymbol{x}, \boldsymbol{\theta})=\mathcal{N} \mathcal{W}(\boldsymbol{\mu}, \boldsymbol{\Lambda} \mid \boldsymbol{m}, \boldsymbol{L}, \kappa, \nu), \quad\{\boldsymbol{m}, \boldsymbol{L}, \kappa, \nu\}=\boldsymbol{\Omega}=\boldsymbol{f}(\boldsymbol{x} ; \boldsymbol{\theta}),
\end{equation}
where $\boldsymbol{\Omega}=\{\boldsymbol{m}, \boldsymbol{L}, \kappa, \nu\}$ is the parameters of the Normal-Wishart predicted by neural network. Then we obtain the posterior predictive,
\begin{equation}
\mathrm{p}(\boldsymbol{y} \mid \boldsymbol{x}, \boldsymbol{\theta})=\mathbb{E}_{\mathrm{p}(\boldsymbol{\mu}, \boldsymbol{\Lambda} \mid \boldsymbol{x}, \boldsymbol{\theta})}[\mathrm{p}(\boldsymbol{y} \mid \boldsymbol{\mu}, \boldsymbol{\Lambda})]=\mathcal{T}\left(\boldsymbol{y} \mid \boldsymbol{m}, \frac{\kappa+1}{\kappa(\nu-K+1)} \boldsymbol{L}^{-1}, \nu-K+1\right)
\end{equation}
And similar to the distribution uncertainty, we calculate the knowledge uncertainty based on mutual information,
\begin{equation}
\underbrace{\mathcal{I}[\boldsymbol{y},\{\boldsymbol{\mu}, \boldsymbol{\Lambda}\}]}_{\text {Knowledge Uncertainty }}=\underbrace{\mathcal{H}\left[\mathbb{E}_{\mathrm{p}(\boldsymbol{\mu}, \boldsymbol{\Lambda} \mid \boldsymbol{x}, \boldsymbol{\theta})}[\mathrm{p}(\boldsymbol{y} \mid \boldsymbol{\mu}, \boldsymbol{\Lambda})]\right]}_{\text {Total Uncertainty }}-\underbrace{\mathbb{E}_{\mathrm{p}(\boldsymbol{\mu}, \boldsymbol{\Lambda} \mid \boldsymbol{x}, \boldsymbol{\theta})}[\mathcal{H}[\mathrm{p}(\boldsymbol{y} \mid \boldsymbol{\mu}, \boldsymbol{\Lambda})]]}_{\text {Expected Data Uncertainty }}
\end{equation}


Similar to~\cite{malinin2020regression}, ~\cite{amini2020deep} accomplished uncertainty quantification in regression model by placing evidential priors (Normal Inverse-Gamma distribution) over the original Gaussian likelihood function and training the neural network to infer the hyperparameters of the evidential distribution. Furthermore, ~\citet{russell2019multivariate} modeled multivariate uncertainty for regression problems with neural networks, incorporated both aleatoric and epistemic sources of heteroscedastic uncertainty by training a deep uncertainty covariance matrix model directly using a multivariate Gaussian density loss function, or indirectly using end-to-end training through a Kalman filter.

We summarized the key components of belief models and ML/DL models to compare their features in Table~\ref{tab:comparison-BMs}.  In Table~\ref{tab:notations-meanings}, we summarize the notations and their meanings used in each theory.

\begin{table}[ht]
\centering
\caption{Notations and Their Meanings}
\label{tab:notations-meanings}
\small 
\begin{tabular}{|P{3cm}|P{10.5cm}|}
\hline
{\bf Notation}   &  {\bf Meaning} \\
\hline
\multicolumn{2}{|c|}{\cb {\bf TVL}} \\
\hline
$\mathcal{K}$& Kleene algebra\\
\hline
$\mathcal{K}_{3}$& Kleene’s three valued logic\\
\hline
$I$& Information system\\
\hline
$IND$& The set of equivalence relationships\\
\hline
$X=(\underline{R}X, \overline{R}X)$& Rough set\\
\hline
$\mathcal{F}$& The set of all logic functions\\
\hline
$\mathcal{A}_\mathcal{K}$& The set of all Kleene algebras\\
\hline
$\mathcal{A}_\mathcal{RS}$& The collections of all rough sets over all possible information systems\\
\hline
\multicolumn{2}{|c|}{\cb {\bf DST}} \\
\hline
$\Theta$ & A set of all propositions (also known as the \textit{frame of discernment (FOD)}) \\
\hline
$P(\Theta) $ & A power set (also known as the \textit{powerset of FOD}) \\
\hline
$A$ & Focal element in $P(\Theta)$ \\
\hline
$m$ & Belief mass (also known as evidence) \\
\hline
$Bel(A)$ & Belief of $A$ \\
\hline
$pl(A)$ & Plausibility of $A$ (i.e., $1-Dis(A)$) \\
\hline
$Dis(A)$ & Disbelief of $A$ (i.e., $1-pl(A)$) \\
\hline
IOU & Interval of Uncertainty \\
\hline
$U^T(A)$ & Total uncertainty in DST \\
\hline
\multicolumn{2}{|c|}{\cb {\bf TBM} (inherits all notations from DST)} \\
\hline
$m_B(A)$ & Belief mass supporting propositions $A$ when the conditional evidence supports proposition $B$ \\
\hline
$\overline{B}$ & A set not support B\\
\hline
$BetP$ & Probability transformed through the pignistic probability function for decision making \\
\hline
$X$ & \textit{Boolean algebra} of the subset of $\Omega$ \\
\hline
\multicolumn{2}{|c|}{\cb {\bf DSmT} (inherits all notations from DST)} \\
\hline
$D^\Theta$ & Hyper-power set of frame $\Theta$\\
\hline
$G^\Theta$ & Any set including power set, hyper-power set, and super-power set\\
\hline
$qm(\theta)$& Qualitative beliefs of evidence $\theta$\\
\hline
$m^I(A)$& Imprecise beliefs of evidence $A$\\
\hline
$E_H$& Normalized Shannon's entropy\\
\hline
$PIC$& Probabilistic information content score\\
\hline
$GPT$& Generalized pignistic transformation\\
\hline
$DSmP_\epsilon$ & Probability transformation with a subjective probability measure $\epsilon$\\
\hline
$\mathcal{M}^0(\Theta)$ & DST model, the most restricted DSm model\\
\hline
$\mathcal{M}^f(\Theta)$& Free DSm model, without constraints on the elements\\ 
\hline
$\mathcal{M}(\Theta)$& Hybrid DSm models\\
\hline
$L$& Set of labels for qualitative beliefs\\
\hline
$I_t$& Total ignorance, as the union of all hypotheses elements\\
\hline
\multicolumn{2}{|c|}{\cb {\bf IDM}} \\
\hline
$\Omega = [\omega_1, \dots, \omega_k]$ & Set of worlds (truth) / Sample space / Exhaustive set of all possible outcomes \\
\hline
$\overline{P}(A|n)$ \& $\underline{P}(A|n)$ & Upper and lower posterior probabilities from IDM, where $n$ is the number of observations towards event $A$ \\
\hline
$A$ & An observed event \\
\hline
$N$ & Total number of observations for all events\\
\hline
$s$ & Positive constant used in IDM \\
\hline
$\theta$ & Probability distribution of the observations  
\\
\hline
\end{tabular}
\end{table}

\begin{table}[ht]
\centering
\caption*{Table 2. Notations and Their Meanings (cont'd)}
\label{tab:notations-meanings2}
\small 
\begin{tabular}{|P{3cm}|P{10.5cm}|}
\hline
{\bf Notation}   &  {\bf Meaning} \\
\hline
\multicolumn{2}{|c|}{\cb {\bf Bayesian Inference}} \\
\hline
$p(\theta)$& Prior distribution of the population characteristics $\theta$ \\
\hline
$\mathcal{Y}$& Sample space\\
\hline
$p(y|\theta)$& The belief that $y$ is the dataset if $\theta$ is true\\
\hline
$p(\theta|y)$& Posterior distribution based on the observed dataset $y$\\
\hline
$\Theta$ & Parameter space for the set of all possible parameter values \\
\hline
$\mathrm{Var}^{Bin}, \mathrm{Var}^{Poiss}, \mathrm{Var}^{MC}$& Variance of Binomial, Poisson, and Monte Carlo approximation models\\
\hline
$\tau_n^2$& Posterior variance in normal model\\
\hline
$1/\tau_n^2$& Posterior inverse variance or posterior precision \\
\hline
$Y$& Dependent variables\\
\hline
$X$& Independent variables\\
\hline
$\mathrm{dbinom}(y, n, \theta)$ & Binomial distribution\\
\hline
$\mathrm{dpois}(y,\theta)$& Poisson distribution\\
\hline
$\overline{\theta}$& Mean $\theta$ from $S$ number of samples\\
\hline
\multicolumn{2}{|c|}{\cb {\bf Fuzzy Logic}} \\
\hline
$\mathscr{T}$& The fuzzy logic truth value set\\
\hline
$l_\tau$& \L ukasiewicz logic\\
\hline
$\tau$& Fuzzy set\\
\hline
$P(A)$& The probability of a fuzzy event $A$\\
\hline
$H^P(A)$& The entropy of a fuzzy event $A$\\
\hline
\multicolumn{2}{|c|}{\cb {\bf Subjective Logic}} \\
\hline
$x$ & A proposition \\
\hline
$\bm{b}_X$ & Belief mass distribution over $\mathbb{X}$ \\
\hline
$\mathbb{X}$ & A domain or a set of propositions where $x \in \mathbb{X}$ or A hyper variable in $\mathscr{R}(\mathbb{X})$ \\
\hline
$u_X$ & Uncertainty (vacuity) mass \\
\hline
$\bm{a}_X$ & Base rate (or prior belief) distribution over $\mathbb{X}$ \\
\hline
$\mathbf{P}_X(x)$ & Projected probability of belief $x$ \\
\hline
$\mathbf{u}^F_X (x)$ & Focal uncertainty \\
\hline
$\bm{b}^{\mathrm{Diss}}_{X}(x_{i})$ & Dissonance per singleton belief \\
\hline
$b^{\mathrm{TV}}_{X}$ & Total vagueness \\
\hline
$\dot{b}_{X}^{\mathrm{Diss}}$ & Total dissonance \\
\hline
$\mathcal{R}(\mathbb{X})$ & Hyper domain \\
\hline
$\bm{p}_X^H$ & Hyper-probability distribution \\
\hline
$\alpha_X$ & Strength vector over $\kappa$ number of $x$'s where $x \in \mathscr{R}(\mathbb{X})$ \\
\hline
$\mathrm{Dir}_X^H (\bm{p}_X^H; \alpha_X (x))$ & Dirichlet hyper-PDF \\
\hline
\multicolumn{2}{|c|}{\cb {\bf Evidential Deep Neural Networks}} \\
\hline
${\bf Dir(\cdot)}$& Probability density function of Dirichlet distribution\\
\hline
${\bf Cat(\cdot)}$& Probability density function of Categorical distribution\\
\hline
${\bf \alpha}$& Parameter of Dirichlet distribution\\
\hline
${\bf p}$& Class probability\\
\hline
${e}$& Evidence in subjective opinion\\
\hline
${b}$& Belief mass in subjective opinion \\
\hline
${u}$& Uncertainty mass in subjective opinion, i.e., vacuity\\
\hline
${diss}$& Dissonance uncertainty\\
\hline

\end{tabular}
\end{table}

\begin{table}[ht]
\centering
\caption*{Table 2. Notations and Their Meanings (cont'd)}
\label{tab:notations-meanings3}
\small 
\begin{tabular}{|P{3cm}|P{10.5cm}|}
\hline
{\bf Notation}   &  {\bf Meaning} \\
\hline
\multicolumn{2}{|c|}{\cb {\bf Fuzzy Deep Neural Networks}} \\
\hline
$\x$&  A set of visible units $\x \in \{0, 1\}^D$\\
\hline
$\h$&  A set of hidden units $\h \in \{0, 1\}^H$\\
\hline
$P(\x;\theta)$&  Joint probability of $\x$\\
\hline
$\mathcal{P}$&  Pythagorean fuzzy set (PFS)\\
\hline
$\mu_p(x)$&  Membership function of $x$\\
\hline
$\nu_p(x)$&  Non-membership function of $x$\\
\hline
$\pi_p(x)$&  Hesitant function of $x$\\
\hline
$\beta$&  $\beta=\mathcal{P}(\mu_\beta, \nu_\beta)$ is Pythagorean fuzzy number (PFN)\\
\hline
$h(\beta)$&  Accuracy function of $\beta$\\
\hline
$s(\beta)$&  Score function of $\beta$\\
\hline
\multicolumn{2}{|c|}{\cb {\bf Rough Deep Neural Networks}} \\
\hline
$\theta_U$&  Parameters of upper bound neuron  \\
\hline
$\theta_L$&  Parameters of lower bound neuron  \\
\hline
$W_U$ &  Weight of the upper bound neuron  \\
\hline
$W_L$ &  Weight of the lower bound neuron  \\
\hline
$b_U$ &  Bias of the upper bound neuron  \\
\hline
$b_L$ &  Bias of the lower bound neuron  \\
\hline
$\alpha$ &  Coefficient of the upper bound output  \\
\hline
$\beta$ &  Coefficient of the lower bound output  \\
\hline
$f^1$ &  {\em Sigmoid} function  \\
\hline
$f^2$ &  {\em Linear} function  \\
\hline
\end{tabular}
\end{table}

\section{Sources of Uncertainty}

\subsection{Humans}
Human beings make mistakes in daily life, either consciously or unconsciously, which leads to uncertainty. For example, humans make errors in perception, imperfect cognition, irrational propensity because of lacking of cognitive ability to judge fairly and accurately, or bias, prejudice, delusional beliefs, or distorted memories. They will all impact on decision making. Even some humans have malicious intent, which also intentionally adds uncertainty into an entity (e.g., other humans or systems) to disrupt its decision making process.


\subsection{Environments}
Environments can be very complicated. For example, the system is in a dynamic environment whose parameters keep changing, or have high temporal operations input that requests the ability to quickly communicate with the environments, such as some external factors including weather, terrains, and destruction. All those aforementioned reasons can lead to generating uncertainty. On the other hand, a system itself has alternative structures that increase the level of uncertainty, such as a centralized, decentralized, and distributed system.

\subsection{Machines}
Systems run on machines. Its certainty rely on machines as well. If machines should have non-functional hardware or software, or compromised software, an uncertainty level can increase.

\subsection{Networks}
Networks, as the carrier of information, will create uncertainty. Congestion is normal in networks, which cause transfer delay, and networks itself has the vulnerability to cyber attacks, or sometimes result in unavailability. All those conditions generate uncertainty to the systems who use networks.

\section{Procedures}
Uncertainty is derived by two kinds of procedures. One is based on objective evidence, which should be able to be implemented by formal and repeatable procedure. The other is based on subjective evidence, which includes beliefs, judgements, and opinions that are mostly generated by humans who use bias or subjective feeling in their decision-making process.

\section*{Acknowledgement}
This work is partly supported by the Army Research Office under Grant Contract Number W91NF-20-2-0140 and NSF under the Grant Number 2107449, 2107450, and 2107451. The views and conclusions contained in this document are those of the authors and should not be interpreted as representing the official policies, either expressed or implied, of the Army Research Laboratory or the U.S. Government. The U.S. Government is authorized to reproduce and distribute reprints for Government purposes notwithstanding any copyright notation herein.

\bibliographystyle{ACM-Reference-Format}
\bibliography{ref}